\documentclass{article} %
\usepackage{times}

\usepackage{amsmath,amsfonts,bm}

\def\eqref#1{equation~\ref{#1}}
\def\1{\bm{1}}

\DeclareMathAlphabet{\mathsfit}{\encodingdefault}{\sfdefault}{m}{sl}
\SetMathAlphabet{\mathsfit}{bold}{\encodingdefault}{\sfdefault}{bx}{n}

\usepackage{natbib}
\setcitestyle{authoryear,round,citesep={;},aysep={,},yysep={;}}
\usepackage{microtype}
\usepackage{hyperref}
\usepackage{graphicx}
\usepackage{booktabs}
\usepackage{booktabs}
\usepackage{subcaption}
\usepackage{url}
\usepackage{pdfpages}
\usepackage{tikz}
\usepackage[disable,textsize=tiny,textwidth=8em,backgroundcolor=yellow]{todonotes}

\usepackage[textwidth=6in]{geometry}
\usepackage{wrapfig}
\usepackage{lipsum}
\usepackage{caption}

\newcommand{\insertFancyImageForFigure}[9]
{ %
  \hspace{0.2em}
  \begin{minipage}{2.2em}
    \vspace{-#9}\vspace{-#9}\vspace{-#9}\vspace{-#9}\vspace{-#9}\vspace{-#9}\vspace{-#9}\vspace{-#9}\vspace{-#9}\vspace{-#9}
    \noindent \\ \vspace{#9}
    \textsf{adv} \\ \vspace{#9}
    \textsf{ret} \\ \vspace{#9}
    \textsf{agr} \\ \vspace{#9}
    \textsf{conf} \\ \vspace{#9}
    \textsf{priv}
  \end{minipage}
\begin{tikzpicture}
    \draw (0, 6) node[inner sep=0] {\includegraphics[width=0.92\textwidth]{#1}};
    \shade[left color=red,right color=green] (-0.09-#2+#3,#6) rectangle (0.05-#2+#3,#8);
    \draw [line width=0.165cm, white, dash pattern={on 0.02cm off 0.1cm}] (-0.02-#2+#3,#7) --(-0.02-#2+#3,#8);
    #4
\end{tikzpicture} %
\vspace{#5}
}
\newcommand{\insertFancyImageForFigureImagenet}[9]
{ %
  \hspace{0.2em}
  \begin{minipage}{2.2em}
    \vspace{-#9}\vspace{-#9}\vspace{-#9}\vspace{-#9}\vspace{-#9}\vspace{-#9}\vspace{-#9}\vspace{-#9}\vspace{-#9}\vspace{-#9}\vspace{-#9}\vspace{-#9}\vspace{-#9}
    \noindent \\ \vspace{#9}
    \textsf{adv} \\ \vspace{#9}
    \textsf{agr} \\ \vspace{#9}
    \textsf{conf}
  \end{minipage}
\begin{tikzpicture}
    \draw (0, 6) node[inner sep=0] {\includegraphics[width=0.92\textwidth]{#1}};
    \shade[left color=red,right color=green] (-0.09-#2+#3,#6) rectangle (0.05-#2+#3,#8);
    \draw [line width=0.165cm, white, dash pattern={on 0.02cm off 0.1cm}] (-0.02-#2+#3,#7) --(-0.02-#2+#3,#8);
    #4
\end{tikzpicture} %
\vspace*{#5}
}
\newcommand{\insertOutliersToPrototypesImage}[3]
{ %
  \insertFancyImageForFigure{#1}{#2}{#3}{}{0.1cm}{3.95}{4}{8.1}{1.1em}
}

\newcommand{\insertOutliersToPrototypesImageImagenet}[3]
{ %
  \insertFancyImageForFigureImagenet{#1}{#2}{#3}{}{-1cm}{5.6}{5.65}{8.1}{1.1em}
}

\begin{document}

\title{\vspace{-4em}Distribution Density, Tails, and Outliers in Machine Learning: Metrics and Applications}

\author{Nicholas Carlini \qquad {\'U}lfar Erlingsson \qquad Nicolas Papernot \\ \emph{Google Research}}
\date{}

\maketitle

\begin{abstract}

  We develop techniques to quantify the degree to which a given
  (training or testing) example is an outlier in the underlying distribution.
  We evaluate five methods to score examples
  in a dataset by how
  well-represented the examples are,
  for different plausible definitions of ``well-represented'',
  and apply these to four common datasets:
  MNIST, Fashion-MNIST, CIFAR-10, and ImageNet.
  Despite being independent approaches, we find all five
  are highly correlated, suggesting that the notion of being well-represented
  can be quantified.
  Among other uses, we find these methods can be combined to identify
  (a) \emph{prototypical examples} (that match human expectations);
  (b) \emph{memorized} training examples; and,
  (c) \emph{uncommon submodes} of the dataset.
  Further, we show how we can utilize our metrics to determine an improved ordering
  for curriculum learning, and impact adversarial robustness.  
  We release all metric values on training and test sets we studied. 
\end{abstract}

\makeatletter
\newcommand*{\centerfloat}{%
  \parindent \z@
  \leftskip \z@ \@plus 1fil \@minus \textwidth
  \rightskip\leftskip
  \parfillskip \z@skip}
\makeatother

\vspace{2em}
\begin{center}
  \centerfloat
  \includegraphics{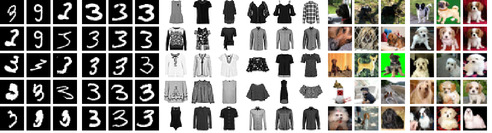}
      
      \captionof{figure}{
        Sorting images sampled from the MNIST ``3'' class, Fashion-MNIST ``shirt''
        class, and CIFAR-10 ``dog'' class using our five metrics.
        Outliers are shown on the left, and well represented examples on the right.
        Notice, for example, the mislabeled ``9'' for MNIST as a digit that is an outlier, or how
        many of the poorly represented Fashion-MNIST ``shirts'' in fact belong in the
        different ``t-shirt'' or ``dress'' class.
      }
	\label{fig:big_figure}
    \end{center}

\section{Introduction}

Machine learning (ML) is now applied to problems with sufficiently large
datasets that it is difficult to manually inspect each training
and test point.
This
drives interest in research that seeks to understand the dataset and
the underlying data distribution.
Potential uses of these techniques are numerous. 
On the one hand, they contribute to  improving
how ML is perceived by end users (e.g., 
one of the motivations behind interpretability
efforts). On the other hand, they also help
ML practitioners glean insights into the learning procedure.
This surfaces the need for tools that enable one to (1) measure and
characterize the contribution of each training point to the learning
procedure and (2) explain the different failure modes observed
on individual test points when the model infers.

Towards this goal, prior work has investigated model
interpretability, identifying
training and test points that
are prototypical~\citep{kim2014bayesian}, or
applying \emph{influence functions} to measure the 
contribution of individual training points to the final model~\citep{koh2017understanding}.
While defining precisely a prototype remains
an open problem, a common intuitive definition of the notion is 
that prototypes should be ``a relatively small number of samples
from a data set which, if well chosen, can  serve  as 
a summary of the original data set''~\citep{bien2011prototype}.
In addition to these two examples, 
there is a wealth of related efforts discussed below.

In this work, we take an orthogonal direction and show
that rather than  trying to identify a single metric or
technique
to identify ``prototypes'', simultaneously considering a variety 
of metrics can be more effective to discover properties
of the training data.
In particular, we introduce five metrics for measuring to what extent
a specific point is well represented or an outlier in a dataset.

We explicitly \emph{do not define what we mean by well-represented}
or outlier specifically
because we are interested in the interplay between different
metrics that may fall under that definition.
Indeed, we find that while the different metrics are highly
correlated for most training and test inputs, their disagreement
are highly informative.

In more detail, our metrics are based on 
adversarial robustness,
retraining stability, 
ensemble agreement,
and differentially-private learning. 
We demonstrate that in addition to supporting
use cases previously studied in the literature 
(e.g., identifying
prototypes), studying the interplay between these five 
metrics allows us to identify other types of 
 examples that help form an
 understanding of the training and inference procedures.
 They provide a more complete picture of a model's training
 and test performance than can be captured by accuracy alone.
 For instance, disagreements between our metrics distinguish \textit{memorized training examples}---that 
 models overfit on to in order to learn,
 or \textit{uncommon submodes}---not sufficiently well-represented in the
training data for a privacy-preserving model to 
recognize them at test time. 
These results hold for all the datasets we consider: 
MNIST, Fashion-MNIST, CIFAR-10, and ImageNet.
We release the results of running our metrics on these
datasets
to help other researchers interested in building
on our results.

Usefully,
there are advantages to
training models using only the well-represented examples:
the models learn much faster,
their accuracy loss is not great
and occurs almost entirely on outlier test examples,
and the models are both 
easier to interpret and more adversarially robust. 
Conversely, at the same sample complexity,
significantly higher overall accuracy can be achieved by
training models exclusively on outliers---once 
erroneous and misleading examples
have been eliminated from the dataset %
automatically through an analysis of the disagreement between our metrics.

As an independent result,
we show that predictive stability under retraining
strongly correlates with adversarial distance,
and may be used as an approximation. This is 
particularly interesting for tasks where defining the adversary's goal
when creating an adversarial example~\citep{biggio2013evasion,szegedy2013intriguing} 
can be difficult (e.g., in sequence-to-sequence
language modeling).

\section{Identifying Outlier Examples}
\label{sec:def}
\label{sec:desire}

It is important to understand the underlying
datasets (both training and testing) used for machine learning
models.
In the following, we introduce the five metrics
that underly our approach for interpreting datasets.
Each metric we develop scores examples on a continuum where
in one direction the examples are somehow more well-represented
in the dataset, and the other direction they are less
represented---more of an outlier---in the dataset.

We do not define \emph{a priori}
what we mean by well-represented: %
Rather,
we define the term with respect to our different algorithms for
computing this.
As we will demonstrate, our rankings agree with the
definition of prototypes in many ways.
However, their disagreement are useful to identify training
and test points that are important for forming an understanding
of the training and inference procedures.
Indeed, in Section~\ref{ssec:comparing-metrics} we demonstrate
how the metrics allow us to identify memorized exceptions
or uncommon submodes at scale in the data.

\subsection{Metrics for Identifying Representitive Examples}
\label{ssec:metrics}
Each of the metrics below begins corresponds to an definition for what
one might mean by saying an example is representative or an outlier.
For each, we provide a concrete method
for measuring this informally-specified quantity.

We study five metrics that we found generalizable and useful;
clearly these are not the only possible metrics, and we encourage future
work to study other metrics.
However, we believe these metrics to cover a wide range of what one
might mean by representative.
Other definitions which we considered were either unstable \footnote{
In one attempt at a metric, we defined how representative an example is
with respect to the magnitude of the gradient of the loss function on
a pre-trained model and found it
varied significantly across different pre-trained
models.} or model-specific \footnote{In another metric we rejected, we 
found that sorting examples by when they were
learned during the training process gave different orderings when
applied to 
different model architectures.}.
All of the algorithms we give below are both stable and appear to
be consistent properties of the training data, and not the model
(e.g., architecture).

\textbf{Adversarial Robustness (\textsf{adv}):}
Examples that well represent the dataset should be
more adversarially robust, i.e., more difficult to find an input perturbation which
makes them change classification.
Indeed, as a measure of \emph{prototypicality},
this exact measure (the distance to the decision boundary measured by an adversarial-example attack
was)
was recently proposed and utilized by~\citet{stock2017convnets}.
Specifically, for an example $x$,
the measure finds the perturbation $\delta$ with minimal  $\lVert\delta\rVert$
such that the original $x$ and the adversarial example
$x+\delta$ are classified differently~\citep{biggio2013evasion,szegedy2013intriguing}.

To compare prototypicality,
the work
of~\citet{stock2017convnets}
that inspired our current work
used a simple and efficient $\ell_\infty$-based adversarial-example attack
based on an iterative gradient descent introduced
by~\citet{kurakin2016adversarial}. That attack procedure computes
gradients to find directions that will increase
the model's loss on the input within an $\ell_\infty$-norm ball. 
They define prototypicality as the number of gradient descent iterations necessary
to change the class of the perturbed input.

Instead our metric (for short, \textsf{adv})
ranks by the 
$\ell_2$ norm
(or faster, less accurate $\ell_\infty$ norm)
of the minimal-found adversarial perturbation~\citep{carlini2017towards}.
This is generally more accurate at measuring the distance to
the decision boundary, but comes at a
performance cost (it is on average 10-100$\times$ slower).

\textbf{Holdout Retraining (\textsf{ret}):}
A model should treat a well-represented example the same
regardless of whether or not it is used in the training process:
if the example is not used, a well-represented example should have
sufficient support in the training data for its omission to not be important.

Assume we are given a training dataset $\mathcal{X}$, a disjoint holdout dataset
$\mathcal{\bar{X}}$, and an example $x \in \mathcal{X}$ to assess how represented it
is in the dataset.
To begin, we train a model $f(\cdot)$ on the data $\mathcal{X}$ to obtain
model weights $\theta$.
We train this model just as how we would typically do---i.e., with the same
learning rate schedule, hyper-parameter settings, etc.
Then, we fine-tune the weights of this first model $f_\theta(\cdot)$ on the held-out training data $\mathcal{\bar{X}}$ 
to obtain new weights $\bar\theta$.
To perform this fine-tuning, we use a smaller learning rate and train until the training
loss stops decreasing.
(We have found it is important to obtain $\bar\theta$ by fine-tuning $\theta$ as opposed to
training from scratch; otherwise, the randomness of training 
leads to unstable rankings that
yield specious results.)
Finally, given these two models,
we measure how well-represented the example $x$ is as the difference
$\lVert{}f_\theta(x)-f_{\bar\theta}(x)\rVert$.
The exact choice of metric $\lVert \cdot \rVert$ is not important;
the results in this paper use the symmetric KL-divergence.

While this metric is similar
to the one considered in~\citep{ren2018learning}, it differs in important ways:
notably, our holdout retraining metric is conceptually simpler, more stable 
numerically, and  more computationally efficient
(because it does not require a backward pass to estimate gradients in addition
to the forward pass needed to compare model outputs).
Since our metric is only meaningful for data used to train the model, in
order to measure how well represented arbitrary test points are,
we actually \emph{train on the test data}
and perform holdout retraining on the original training data.

\textbf{Ensemble Agreement (\textsf{agr}):}
Well-represented examples should be easy for many types of models to learn,
and not only models which are nearly perfect.
We train multiple models of varying capacity (i.e., number of parameters)
on different subsets of the
training data (see Appendix~\ref{apx:protofigs}).
The \textsf{agr} metric ranks
examples based on
the agreement within this ensemble,
as measured by the symmetric JS-divergence between the models' output.
Concretely, we train many
models $f_{\theta_i}(\cdot)$ and,
for each example $x$,
evaluate
the model
predictions, and then compute the following value to order the examples:
\begin{equation}
{1 \over N^2}\sum_{i=1}^N \sum_{j=1}^N \text{JS-Divergence}(f_{\theta_i}(x), f_{\theta_j}(x))
\end{equation}

\textbf{Model Confidence (\textsf{conf}):}
Models should be confident on examples that are well-represented.
Based on an ensemble of models
like that used by the \textsf{agr} metric,
the \textsf{conf} metric
ranks examples by
the mean confidence in the
models' predictions,
i.e., ranking each example $x$
by:
\begin{equation}
{1 \over N}\sum_{i=1}^N \max f_{\theta_i}(x)
\end{equation}

\textbf{Privacy-preserving Training (\textsf{priv}):} 
We can expect well-represented examples 
to be classified properly by models
even when trained
with guarantees of
differential privacy~\citep{abadi2016deep,papernot2016semi}.
(Informally, differential privacy states that whether or not any
given training example is in the training data, the learned models
will be statistically indistinguishable.)
However,
such privacy-preserving models
should exhibit significantly reduced accuracy
on any rare or exceptional examples,
because differentially-private learning
attenuates gradients and
introduces noise to prevent
the details about
any specific training examples from being memorized.
Outliers are disproportionally likely to be impacted
by this attenuation and added noise,
whereas the common signal found
across well-represented examples must have been preserved
in models trained to reasonable accuracy.

Our \textsf{priv} metric
is based on training an ensemble of models with increasingly greater $\varepsilon$ privacy
(i.e., more attenuation and noise)
using $\varepsilon$-differentially-private
stochastic gradient descent~\citep{abadi2016deep}.
Our metric
then ranks how well-represented an example is
based on the minimum $\varepsilon$ 
(i.e., maximum privacy protection)
at which the example is
correctly classified
in a reliable manner
(which we take as being also classified correctly
in 90\% of less-private models).
This ranking
embodies the intuition that
the more tolerant an example is to noise and attenuation
during learning, the more well-represented it must be.

\begin{figure*}
  \centering
  \begin{subfigure}{.4\columnwidth}
    \includegraphics[width=\linewidth]{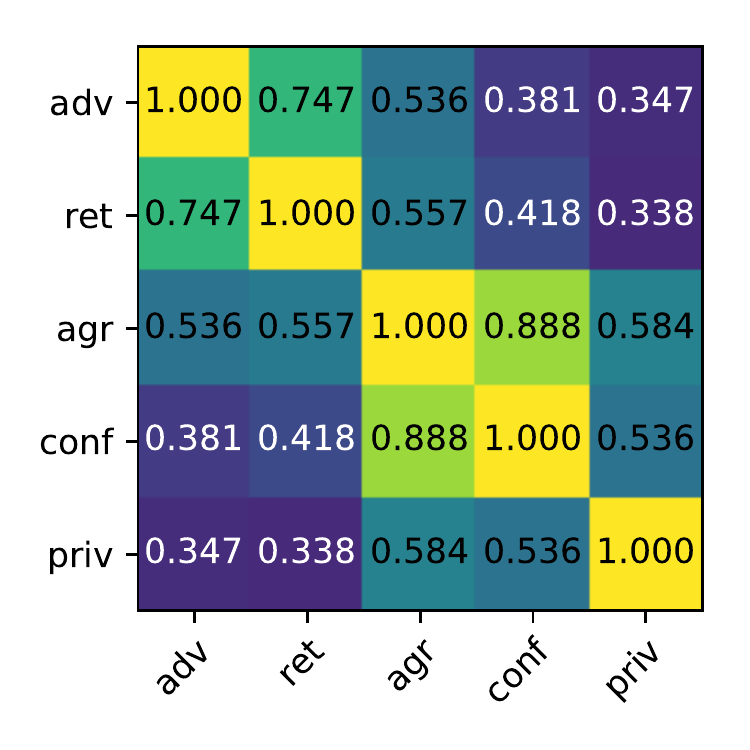}
    \caption{MNIST\vspace*{-.5ex}}
  \end{subfigure}
  \begin{subfigure}{.4\columnwidth}
    \includegraphics[width=\linewidth]{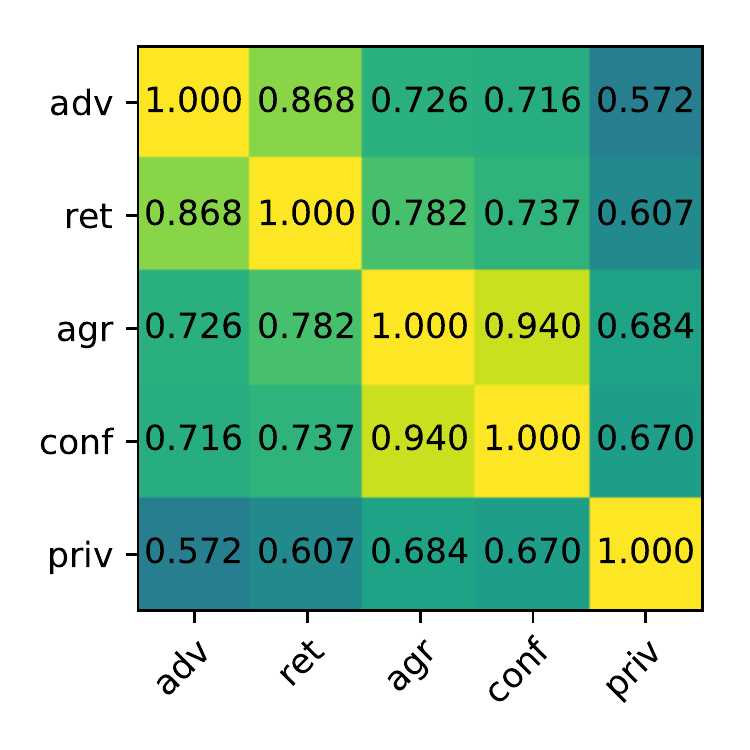}
    \caption{Fashion-MNIST\vspace*{-.5ex}}
  \end{subfigure}
  
  \begin{subfigure}{.4\columnwidth}
    \includegraphics[width=\linewidth]{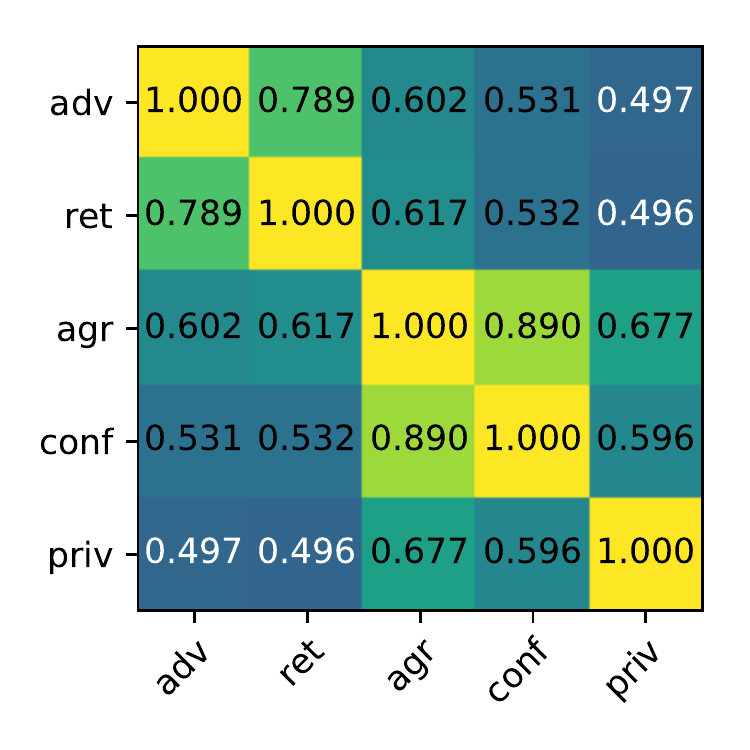}
    \caption{CIFAR-10\vspace*{-.5ex}}
  \end{subfigure}
  \begin{subfigure}{.4\columnwidth}
    \includegraphics[width=\linewidth]{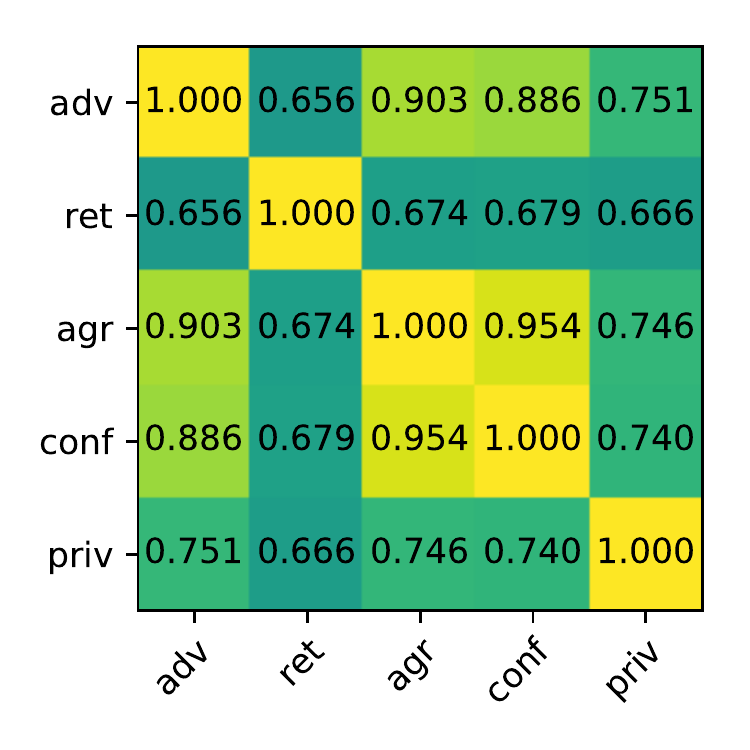}
    \caption{ImageNet\vspace*{-.5ex}}
  \end{subfigure}
\caption{Correlation coefficients for our five prototypicality metrics
on four common datasets.\vspace*{-2ex}}
\label{tbl:correlations}
\end{figure*}

\section{Evaluating the Five Different Metrics}
\label{sec:eval}

As the first step in an evaluation,
it is natural to consider to what extent
these metrics are different methods of evaluating the
same underlying property.
We find that the are highly correlated across the four
datasets we study: MNIST~\citep{lecun2010mnist}, Fashion-MNIST~\citep{xiao2017/online}, CIFAR-10~\citep{krizhevsky2009learning}, and ImageNet~\citep{russakovsky2015imagenet}.
In particular, we observe a strong correlation between 
the adversarial distance and holdout retraining metrics, 
which is of independent interest: the holdout 
retraining metric could serve as a substitute for adversarial
distance in tasks where adversarial examples are ill-defined.

Our metrics
are widely applicable, as they
are not specific to any learning task or model
(some, like
\textsf{ret} and \text{priv}
might be applicable even to unsupervised learning),
and experimentally
we have confirmed
that the metrics are model-agnostic in the sense that
they give overall the same results
despite large changes in hyperparameters
or even the model architecture.
We also show that our metrics are consistent
with human perception of representativeness.

An important application of our metrics is that
studying their (relatively rare) disagreements allow us
to inspect datasets at scale. 
In particular, we show how to identify two types
of examples:
\textit{memorized exceptions} and 
\textit{uncommon submodes}.
They can be used to form an understanding of
performance at training and test time that is more
precise than what an accuracy measurement can offer.
(Due to space constraints,
experimental results supporting some of the above observations
are given in the Appendix.)

We release the full results of running our metrics on each of the
four datasets in the Appendix; we encourage the
interested reader to examine the results, we believe the
results speak for themselves.

\begin{table*}
  \begin{tabular}{llll}
    & \multicolumn{1}{c}{MNIST} & \multicolumn{1}{c}{Fashion-MNIST} & \multicolumn{1}{c}{CIFAR-10} \\
    \rotatebox{90}{\hspace{1.2em}Pick Worst} & 
    \includegraphics[width=.29\linewidth,trim={0 0 0 1cm},clip]{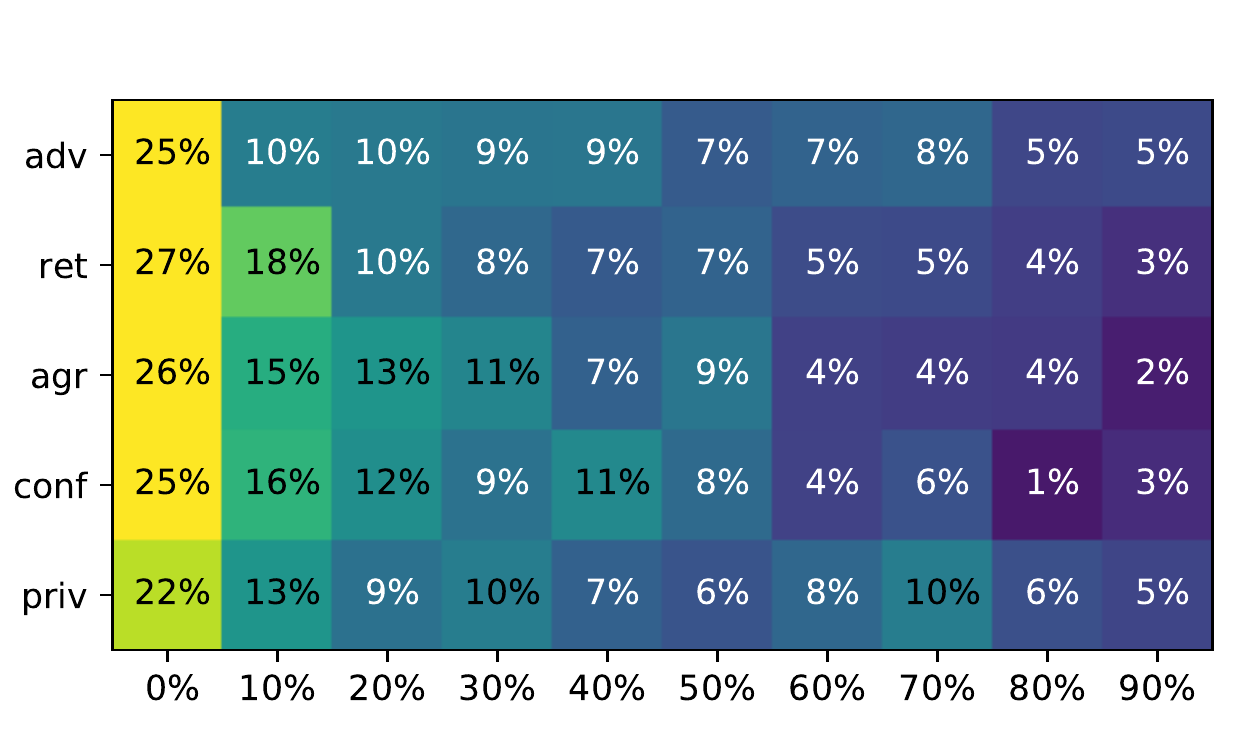} &
    \includegraphics[width=.29\linewidth,trim={0 0 0 1cm},clip]{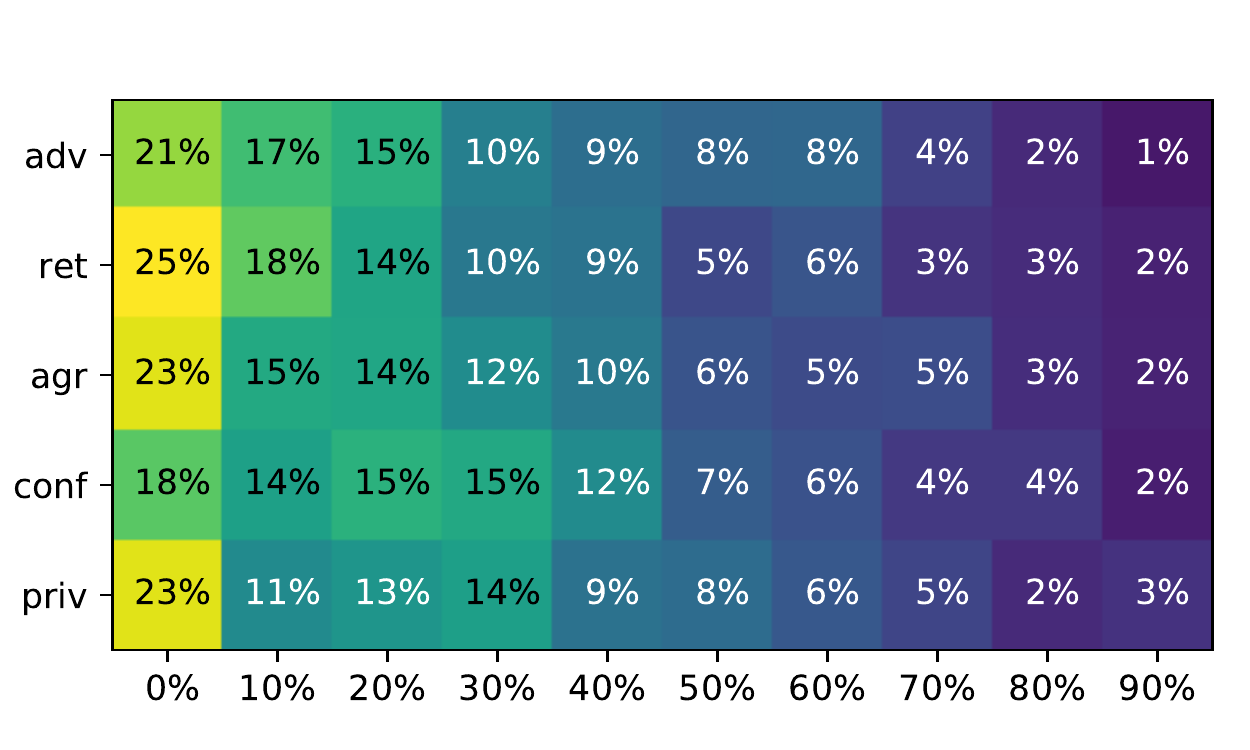} &
    \includegraphics[width=.29\linewidth,trim={0 0 0 1cm},clip]{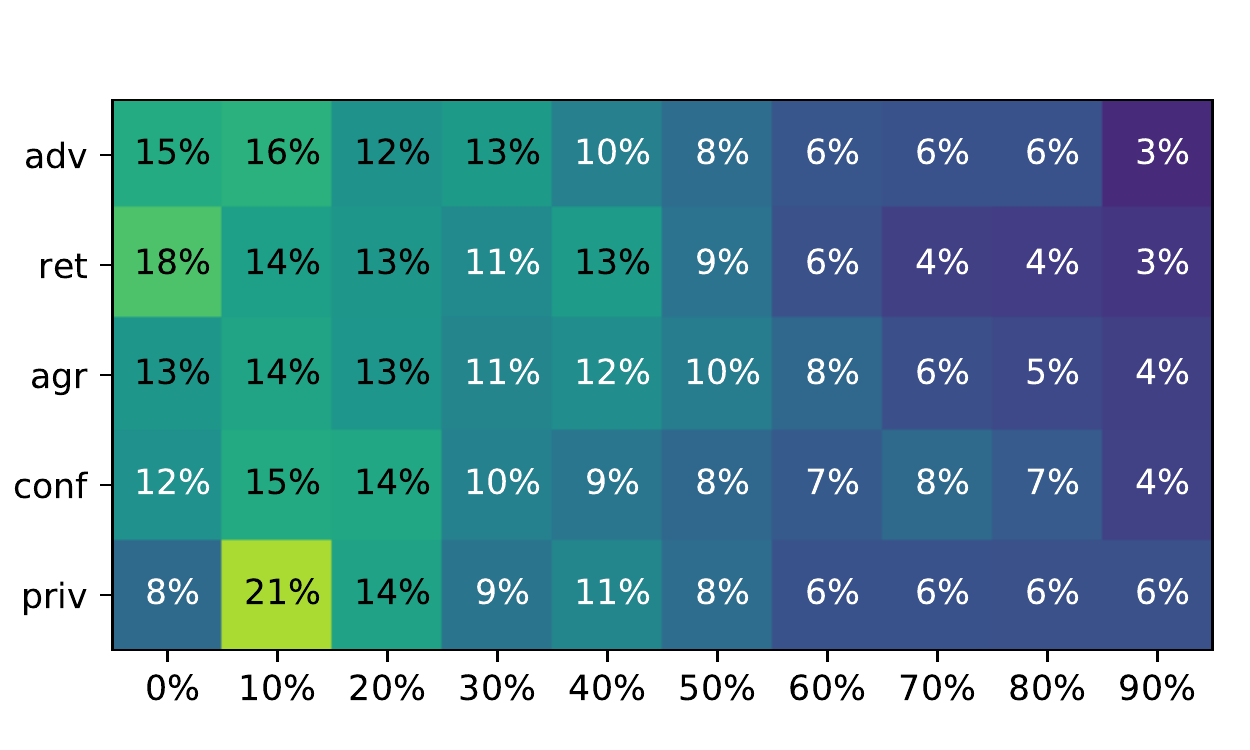} \\
    \rotatebox{90}{\hspace{1.5em}Pick Best} & 
    \includegraphics[width=.29\linewidth,trim={0 0 0 1cm},clip]{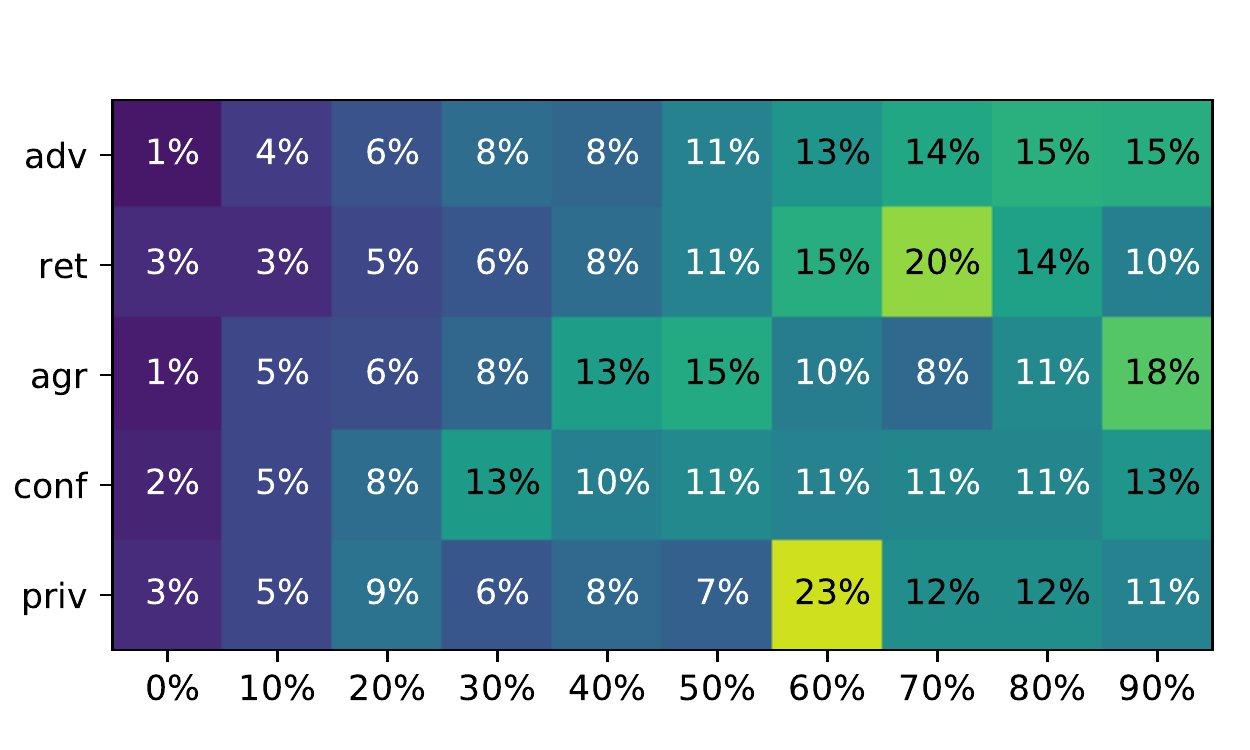} &
    \includegraphics[width=.29\linewidth,trim={0 0 0 1cm},clip]{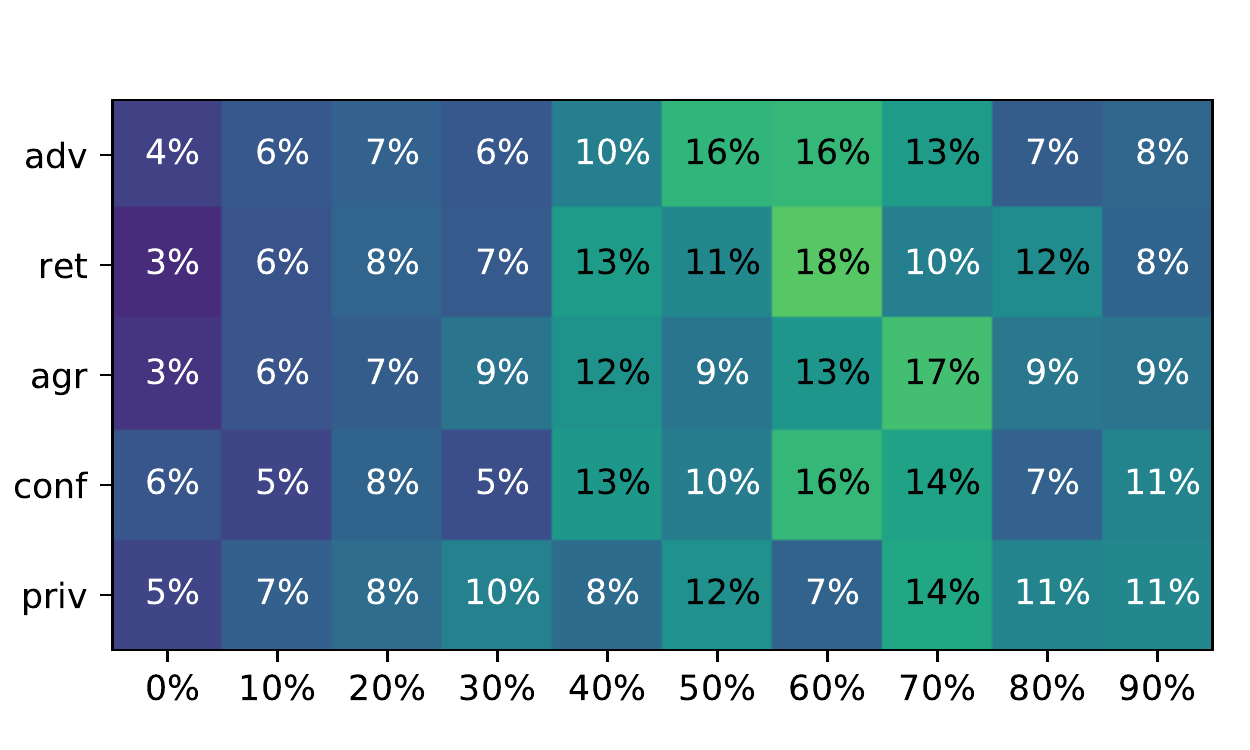} &
    \includegraphics[width=.29\linewidth,trim={0 0 0 1cm},clip]{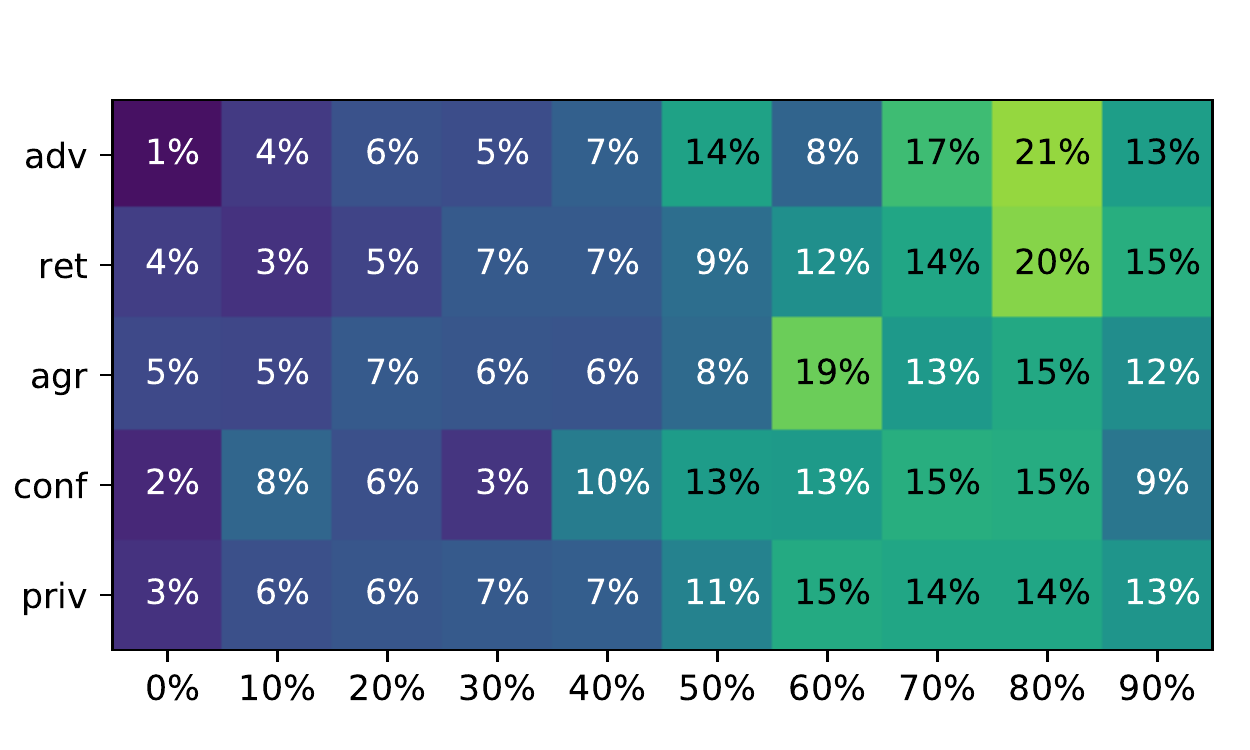} \\
  \end{tabular}
  \vspace{-1em}
  \caption{Results of a human study of
    Mechanical Turk workers selecting the best or worst example among a
    random collection of 9 training-data images.
    For each metric,
    the tables show what percent of workers
    selected examples
    in each 10\% split
    of the metric's sorted ranking
    (e.g.,     
    when shown MNIST digits, $25\%$ human workers selected as worst an example
    that fell in the bottom $10\%$ of examples as ranked by the \textsf{adv} metric.\vspace*{-3ex}
  }
  \label{tbl:humanstudy}

\end{table*}

\subsection{Correlations Between Metrics}

Figure~\ref{tbl:correlations} shows the correlation coefficients
computed pairwise between each of our metrics
for all our datasets,
(the tables are symmetric across the diagonal).
The metrics are overall strongly correlated,
and the differences in correlation
are informative.
Unsurprisingly,
since they measure very similar properties, 
the \textsf{agr} (ensemble agreement) and 
the \textsf{conf} (model confidence)
show the 
highest correlation,

However,
somewhat unexpectedly,
we find that the \textsf{adv} metric (adversarial robustness)
correlates very strongly with the \textsf{ret} metric (retraining distance)
on the smaller three datasets.
This is presumably
because these two metrics 
both measure the distance to a model's decision boundary---even though
\textsf{adv} measures this distance by perturbing each example
while 
\textsf{ret} measures how
the evaluation of each example
is affected when models' decision boundaries themselves
are perturbed.
On ImageNet, \textsf{adv} is most strongly correlated with
ensemble agreement and model confidence; we hypothesize this
is due to the fact that ImageNet is a much more challenging task
and therefore the distance to the decision boundary can be best
approximated by the initial model confidence, unlike on MNIST
where most datapoints (even incorrectly labeled) are assigned probability $0.999$ or higher.

This strong correlation
between \textsf{adv} and \textsf{ret}
is a new result that may be of independent interest
and some significance.
Measurement of adversarial distance is a useful and highly-utilize technique,
but it is undefined or ill-defined on many learning tasks
and its computation is difficult, expensive, and hard to calibrate.
On the other hand,
given any holdout dataset
and any measure of divergence,
the \textsf{ret} metric we define in Section~\ref{ssec:metrics}
should be easily computable for any ML model or task.

\subsection{\mbox{Qualitative Evaluation Inspection and Human Study}}
\label{sec:humans}

Clearly these metrics do consistently measure \emph{some}
quantity which is inspired by how well
represented individual examples are,
but we have not yet provided evidence that it does so.
To do this, we will show that our methods for identifying examples that
are well represented in the dataset match human intuition for examples
that are of a high quality.

To begin, we
perform a subjective visual inspection
of how the different metrics
rank the example training and test data on different datasets.
As a representative example,
Figure~\ref{fig:big_figure} (Page 1)
and the figures in Appendix~\ref{apx:protofigs}
confirm that there is an obviously apparent
difference between the two extremes on
the MNIST, Fashion-MNIST, and CIFAR-10
training
examples (ImageNet examples are given in the Appendix).

To more rigorously
validate and quantify
how our metrics correlate 
with human perception,
we performed an online human study
using Amazon's Mechanical Turk service.
We presented human evaluators a collection of 
images (all of one class) from the
MNIST, Fashion-MNIST, or CIFAR-10 datasets.
We asked the evaluator to select 
either the image that was \emph{most} or \emph{least} representative
of the class.
In the study, 400 different human evaluators
assessed over 100,000 images.
At a high level, the human evaluators largely agreed
that the images selected by our algorithms as most
representative do in fact match human intuition.

Concretely,
in this study
(an image of the study form is given in the Appendix, Section~\ref{apx:humanstudy}),
each human evaluator
saw a 3x3 grid of 9 random images
and was asked to pick the worst image---or the best image---and
this was repeated multiple times.
Evaluators
exclusively 
picked either
best or worst images
and were only shown random images from one output class
under a heading with the label name of that class;
thus one person would pick
only the best MNIST digits ``7''
while another picked only the worst CIFAR-10 ``cars.''
(As dictated by good study design,
we inserted ``Gold Standard'' questions
with known answers to catch
workers answering randomly or incorrectly,
eliminating the few such workers from our data.)
For all datasets,
picking non-representative images proved to be the easier task:
in a side study where 50 evaluators
were shown the same identical 3x3 grids, 
agreement was $80\%$ on the worst image
but only $27\%$ on the best image
(random choice would give $11\%$ agreement).

The results of our human study are presented in Table~\ref{tbl:humanstudy}.
The key takeaway is that the evaluator's assessment is correlated 
with each one of our metrics: evaluators
mostly picked the not well-represented images as the worst examples
and examples that were better represented  as being the best.

\subsection{Comparing Metrics and their Characteristics}
\label{ssec:comparing-metrics}

\begin{figure}
  \centering
    \includegraphics[width=0.5\textwidth]{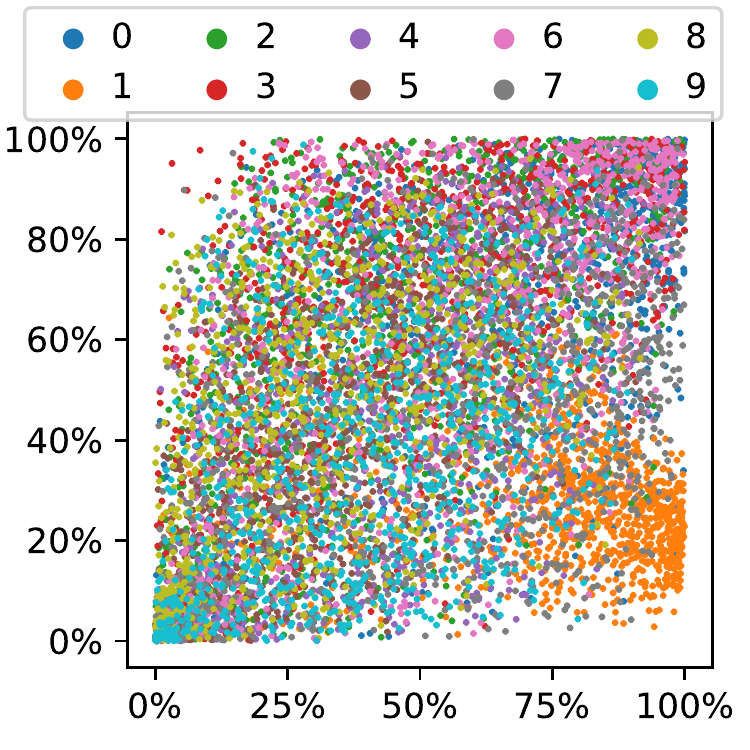}
    \caption{Scatter plot comparing the \textsf{adv} vs.\ \textsf{conf} ranks
      on the MNIST test set. Notice clusters of digits, for examples the 1 class
      that has extremely high average confidence (i.e., they are easy to classify) but
      low adversarial distance (i.e., are easy to perturb to other classes).}
    \label{fig:metricplots}
\end{figure}

Because our five metrics are not perfectly correlated,
there are likely to be many examples
that are determined to be well-represented under one metric but not under another,
as a consequence of the fact that each metric defines ``well-represented'' differently.
To quantify the number and types of
those differences
we can try
looking at their visual correlation
in a scatter plot;
doing so can be informative,
as can be seen in Figure~\ref{fig:metricplots}(a)
where the easily-learned, yet fragile,
examples of class ``1''
in MNIST models
have high confidence
but low adversarial robustness.
The results show substantial 
disagreement between metrics.

To understand disagreements,
we can consider examples that are well represented
in one metric but \emph{not} in others,
first
combining the union of \textsf{adv} and \textsf{ret}
into a single \textsf{boundary} metric,
and the union of \textsf{adv} and \textsf{ret}
into an \textsf{ensemble} metric,
because of their high correlation.

\textbf{Memorized exceptions:~}
Recalling
the unusual dress-looking ``shirt''
of Figure~\ref{fig:big_figure},
and how it seemed to have been memorized with high confidence,
we can intersect the top 25\% well represented
\textsf{ensemble} images
with the bottom-half outliers in both
the \textsf{boundary} and \textsf{priv} metrics.

\begin{wrapfigure}{r}{0.31\textwidth} %
    \vspace*{-3ex}
    \includegraphics[width=0.31\textwidth,trim=2cm 1.2cm 1.6cm 1.2cm,clip]{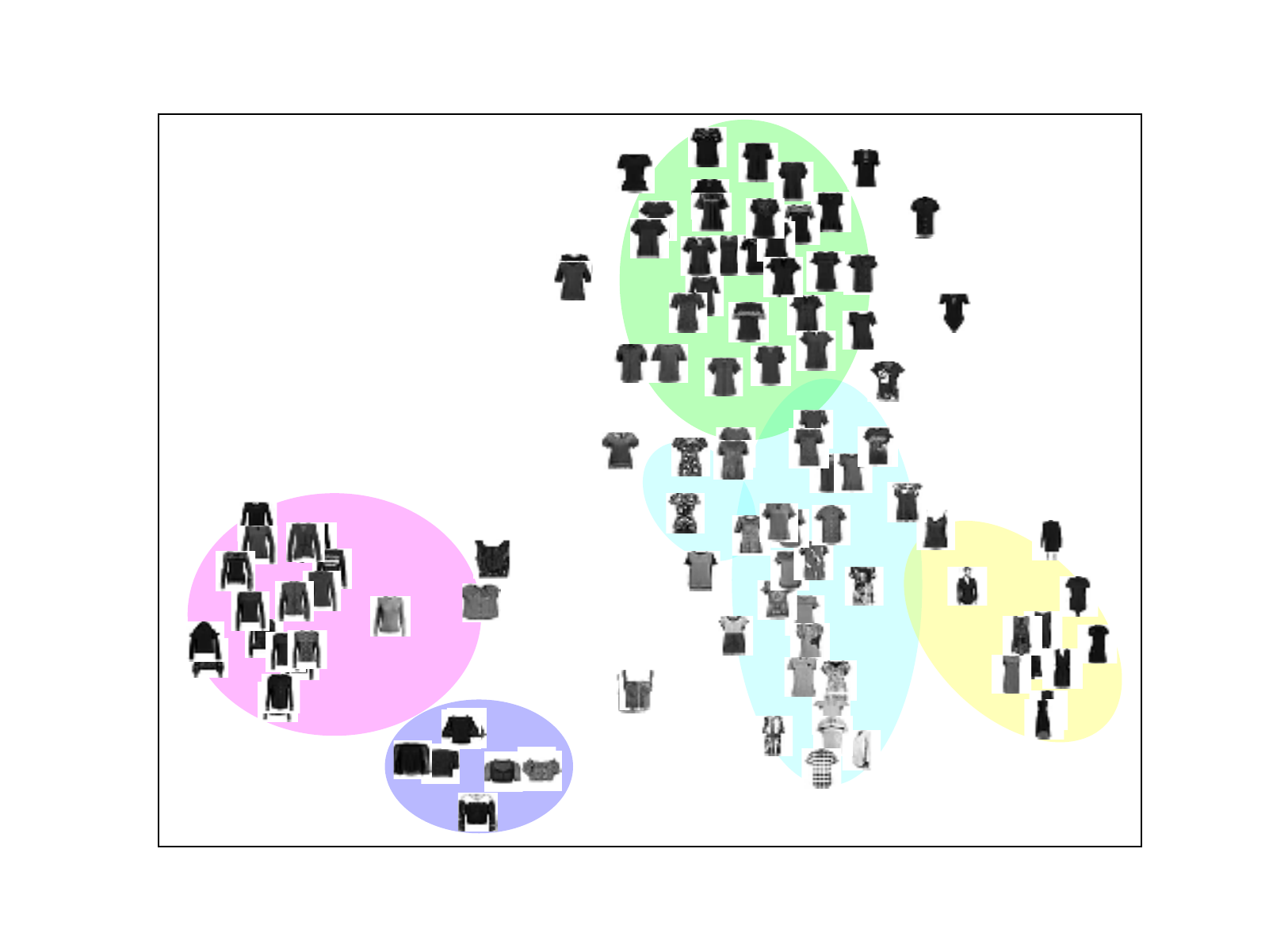}
    \caption{Exceptional ``shirts.''}
    \label{fig:inline}
\end{wrapfigure}
\begin{figure*}
  \begin{subfigure}{.31\textwidth}
	\includegraphics[width=\linewidth,trim=2cm 1cm 1.6cm 1cm,clip]{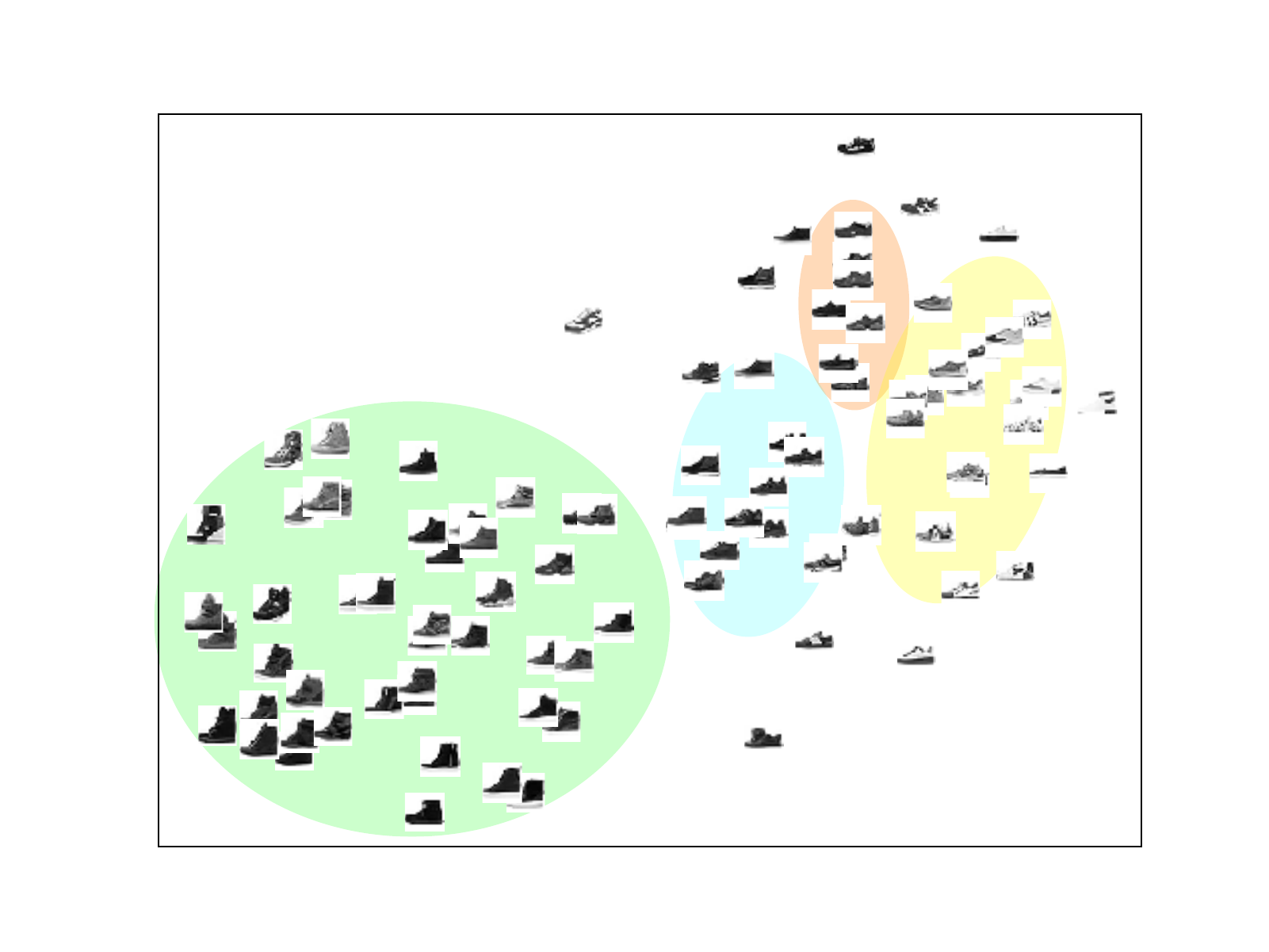}
	\caption{\textbf{Memorized exceptions} in the Fashion-MNIST ``sneaker'' class.\vspace*{-1ex}}
        \label{fig:clustersa}
\end{subfigure}
	\hspace*{\fill}  
\begin{subfigure}{.31\textwidth}
    \includegraphics[width=\linewidth,trim=2cm 1cm 1.6cm 1cm,clip]{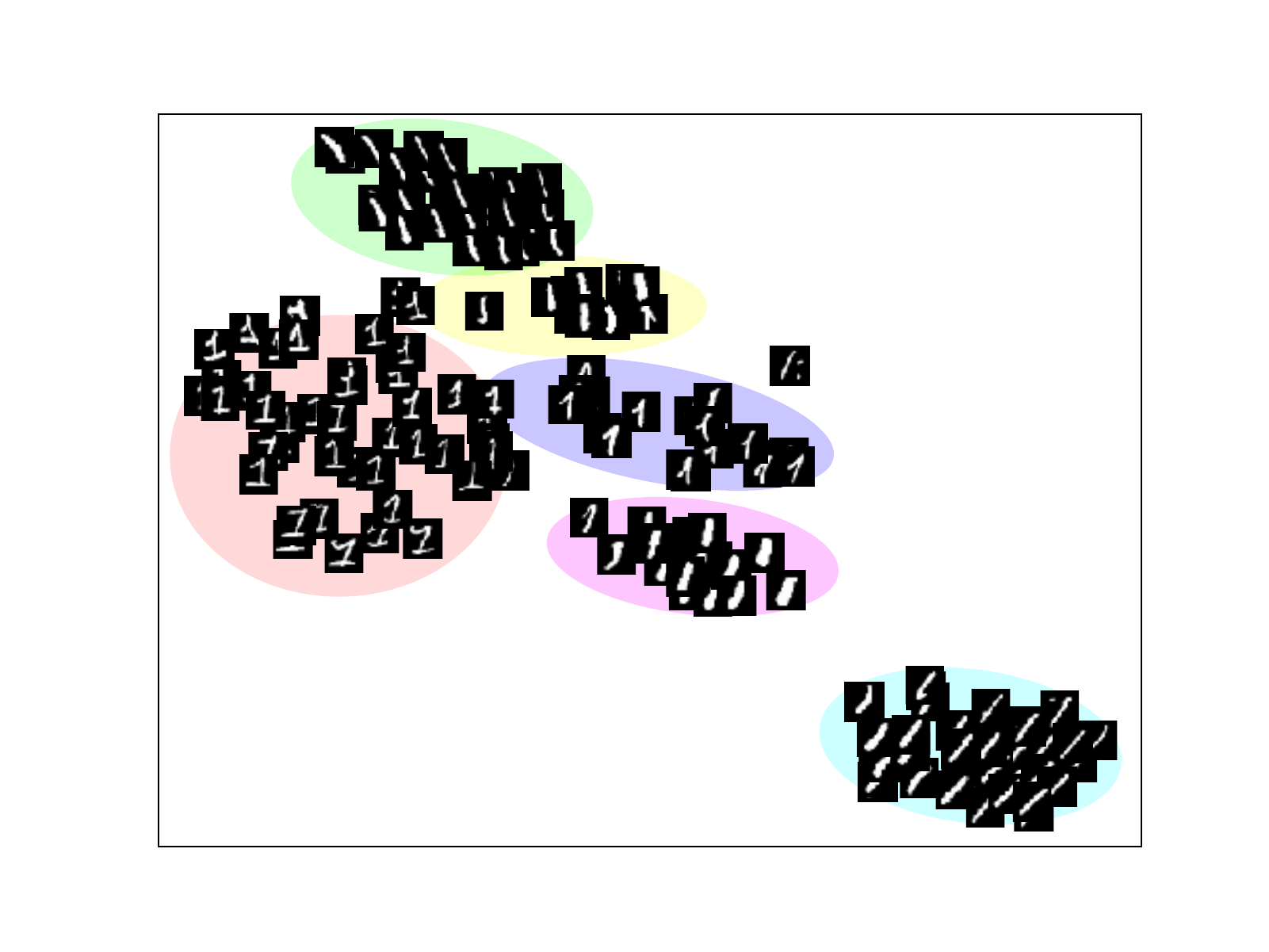}
    \caption{\textbf{Uncommon submodes} found within the MNIST ``1'' class.\vspace*{-1ex}}
    \label{fig:clustersb}
  \end{subfigure}
  \hspace*{\fill}
  \begin{subfigure}{.31\textwidth}
    \includegraphics[width=\linewidth,trim=2cm 1cm 1.6cm 1cm,clip]{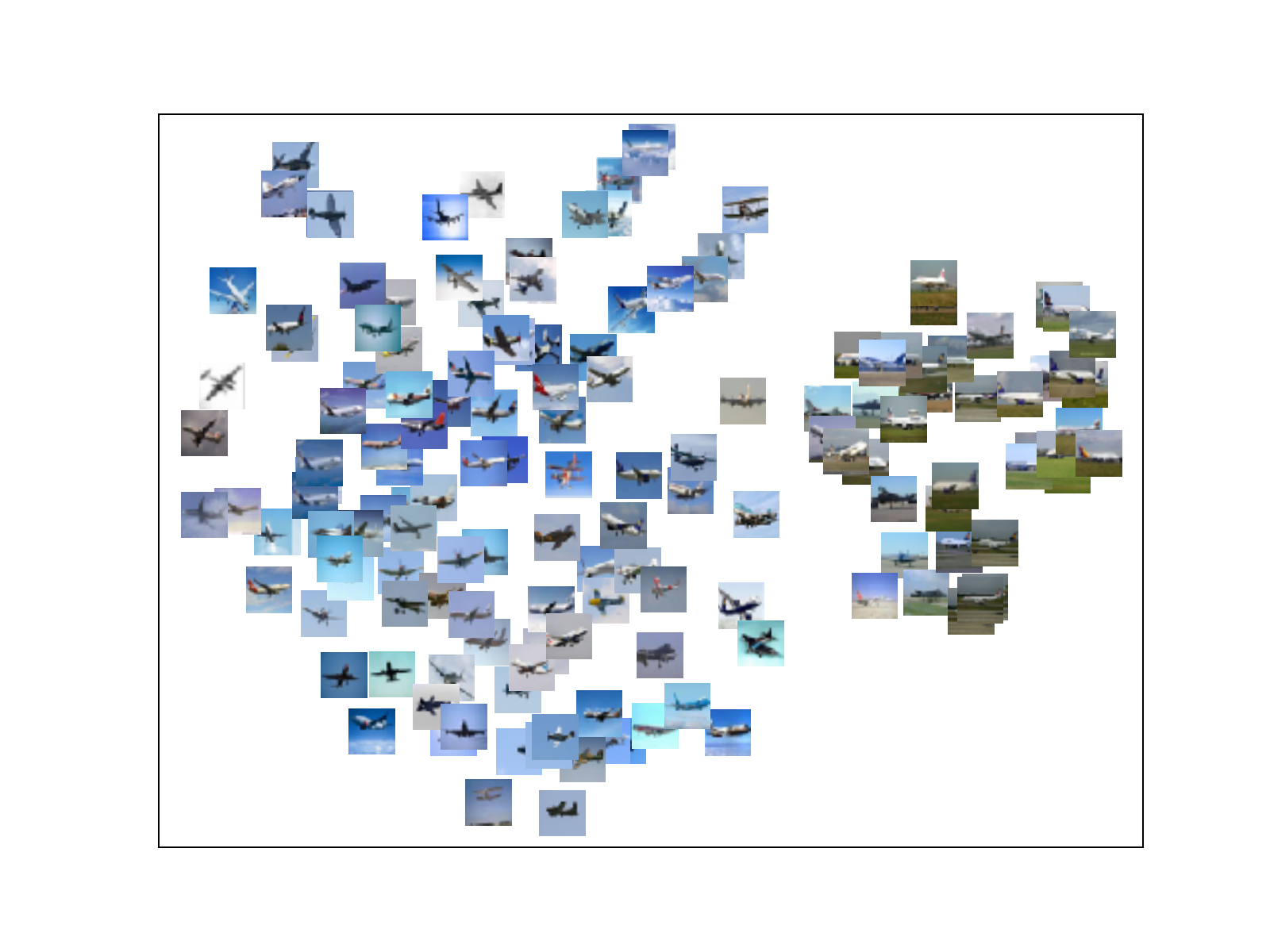}
    \caption{\textbf{Canonical prototypes} in the CIFAR-10 ``airplane'' class.\vspace*{-1ex}}
    \label{fig:clustersc}
  \end{subfigure}
  \caption{Our metrics' sets
    reveal interesting examples, which can be clustered.\vspace*{-2ex}}
\label{fig:clusters}
\end{figure*}

For the Fashion-MNIST ``shirt'' class,
this set---visually shown in Figure~\ref{fig:inline} on the right---includes not only the dress-looking example
but a number of other atypical ``shirt'' images,
including some looking like shorts.
Also apparent in the set are a number of
T-shirt-like and pullover-like images,
which are misleading,
given the other output classes of Fashion-MNIST.
For these sets,
which are likely to include
spurious, erroneously-labeled, and inherently ambiguous examples,
we use the name \emph{memorized exceptions} 
because 
they must be memorized as exceptions
for models to have been able to reach very high confidence
during training.
Similarly, Figure~\ref{fig:clustersa} shows a large (green) cluster 
of highly ambiguous boot-like sneakers, which appear indistinguishable 
from a cluster of memorized exceptions in the Fashion-MNIST ``ankle boot''
class (see Appendix~\ref{apx:protofigs}).

\textbf{Uncommon submodes:~}
On the other hand,
the \textsf{priv} metric is based on
differentially-private learning
which ensures
that no small group of examples
can possibly be memorized:
the privacy stems 
from adding noise and attenuating gradients
in a manner that will mask
the signal from rare examples 
during training.
This suggests
that we can find \emph{uncommon submodes} of the examples
in learning tasks
by intersecting the bottom-most outlier examples
on the \textsf{priv} metric
with the union of most well-represented
in the \textsf{boundary} and \textsf{ensemble} metrics.
Figure~\ref{fig:clustersb}
shows uncommon submodes discovered in MNIST
using the 25\% lowest outliers on \textsf{priv}
and top 50\% well-represented on other metrics.
Notably, all of the ``serif 1s'' in the entire
MNIST training set are found as a submode.

\textbf{Canonical prototypes:~}
Finally,
we can simply consider the intersection of
the sets of all the most represented examples
in all of our metrics.
The differences between our metrics should ensure
that this intersection is free of spurious or misleading
examples;
yet, our experiments
and human study suggest
the set will provide good coverage.
Hence, we call this set \emph{canonical prototypes}.
Figure~\ref{fig:clustersc} shows the airplanes that are canonical prototypes in CIFAR-10.

\begin{figure*}
  \begin{subfigure}{.31\textwidth}
    \includegraphics[width=\linewidth]{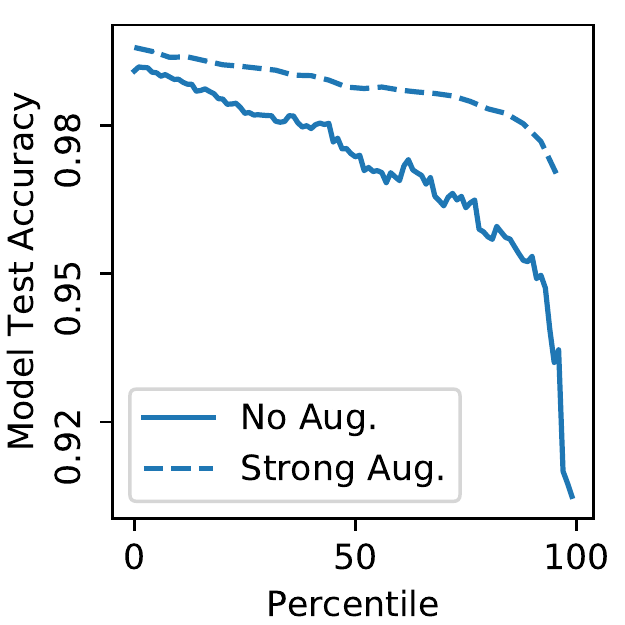}
    \caption{MNIST}
    \label{fig:prototraininga}
  \end{subfigure}
  \hspace*{\fill}
  \begin{subfigure}{.31\textwidth}
    \includegraphics[width=\linewidth]{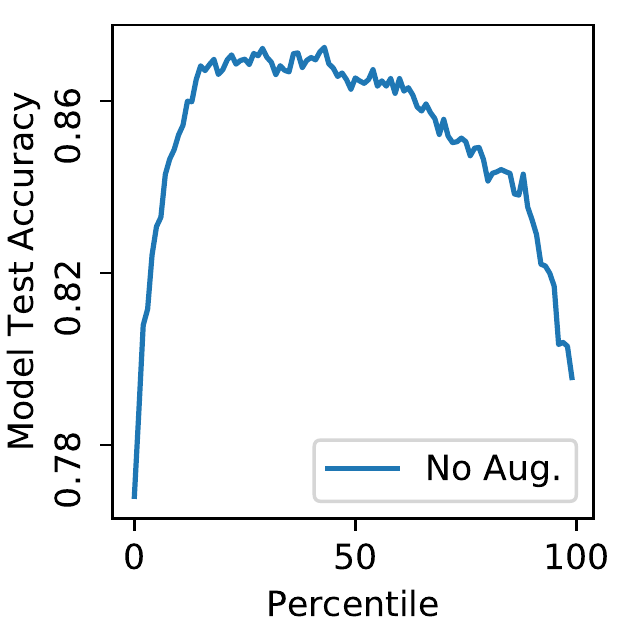}
    \caption{Fashion-MNIST}
    \label{fig:prototrainingb}
  \end{subfigure}
  \hspace*{\fill}
  \begin{subfigure}{.31\textwidth}
    \includegraphics[width=\linewidth]{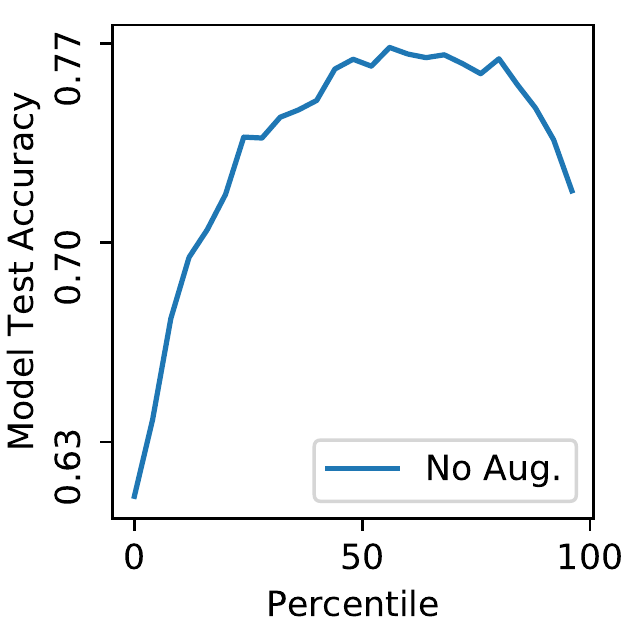}
    \caption{CIFAR-10}
    \label{fig:prototrainingc}
  \end{subfigure}

  \caption{Final test accuracy of a model after trained
    on $5,000$ training examples consecutively ranked by the
    \textsf{adv} metric (so that training on the least representative examples
    are that the 0th percentile, and the most representative at the 100th percentile).
    See text for full details.
    Given only $5,000$ training examples, on MNIST (subplot (a)) training on the outliers is always better,
    however for Fashion-MNIST (b) and CIFAR-10 (c) it is preferable to train using neither
    the most nor least well-represented examples, but those in the middle.
  }
\label{fig:prototraining}
  \begin{subfigure}{.31\textwidth}
    \includegraphics[width=\linewidth]{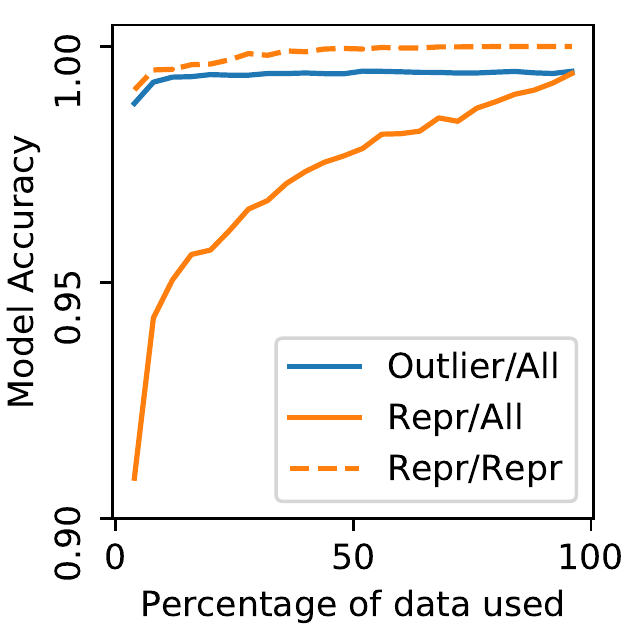}
    \caption{MNIST\vspace*{-1ex}}
  \end{subfigure}
  \hspace*{\fill}
  \begin{subfigure}{.31\textwidth}
    \includegraphics[width=\linewidth]{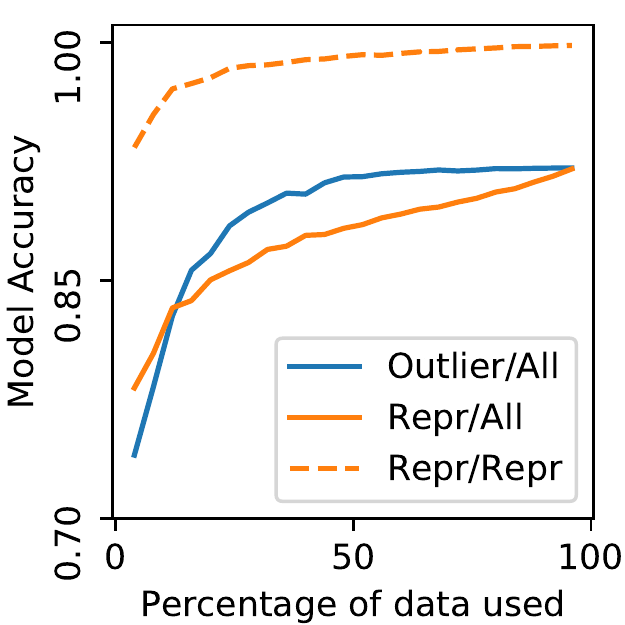}
    \caption{Fashion-MNIST\vspace*{-1ex}}
  \end{subfigure}
  \hspace*{\fill}
  \begin{subfigure}{.31\textwidth}
    \includegraphics[width=\linewidth]{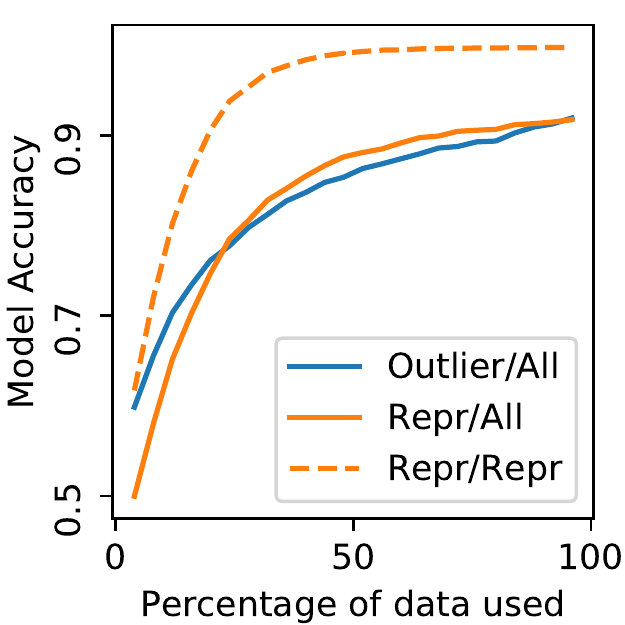}
    \caption{CIFAR-10\vspace*{-1ex}}
  \end{subfigure}
  \caption{Final test accuracy of a model after trained on
    a varrying percentage of training data as sorted by the \textsf{adv} metric.
    The blue solid lines correspond to final test accuracy when the training data consists of only
    the $x\%$ least well-represented examples, and growing to the most represented. The orange solid line
    corresponds to training on the $x\%$ most most well represented examples;
    the orange dashed line corresponds to when testing only on the top $50\%$ most well-represented
    test points.
    See text for full details.
}
\label{fig:prototraining2}
\end{figure*}

To further aid interpretability in Figures~\ref{fig:inline} and~\ref{fig:clusters},
we perform
a combination of dimensionality reduction and clustering.
We apply t-SNE~\citep{maaten2008visualizing}
on the pixel space (for MNIST and Fashion-MNIST) or ResNetv2 feature space (for CIFAR10)
to project the example sets into two dimensions,
and cluster with HDBSCAN~\citep{campello2013density}, a hierarchical and
density-based clustering algorithm
which does not try to assign all points to clusters---which 
not only can improve clusters
but also identify spurious data.
We believe that
other types of
data projection and
clustering
could also be usefully
applied to our metrics,
and offer significant insight into ML datasets.
(See Appendix~\ref{apx:clustering}
for this section's figures shown larger.)

\section{Utilizing Well-Represented Examples}
\label{sec:using}
Our metrics enable us to inspect 
datasets at scale and identify examples in the training and test
sets that are of particular importance to evaluating a model's performance.
For instance, imagine we observe that a non-private model performs better than a private model
because the private model is unable to classify uncommon submodes 
correctly. 
This is desirable because it is harder to protect the privacy of
examples that are not well-represented in the training data. 
However, if one were to limit their evaluation to simply
reporting accuracy, we would conclude that the privacy-preserving
model performs ``worse'' while this is not necessarily the case.

Beyond such applications that provide insights 
into learning and inference,
we now show that our metrics for sorting examples
 according to how well-represented
they are can also be integrated directly in learning procedures
to improve them.
Namely, we look at three model properties: sample complexity, 
accuracy, or robustness.

\subsection{Curriculum Learning}

We perform two experiments on the three datasets
to investigate whether it is better to train on the the well-represented
examples or the outliers---exploring
the
``train on hard data'' vs.\ ``train on easy data''
question of curriculum learning~\citep{ren2018learning}.
To begin, we order all training data according to 
our~\textsf{adv} metric.\footnote{We use the \textsf{adv} metric since
  it is well-correlated to human perception and 
  does not involve model performance in defining it.}

\textbf{Experiment 1.}
First, we experiment with training on splits of $5,000$ training examples
(approximately $10\%$ of the training data) chosen by taking the $k$-th most
well-represented example to the ($k$+5000)-th most well-represented,
for different values of $k$.
As shown in Figure~\ref{fig:prototraining},
we find that the index $k$ that yields the most accurate model varies
substantially across the datasets and tasks.
(In the plot, the x-axis is given in a percentile from $0$ to $100$, obtained
by dividing $k$ by the size of the dataset.)
For example, initially we train a model on \emph{only} the $5,000$ least
represented examples and record the
models's final test accuracy; for MNIST this accuracy is nearly $99\%$ already.
However, for CIFAR-10 it is preferable to take $5,000$ examples
starting at the 60th percentile---that is, examples ordered from $30,000$
to $35,000$ by the \textsf{adv} metric reach a test accuracy of $77\%$.

To summarize the results,
on MNIST, training on the outlier examples
gives the highest accuracy;
conversely, on Fashion-MNIST and CIFAR-10, training on examples that are better
represented
gives the highest accuracy.
We conjecture this is due to the dataset complexity: because nearly all of MNIST
is very easy, it makes sense to train on the hardest,
most outlier examples.
However, because Fashion-MNIST and CIFAR-10
CIFAR-10 are comparably difficult, training on the most well-represented examples is
better given a limited amount of training data: it is simply too hard
to learn on \emph{only} the least representative training examples.

Notably, many of the CIFAR-10 and Fashion-MNIST outliers
appear to be inherently misleading or ambiguous examples,
and several are simply erroneously labeled.
We find that about
$10\%$ of the first 5,000 outliers meet
our definition
of memorized exceptions.
Also, we find that inserting 10\% label noise
causes model accuracy to decrease by about 10\%,
regardless of the split trained on---i.e.,
that
to achieve high accuracy on small training data
erroneous and misleading
outliers 
must be removed---and explaining
the low accuracy
shown on the left
in the graph of Figures~\ref{fig:prototrainingb} and ~\ref{fig:prototrainingc}.

\textbf{Experiment 2.}
For our second experiment, we ask: is it better to train on the $k$-most or $k$-least
well-represented examples?
That is, the prior experiment assumed the amount of data is fixed, and we must choose
which percentile of data to use.
Now, we examine what the best strategy is to apply if we must choose either a prefix
or a suffix of the training data as ordered by our \textsf{adv} metric.

The results are given in Figure~\ref{fig:prototraining2}.
Again, we find the answer depends on the dataset.
On MNIST, training on the $k$-least represented examples is always better for any $k$
than training on the $k$-most represented examples.
However, on Fashion-MNIST and CIFAR-10, training on the well-represented examples
is better when $k$ is small, but as soon as we begin to collect more than
roughly $10,000$ examples for Fashion-MNIST or $20,000$ for CIFAR-10, training
on the outliers begins to give more accurate models.
However, we find that training only on the most well-represented examples
found in the training data gives extremely high test accuracy on the
well-represented  examples found in the test data.

This evidence supports our hypothesis that training on difficult, but not impossibly
difficult, training data is of most value.
The harder the task, the more useful well-represented training examples are.

Also shown in Figure~\ref{fig:prototraining2} is the final test accuracy of a model when only \emph{evaluated}
on the well-represented test examples. Here, we find that the test accuracy is subsantially
higher.

\vspace*{-0.05in}

\subsection{Decision Boundary Analysis}

\vspace*{-0.05in}
While training exclusively on the well-represented examples
often gives inferior accuracy compared
to training on the outliers, the former has the benefit of 
obtaining models with simpler decision boundaries.
Thus, it is natural to ask whether training on well represented examples
gives models that are
more robust to adversarial examples \citep{szegedy2013intriguing}.
In fact, prior work has found that discarding outliers from the training
data can help with both classifying and detecting adversarial examples~\citep{liu2018less}.

To show that simpler boundaries can lead to somewhat higher robustness,
we train models on fixed-sized subsets of the data
where include $5,000$ training points either from the well-represented
training examples or those that are not.
For each model, we then compute the mean $\ell_\infty$ adversarial distance needed to find adversarial
examples.
As shown in Figure~\ref{fig:protoadv} in the Appendix,
the Fashion-MNIST and CIFAR-10 models that are trained on well well represented
examples are \emph{more} robust to adversarial examples than those trained
on a slice of training data that is mostly made up of outliers.
However, these models trained on a slice of $5,000$ well-represented
examples remain comparably robust to a baseline model trained
on the entire data.

\section{Related Work}
\label{apx:related-work}

\paragraph{Prototypes.}
At least since the work of~\citet{Zhang:Protos}
which was based on intra- and inter-concept similarity,
prototypes have been examined
using several metrics
derived from the intuitive notion
that one could find
``quintessential observations that best represent clusters in a dataset''~\citep{kim2014bayesian}.
Several more formal variants
of this definition were proposed in the literature---along 
with corresponding techniques for finding prototypes.~\citet{kim2016examples} select prototypes
according to their maximum mean discrepancy with the data, which assumes the existence
of an appropriate kernel for the data of interest.~\citet{li2017deep} circumvent this
limitation by prepending classifiers
with an autoencoder projecting the input data on a manifold of reduced dimensionality.
A prototype layer, which  serves as the classifier's input, 
is then trained to minimize the distance between inputs and
a set of prototypes on this manifold. 
While this method
improves interpretability by ensuring that prototypes are central to the classifier's
logic, it does require that one modify the model's architecture. Instead, metrics considered
in our manuscript all operate on existing architectures.
\citet{stock2017convnets} proposed to use distance to the boundary---approximately
measured with an adversarial example algorithm---as a proxy for prototypicality.

\paragraph{Other interpretability approaches.} 
Prototypes enable interpretability
because they provide a subset of examples that summarize
the original dataset and best explain a particular
decision made at test time~\citep{bien2011prototype}. 
Other approaches like saliency maps instead synthesize
new inputs to visualize what a neural network has learned. This is typically done
by gradient descent with respect to the input space~\citep{zeiler2014visualizing,simonyan2013deep}.
Because they rely on model gradients, saliency maps can be fragile and only
locally applicable~\citep{fong2017interpretable}.

Beyond interpretability, prototypes are also motivated by additional use cases, some of
which we discussed in Section~\ref{sec:using}. Next, we review related
work in two of these applications: namely, curriculum learning and reducing sample complexity.

\paragraph{Curriculum learning.} Based on the observation that the order
in which training data is presented to the model can improve  performance (e.g., convergence) of optimization during
learning and circumvent limitations of the dataset (e.g., data imbalance or noisy labels), 
curriculum learning seeks to find the best order in which to 
analyze training data~\citep{bengio2009curriculum}. 
This first effort further hypothesizes that easy-to-classify samples should be 
presented early in training while complex samples gradually inserted as learning progresses.
While~\citet{bengio2009curriculum} assumed the existence of hard-coded 
curriculum labels in the dataset, \citet{chinleverage} 
sample an order for the training set by assigning each
point a sampling probability proportional to its leverage score---the
distance between the point and a linear model fitted to the whole data. Instead, we use
metrics that also apply to data that cannot be modeled linearly.

The curriculum may also be generated online during training,
so as to take into account progress made by the learner~\citep{kumar2010self}.  %
For instance,
\citet{katharopoulos2017biased} train an auxiliary LSTM model to predict the loss
of training samples, which they use to sample a subset of training points
analyzed by the learner at each training iteration.
Similarly,~\citep{jiang2017mentornet} have an auxiliary model predict the curriculum.
This auxiliary model is trained using the learner's current feature representation of a smaller
holdout set of data for which ground-truth curriculum is known. 

However, as reported in our experiments, training on easy samples is beneficial
when the dataset is noisy, whereas training on hard examples is on the contrary
more effective when data is clean. These observations
oppose self-paced learning~\citep{kumar2010self} with hard example mining~\citep{shrivastava2016training}.
Several strategies have been proposed to perform better in both settings.
Assuming the existence of a holdout set as well,~\citet{ren2018learning} assign
a weight to each training example that characterizes the alignment
of both the logits and gradients of the learner on training and heldout data.
\citet{chang2017active} propose to train on points with high prediction variance
or whose average prediction is close from the decision threshold.
Both the variance and average are estimated by analyzing a sliding 
window of the history of prediction probabilities throughout training
epochs.

\paragraph{Sample complexity.} Prototypes of a given
task share some intuition with the notion of coresets~\citep{agarwal2005geometric, huggins2016coresets, bachem2017practical, tolochinsky2018coresets} because both prototypes
and coresets describe the dataset
in a more compact way---by returning a (potentially weighted) subset of the original dataset.
For instance, clustering algorithms may rely on both prototypes~\citep{biehl2016prototype}  or coresets~\citep{biehl2016prototype} 
to cope with the high dimensionality of a task. However, prototypes and 
coresets differ in essential ways. In particular, coresets
are defined according to a metric of interest (e.g., the loss that one
would like to minimize during training) whereas prototypes are 
independent of any machine-learning aspects as indicated in our list of
desirable properties for prototypicality metrics from Section~\ref{sec:desire}. 

Taking a different approach,
\citet{wang2018data} apply influence 
functions~\citep{koh2017understanding} to discard training inputs that do not
affect learning.
Conversely, for MNIST, we found in our experiments that removing
individual training examples did not have a measurable impact on the
predictions of individual test examples. Specifically, we trained many
models to 100\% training accuracy where we left one training example
out for each model. There was no statistically significant difference
between the models predictions on each individual test example

\section{Conclusion}

This paper explores metrics for gaining insight into the properties of
datasets commonly used for training deep learning models.
We develop five metrics %
and find that humans agree that the rankings computed
capture human intuition behind
what is meant by a good representative example of the class.

When the metrics disagree on how well-represented an example is, we
can often learn something interesting about that example.
This helps forming an understanding of the performance
of ML models that goes beyond measuring
test accuracy.
For instance, by identifying memorized exceptions in the test data, we
may not weight mistakes that models make on these points as important as mistakes on canonical
prototypes.
Further, by identifying uncommon submodes we can learn where
collecting training points will be useful.
We find
that models trained on well-represented examples often
have simpler decision boundaries
and are thus slightly more adversarially robust,
however training
only on the most represented often yields inferior accuracy compared to
training on outliers.

We believe that  exploring other metrics for assessing properties
of datasets
and developing methods for using them during training is an important
area of future work, and hope that our analysis will be useful
towards that end goal.

\bibliography{iclr2019_conference}
\bibliographystyle{iclr2019_conference}

\clearpage
\onecolumn

\section*{Appendix}

\newpage
\section{Figures of outliers}
\label{apx:protofigs}

The following are training examples from MNIST, FashionMNIST, 
and CIFAR10 that are identified as most
outlier (left of the red bar) or prototypical (right of the green bar). Images are
presented in groups by class. Each row in these groups
corresponds to one of the five metrics in Section~\ref{ssec:metrics}.

\vspace*{-0.1in}

\subsection{MNIST extreme outliers}

All MNIST results were obtained with a CNN 
made up of two convolutional layers (each with kernel size of 5x5 and
followed by a 2x2 max-pooling layer) and
a fully-connected layer of 256 units. It was trained
with Adam at a learning rate of $10^{-3}$ with a $10^{-3}$ decay.
When an ensemble of models was needed
(e.g., for the~\textsf{agr} metric),
these were obtained by using different random initializations.

\insertOutliersToPrototypesImage{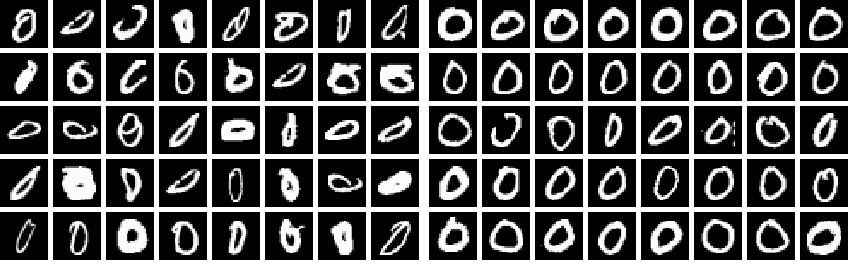}{0.06}{0}

\insertOutliersToPrototypesImage{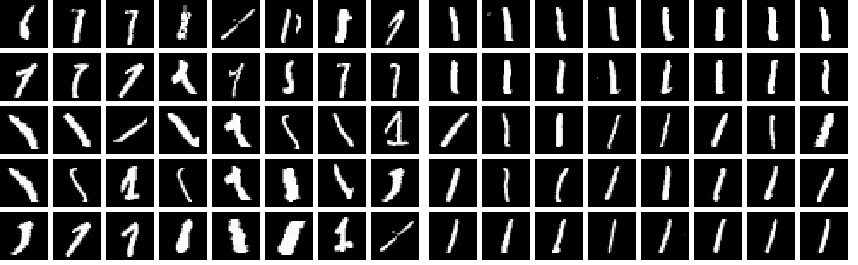}{0.06}{0}

\insertOutliersToPrototypesImage{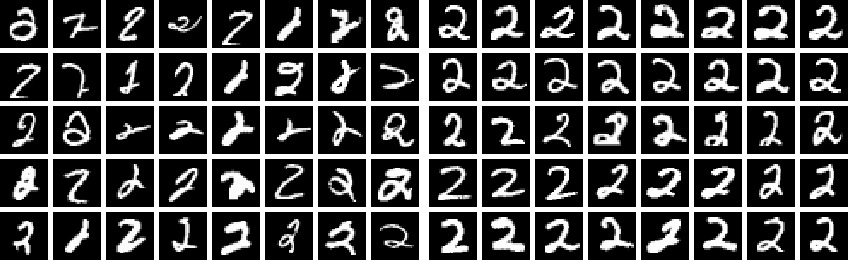}{0.06}{0}

\insertOutliersToPrototypesImage{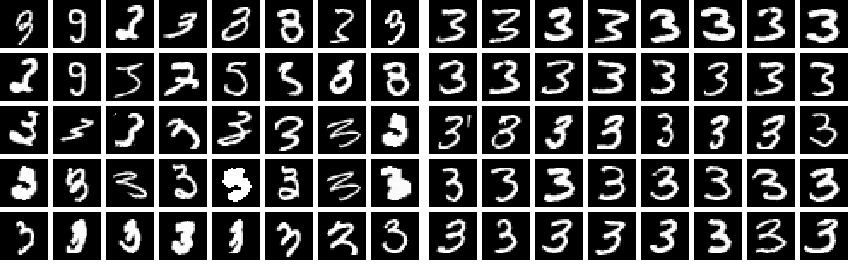}{0.06}{0}

\insertOutliersToPrototypesImage{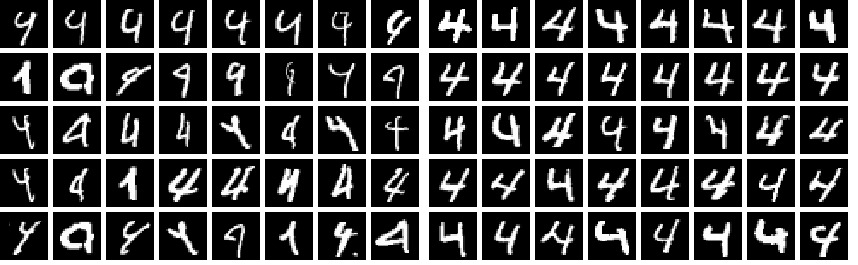}{0.06}{0}

\insertOutliersToPrototypesImage{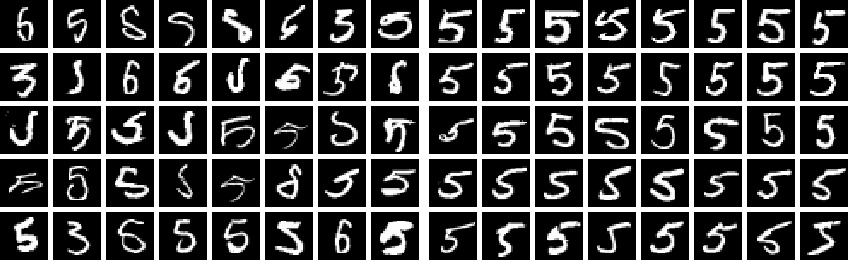}{0.06}{0}

\insertOutliersToPrototypesImage{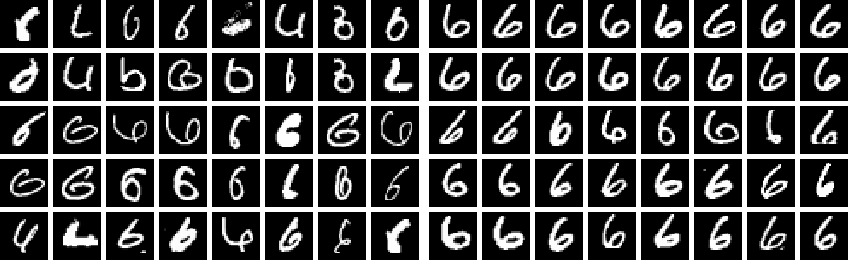}{0.06}{0}

\insertOutliersToPrototypesImage{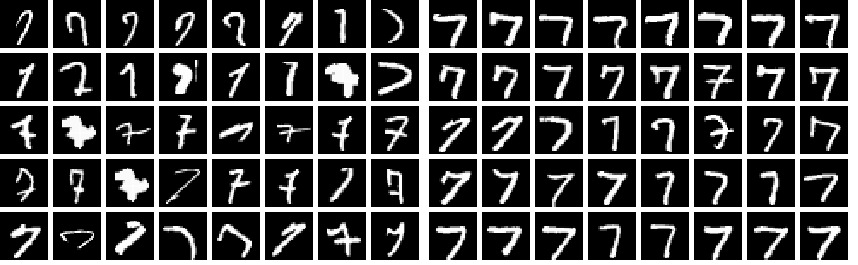}{0.06}{0}

\insertOutliersToPrototypesImage{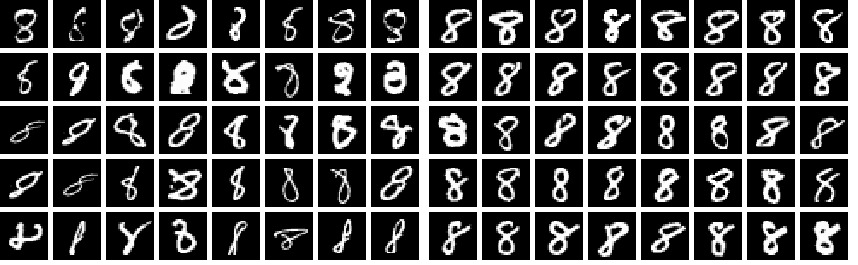}{0.06}{0}

\insertOutliersToPrototypesImage{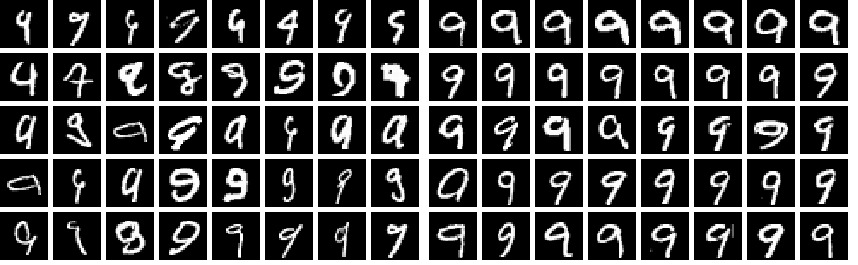}{0.06}{0}

\subsection{Fashion-MNIST extreme outliers}

The Fashion-MNIST model architecture is identical to the one used for MNIST.
It was also trained with the same optimizer and hyper-parameters.

\insertOutliersToPrototypesImage{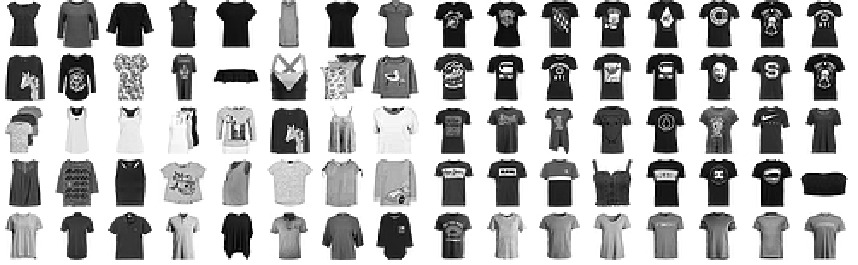}{0}{0}

\insertOutliersToPrototypesImage{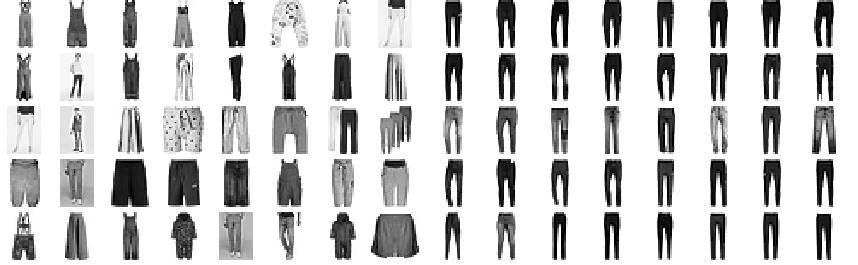}{0}{0}

\insertOutliersToPrototypesImage{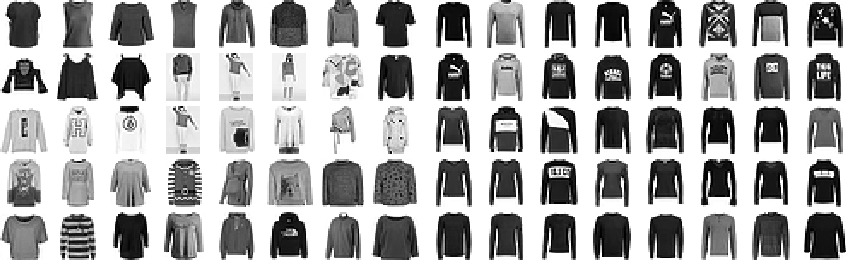}{0}{0}

\insertOutliersToPrototypesImage{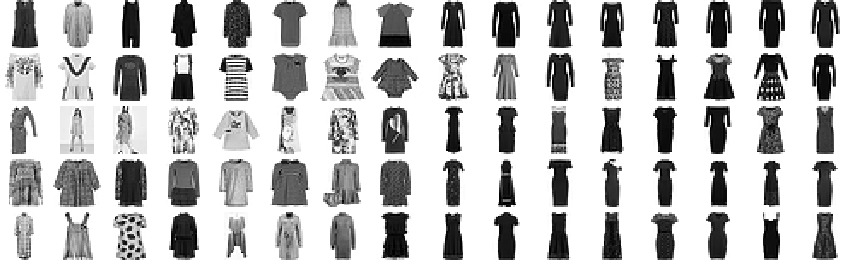}{0}{0}

\insertOutliersToPrototypesImage{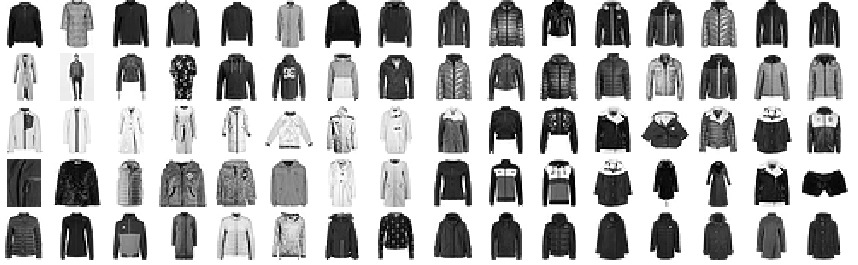}{0.02}{0}

\insertOutliersToPrototypesImage{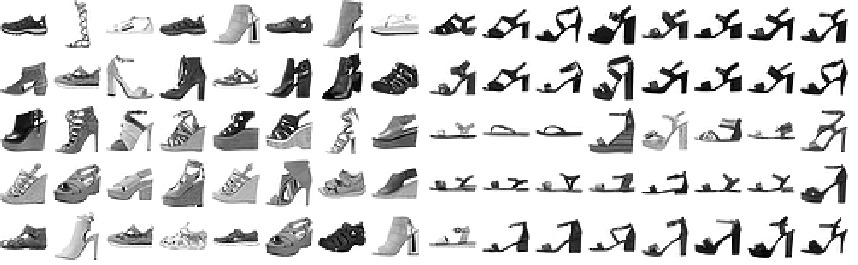}{0.03}{0}

\insertOutliersToPrototypesImage{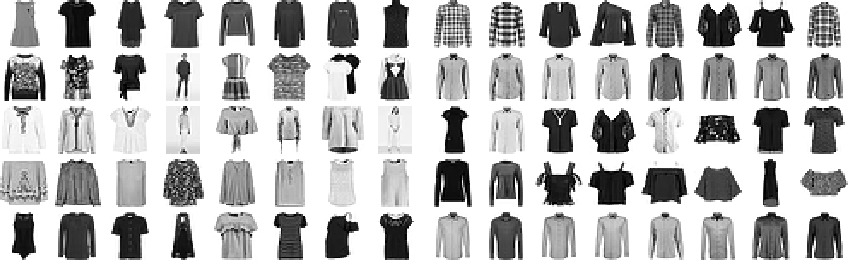}{0.04}{0}

\insertOutliersToPrototypesImage{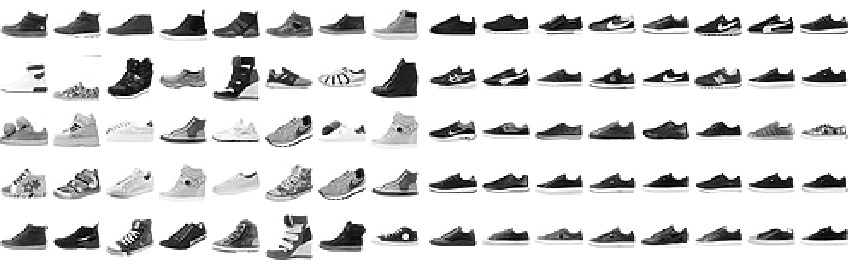}{0.05}{0}

\insertOutliersToPrototypesImage{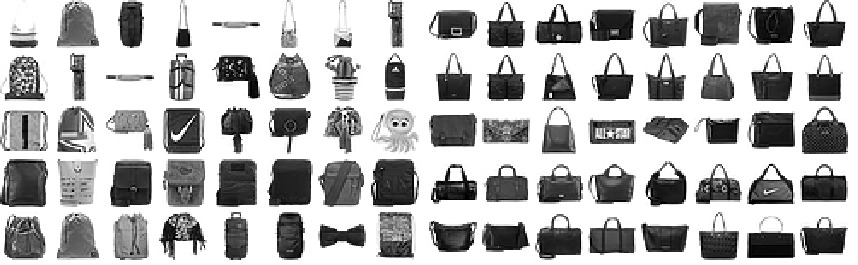}{0.04}{0}

\insertOutliersToPrototypesImage{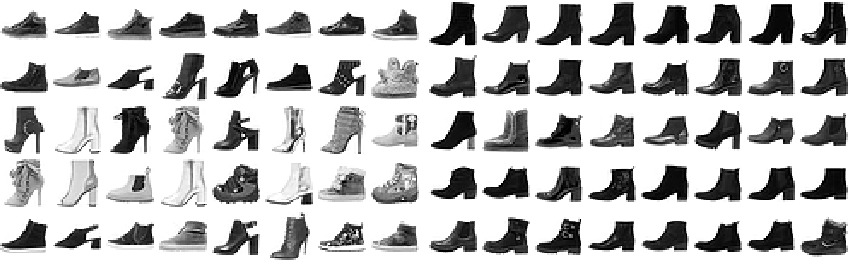}{0.05}{0}

\subsection{CIFAR extreme outliers}

All CIFAR results were obtained with a ResNetv2 trained on batches of $32$ points
with the Adam optimizer for $100$ epochs at an initial learning rate of $10^{-3}$ 
decayed down to $10^{-4}$ after 80 epochs. We adapted the following data augmentation
and training
script: \url{https://raw.githubusercontent.com/keras-team/keras/master/examples/cifar10_resnet.py}
When an ensemble of models was needed
(e.g., for the~\textsf{agr} metric),
these were obtained by using different random initializations.

\insertOutliersToPrototypesImage{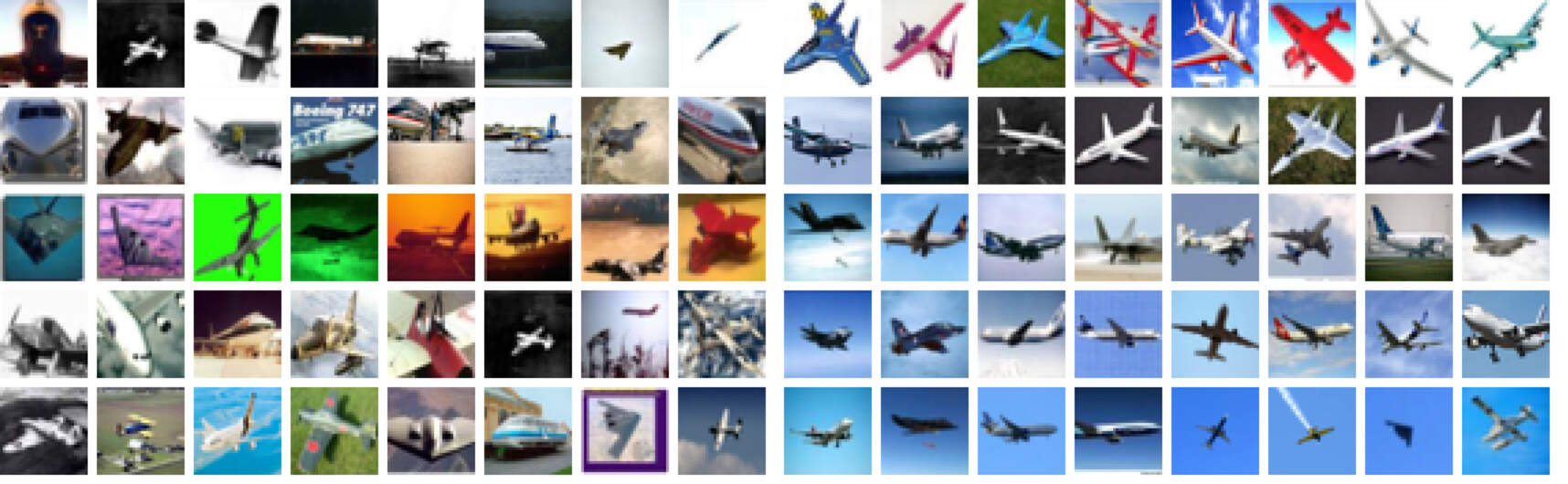}{0.06}{0}

\insertOutliersToPrototypesImage{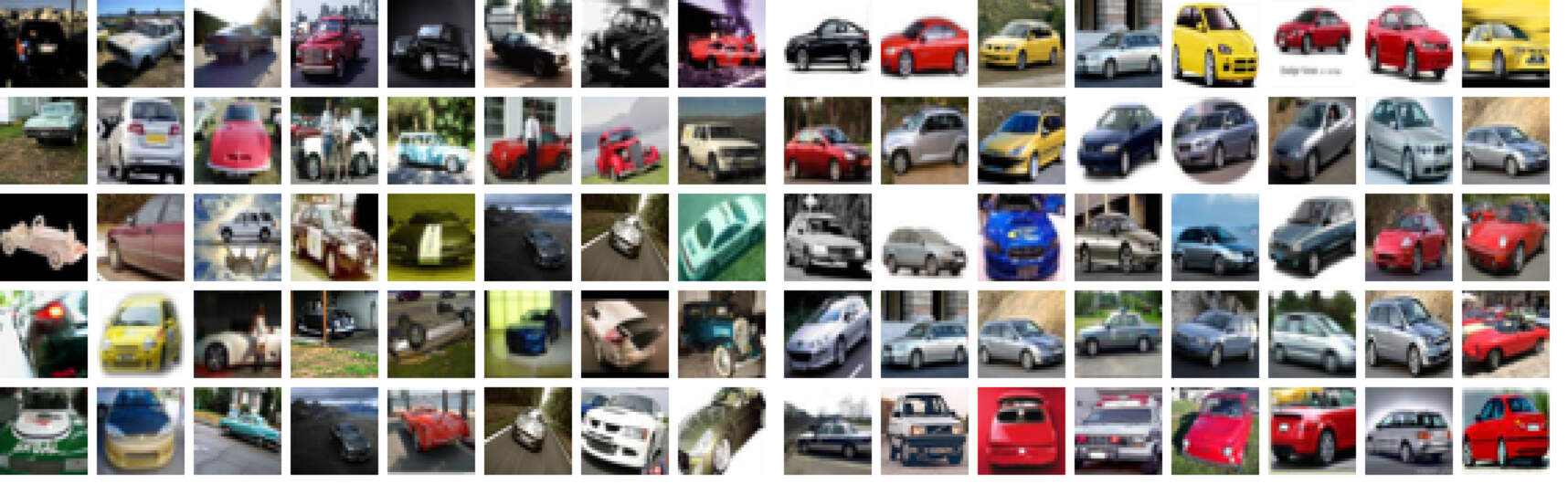}{0.06}{0}

\insertOutliersToPrototypesImage{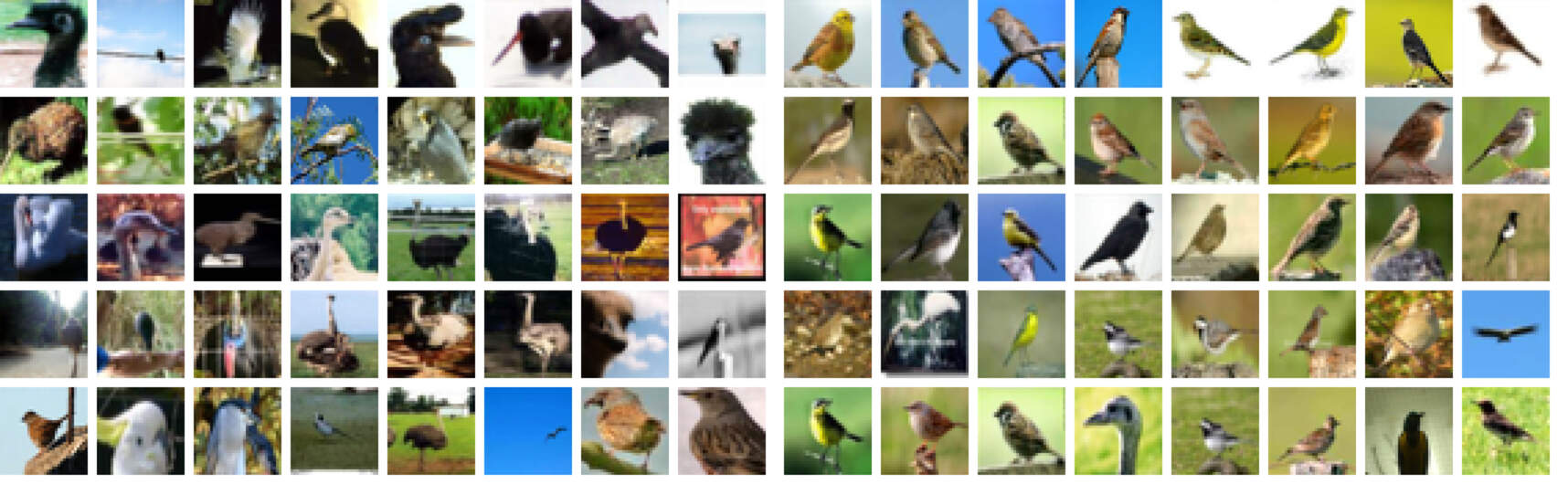}{0.06}{0}

\insertOutliersToPrototypesImage{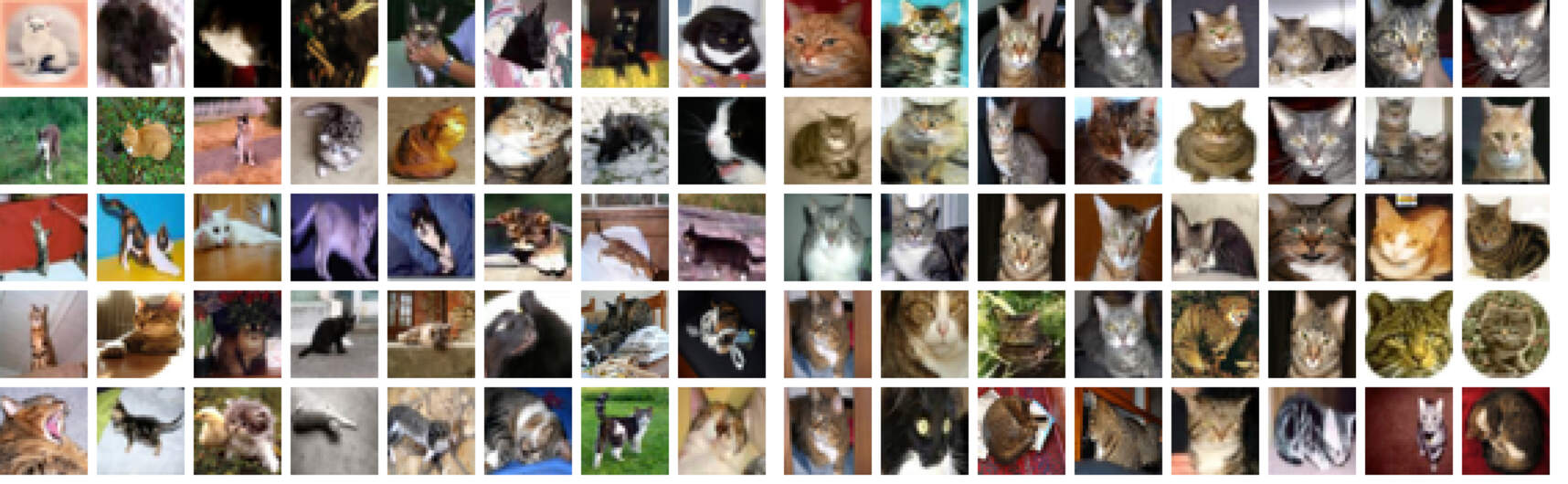}{0.06}{0}

\insertOutliersToPrototypesImage{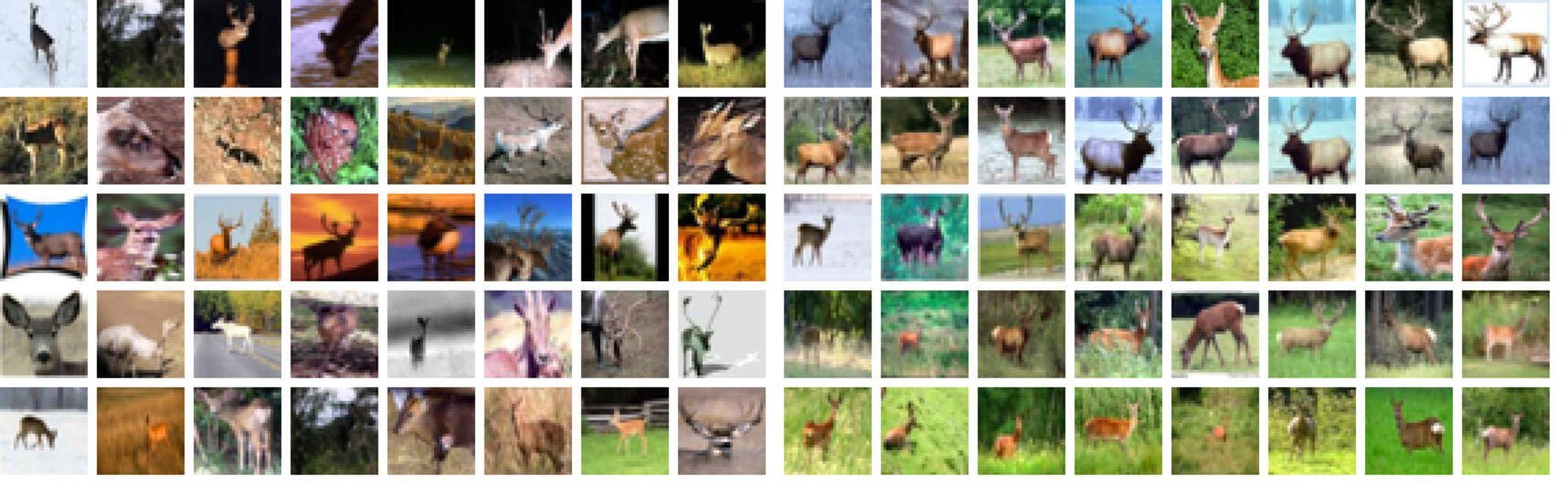}{0.06}{0}

\insertOutliersToPrototypesImage{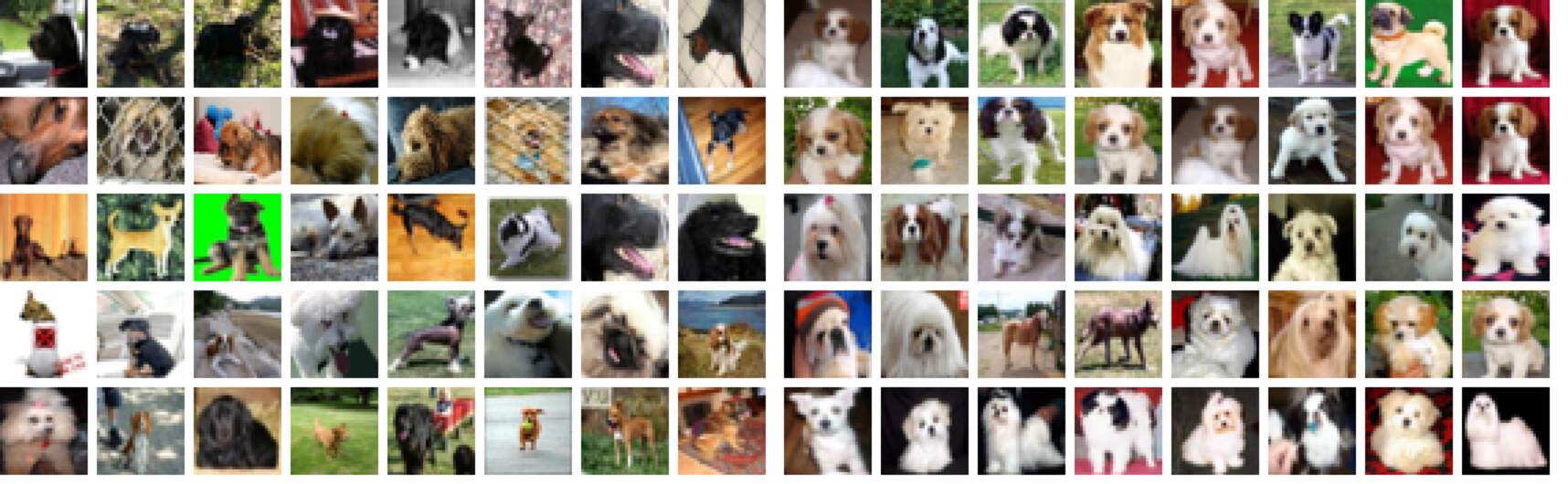}{0.06}{0}

\insertOutliersToPrototypesImage{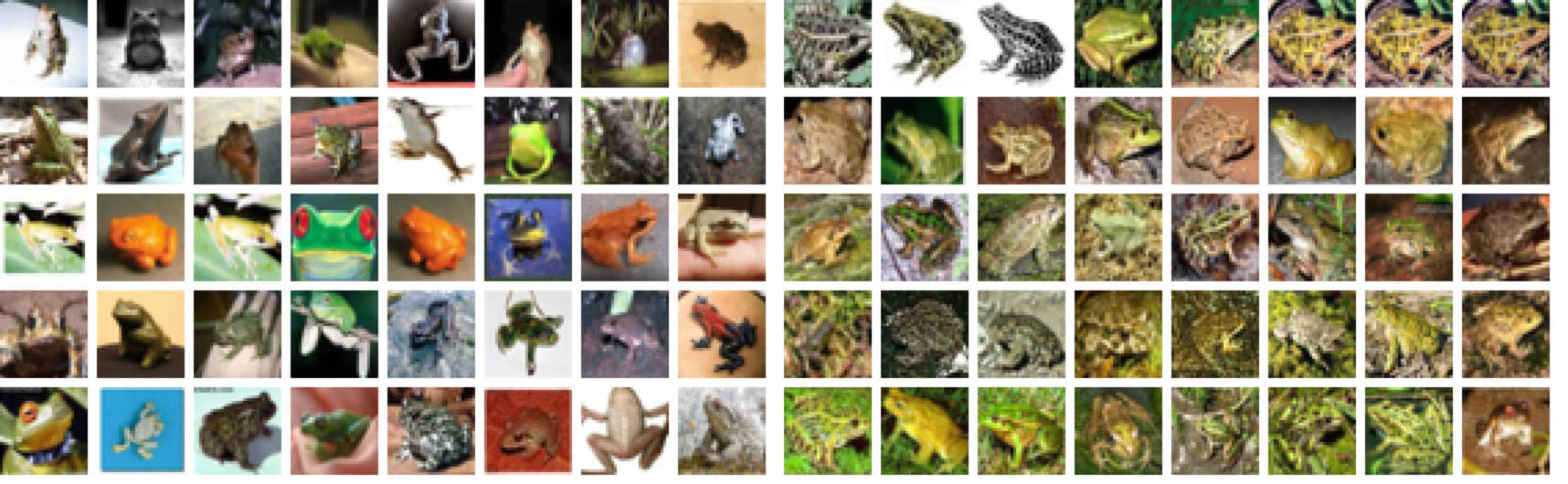}{0.06}{0}

\insertOutliersToPrototypesImage{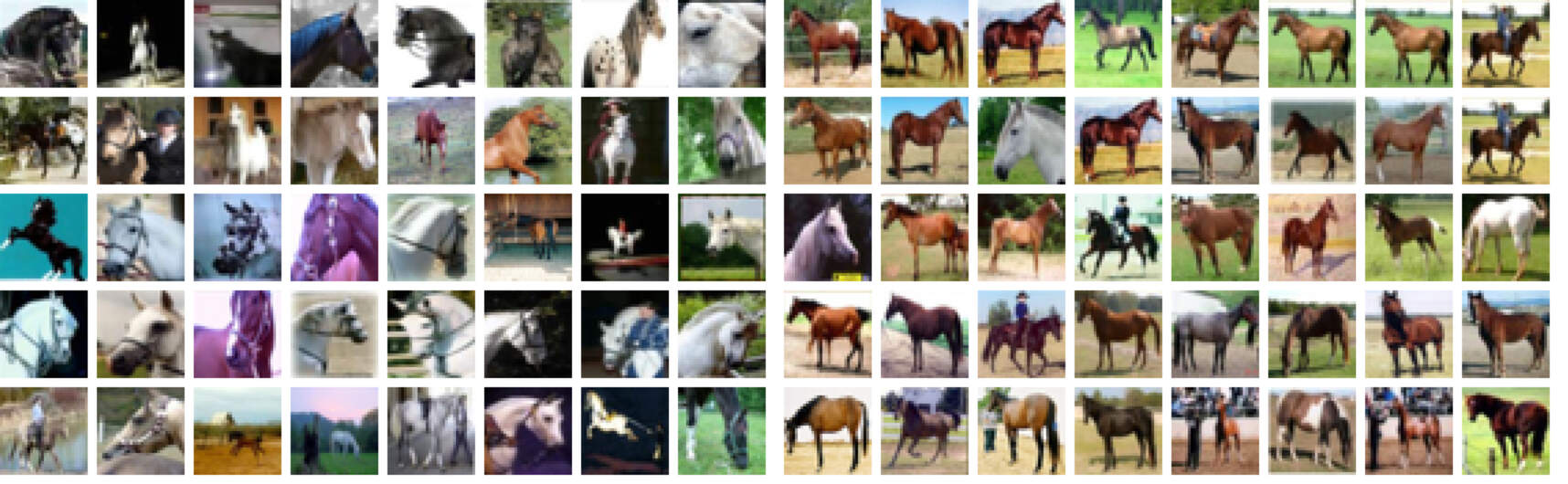}{0.06}{0}

\insertOutliersToPrototypesImage{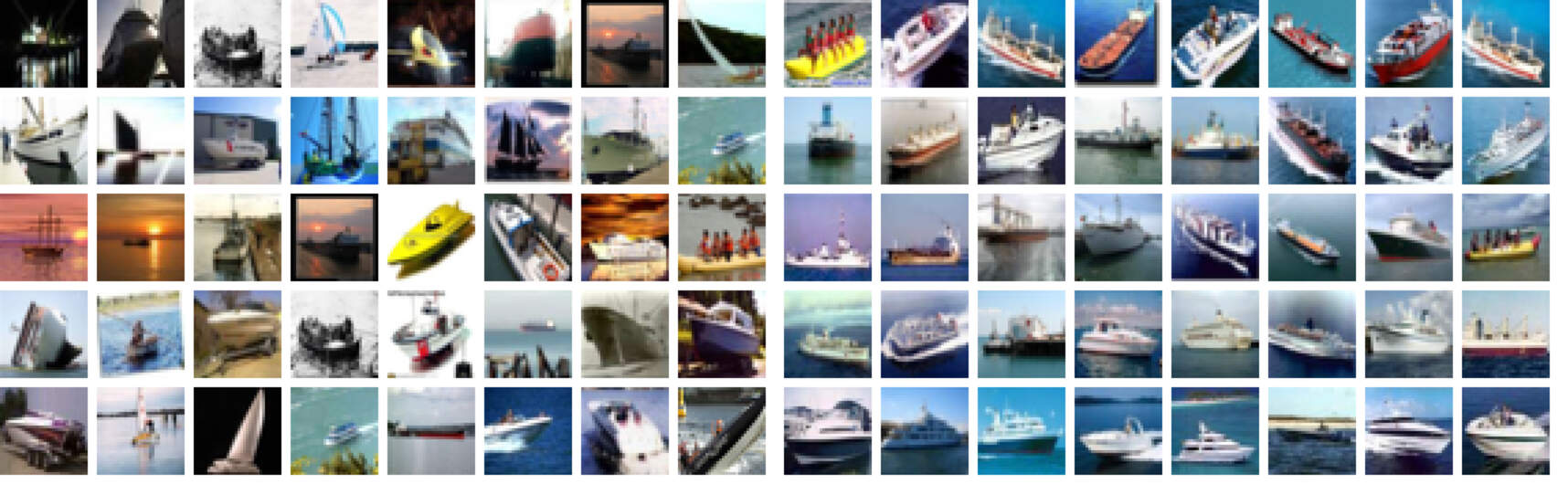}{0.06}{0}

\insertOutliersToPrototypesImage{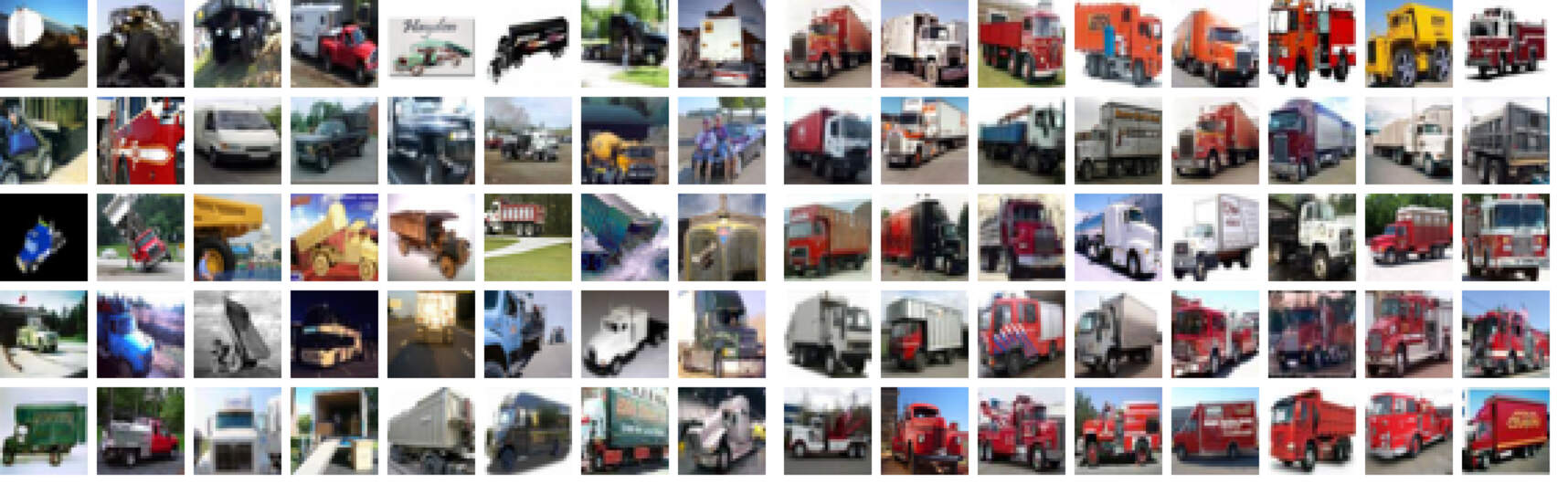}{0.06}{0}

\clearpage
\subsection{ImageNet extreme outliers}

The following pre-trained ImageNet models were used: 
DenseNet121, DenseNet169, DenseNet201
InceptionV3, InceptionResNetV2,
Large NASNet, Mobile NASNet,
 ResNet50, VGG16,
VGG19, and Xception. They are all found in the Keras 
library: \url{https://keras.io/applications}.

\insertOutliersToPrototypesImageImagenet{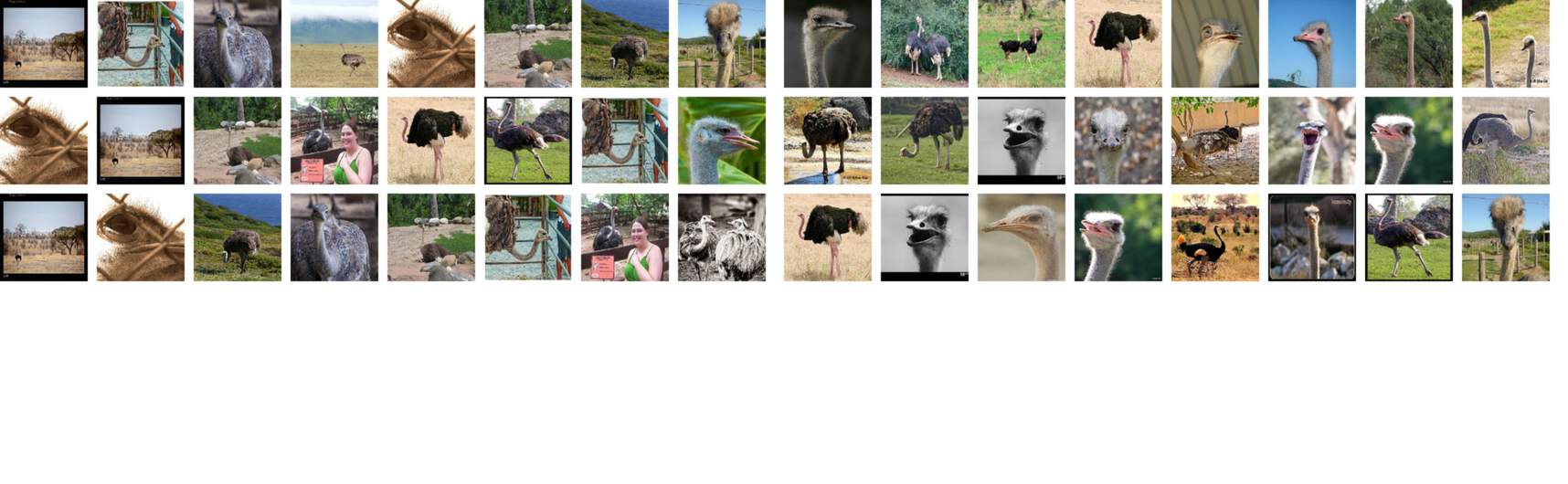}{0.05}{0}

\insertOutliersToPrototypesImageImagenet{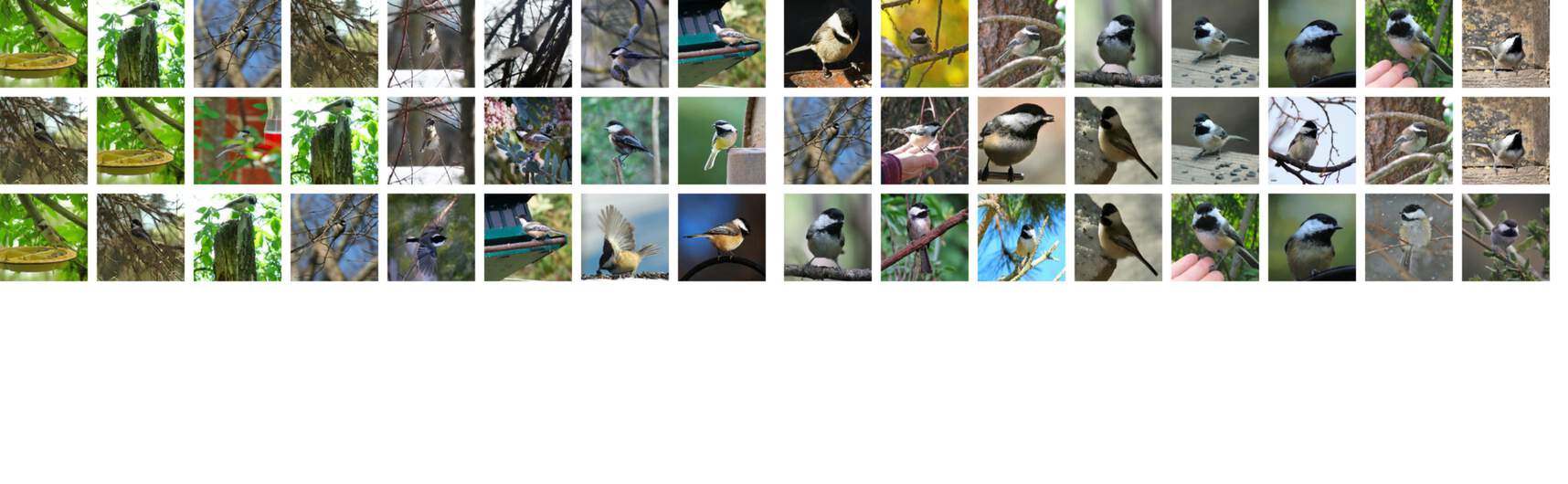}{0.05}{0}

\insertOutliersToPrototypesImageImagenet{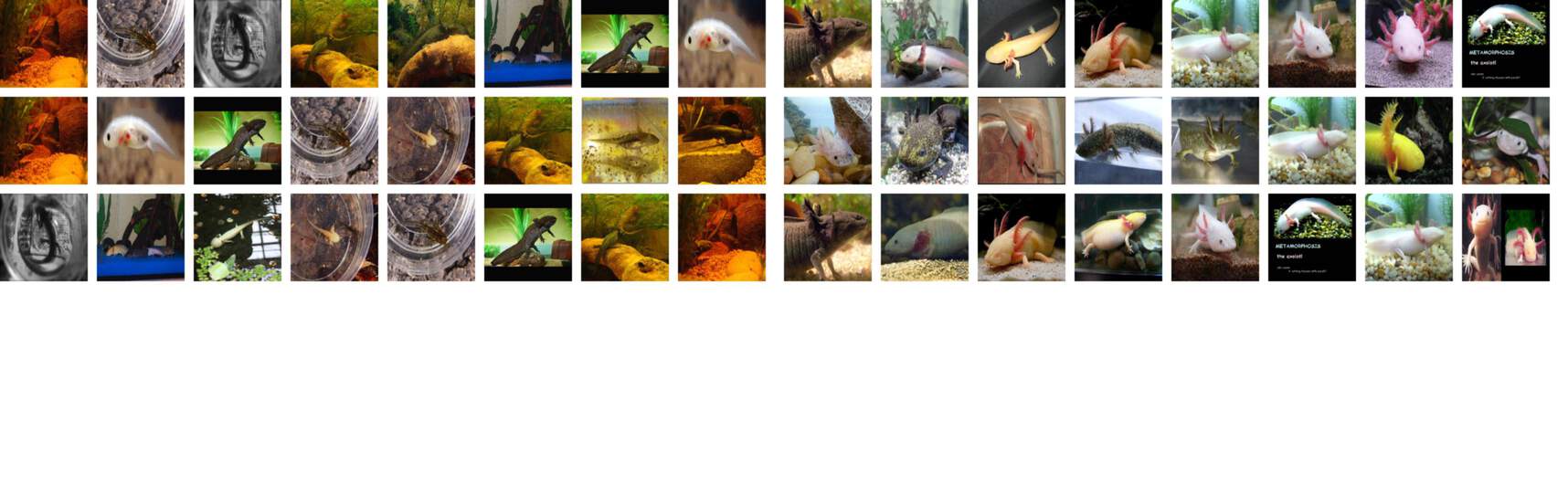}{0.05}{0}

\insertOutliersToPrototypesImageImagenet{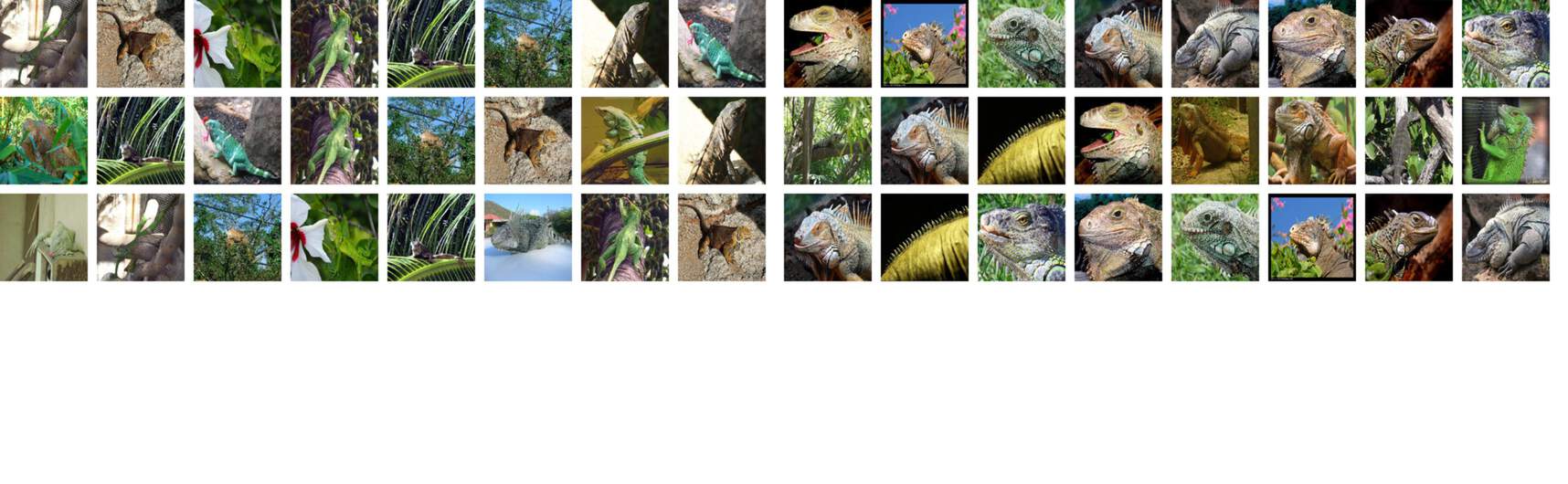}{0.05}{0}

\insertOutliersToPrototypesImageImagenet{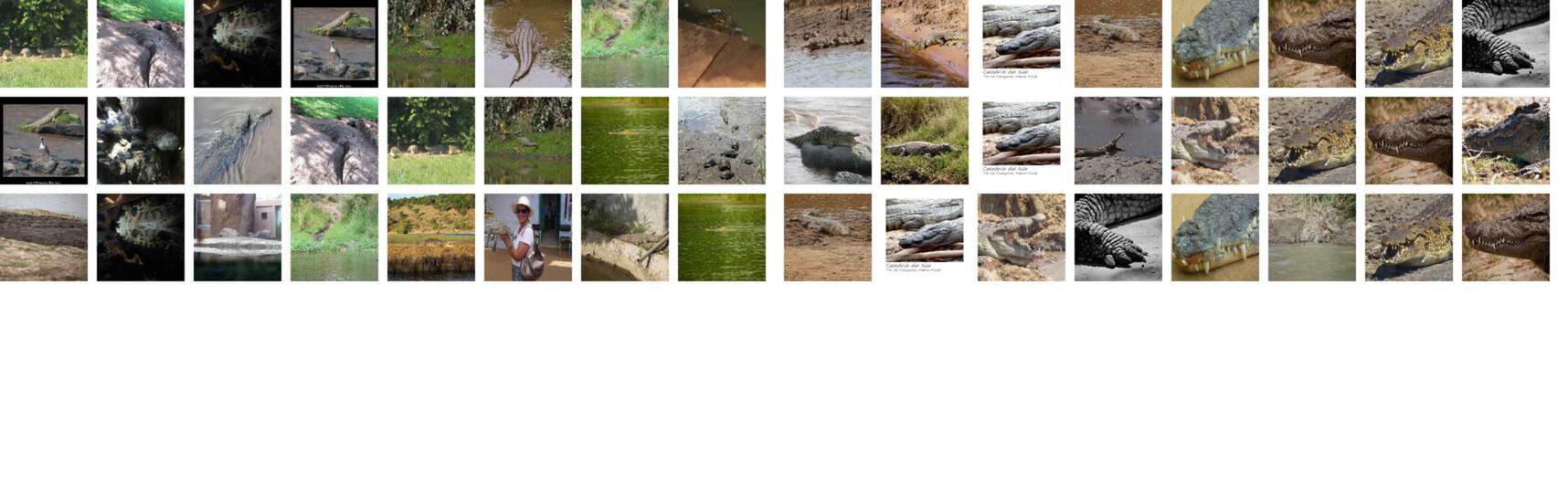}{0.05}{0}

\insertOutliersToPrototypesImageImagenet{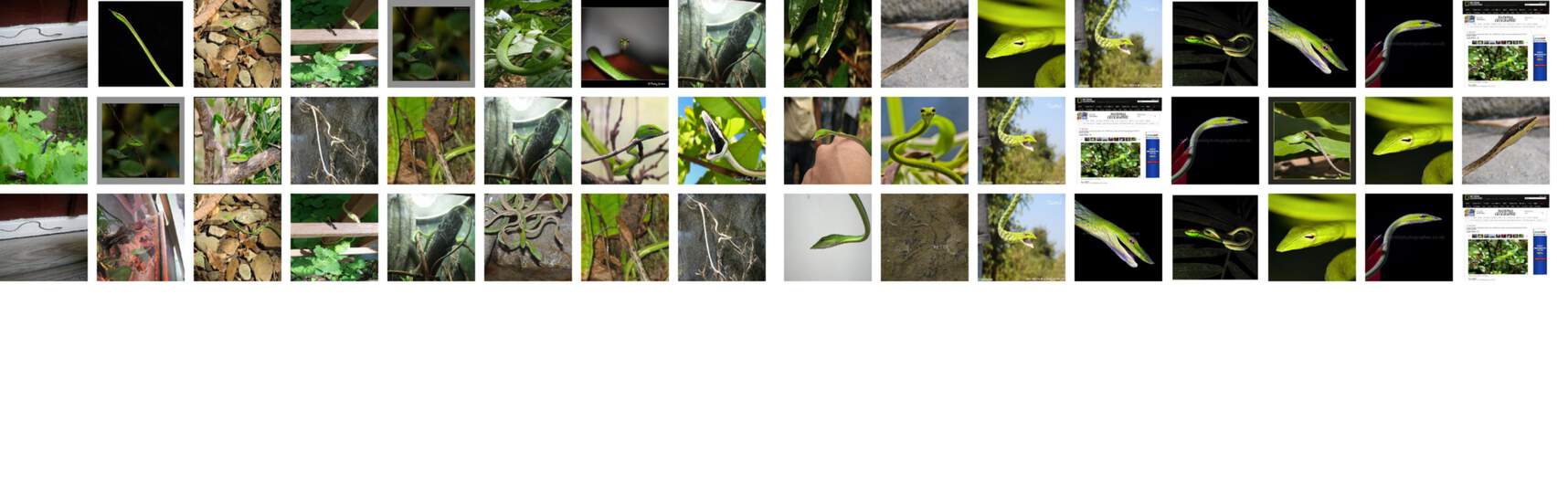}{0.05}{0}
\clearpage

\insertOutliersToPrototypesImageImagenet{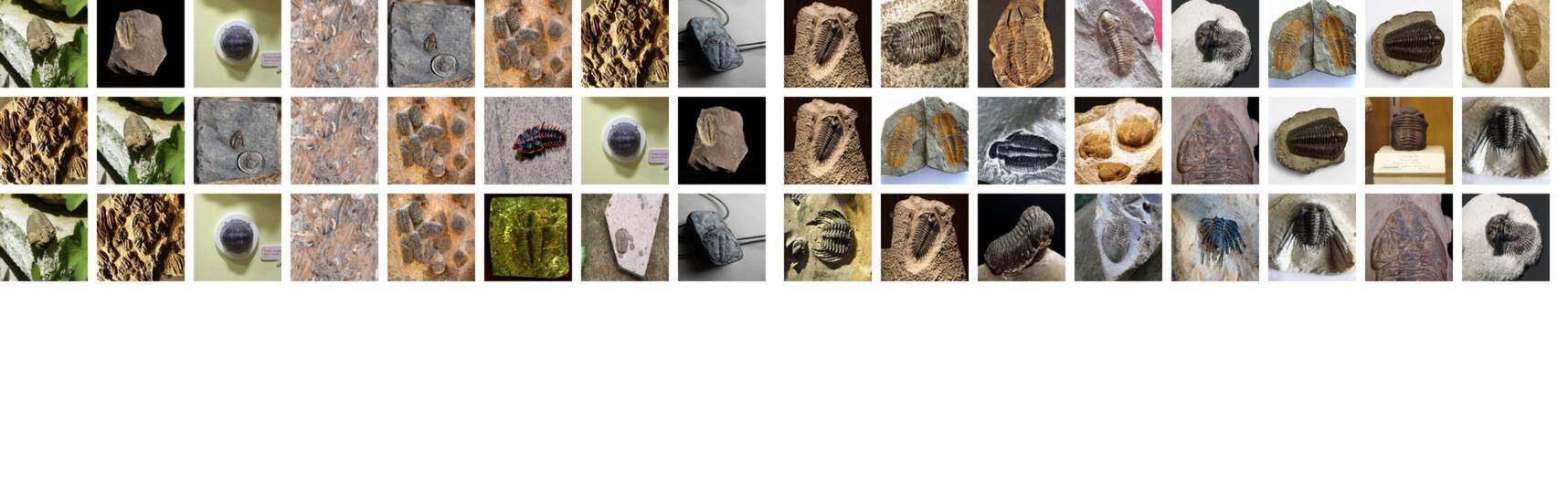}{0.05}{0}

\insertOutliersToPrototypesImageImagenet{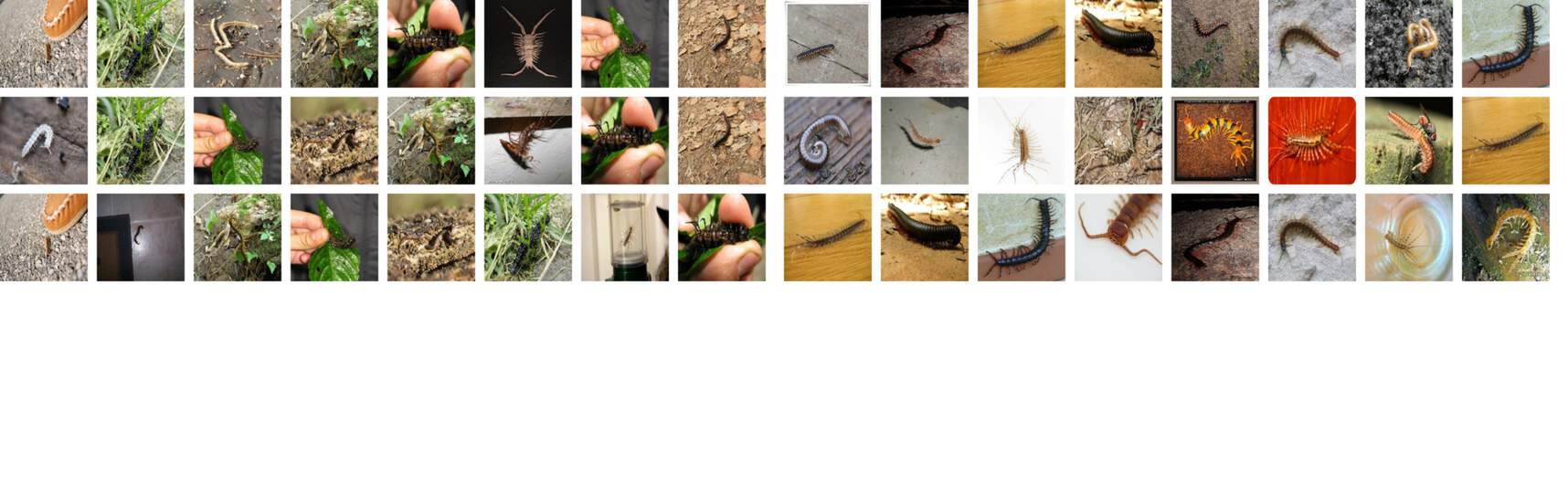}{0.05}{0}

\insertOutliersToPrototypesImageImagenet{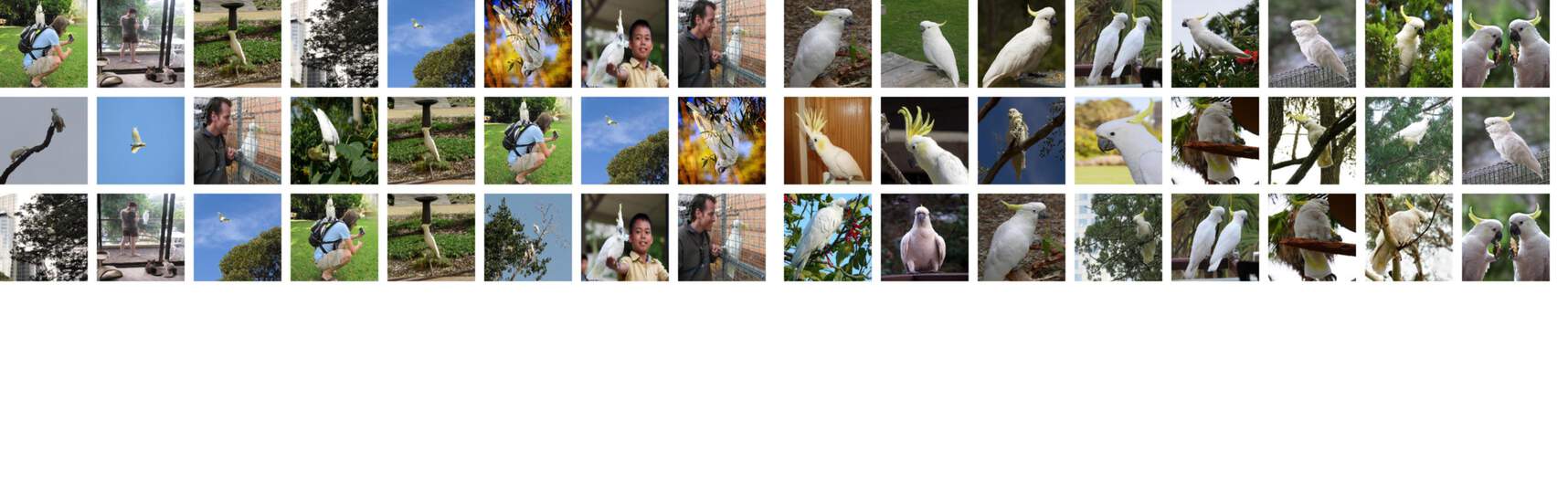}{0.05}{0}

\insertOutliersToPrototypesImageImagenet{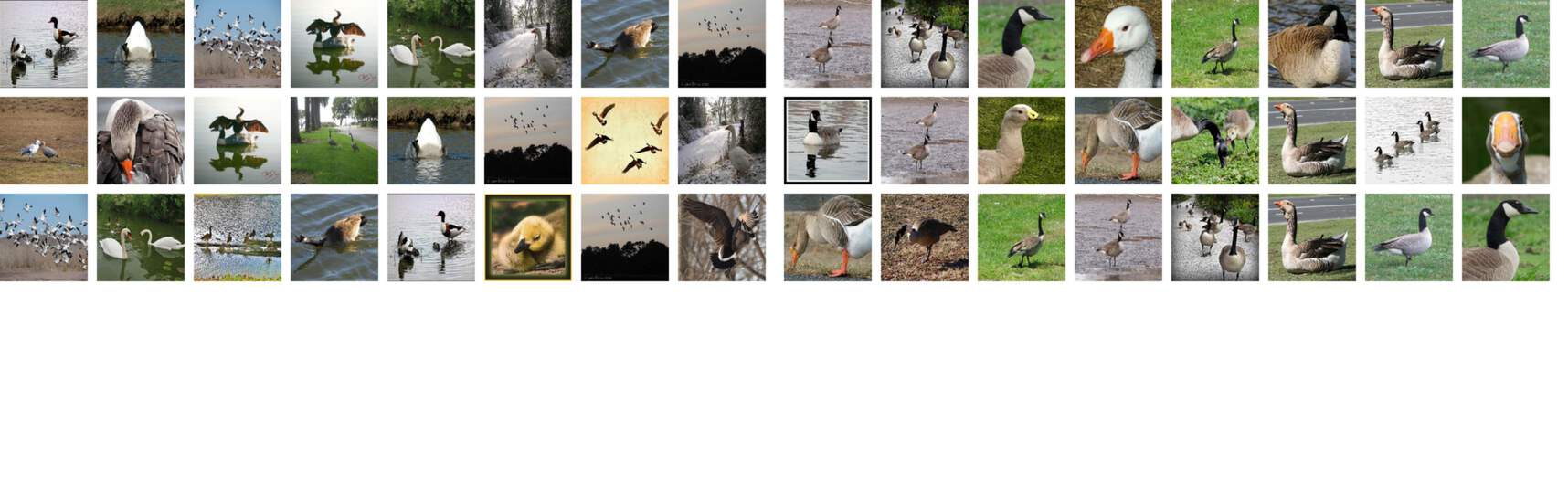}{0.05}{0}

\newpage
\section{Accuracy on well represented data when training only on well represented data}

The  three matrices that follow respectively report the accuracy of MNIST, Fashion-MNIST
and CIFAR-10 models learned on training examples with varying degrees of
prototypicality and evaluated on test examples also with varying degrees of
prototypicality. Specifically, the model used to compute cell $(i,j)$ of a matrix 
is learned on training data that is ranked in the $i^{th}$ percentile 
of ~\textsf{adv} prototypicality. The model is then
evaluated on the test examples whose~\textsf{adv} prototypicality
falls under the $j^{th}$ prototypicality percentile. For all datasets, these matrices show
that performing well on non-outliers is possible even when the model is trained
on outliers. For MNIST, this shows again that training on outliers provides better
performance across the range of test data (from outliers to well represented examples).
For Fashion-MNIST and CIFAR-10, this best performance is achieved by
training on examples that are neither prototypical nor outliers.

\begin{figure}
	\centering
\includegraphics[width=0.9\linewidth]{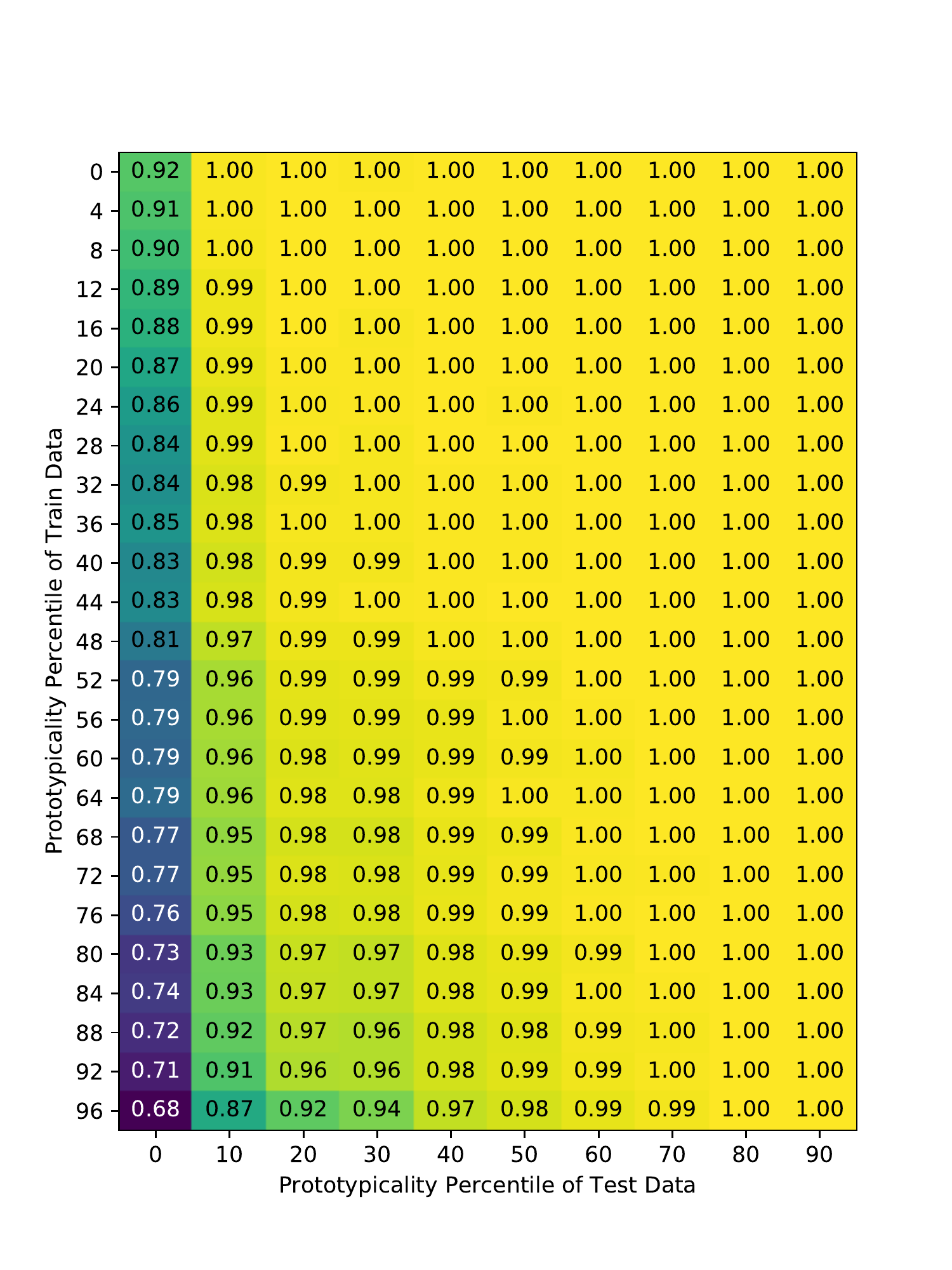}
\caption{MNIST}
\end{figure}

\begin{figure}[p]
	\centering
	\includegraphics[width=0.9\linewidth]{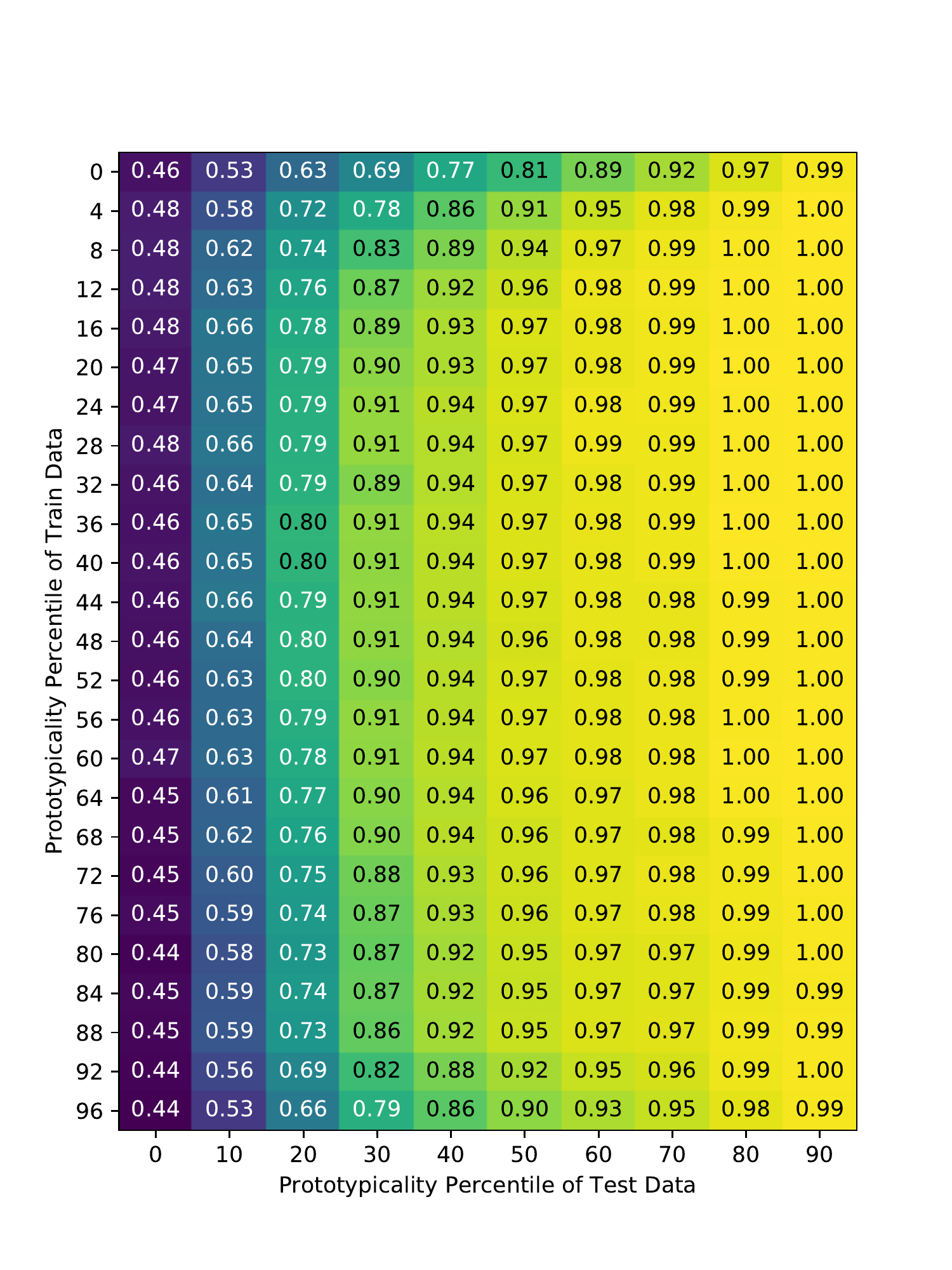}
	\caption{Fashion-MNIST}
\end{figure}

\begin{figure}[p]
	\centering
	\includegraphics[width=0.9\linewidth]{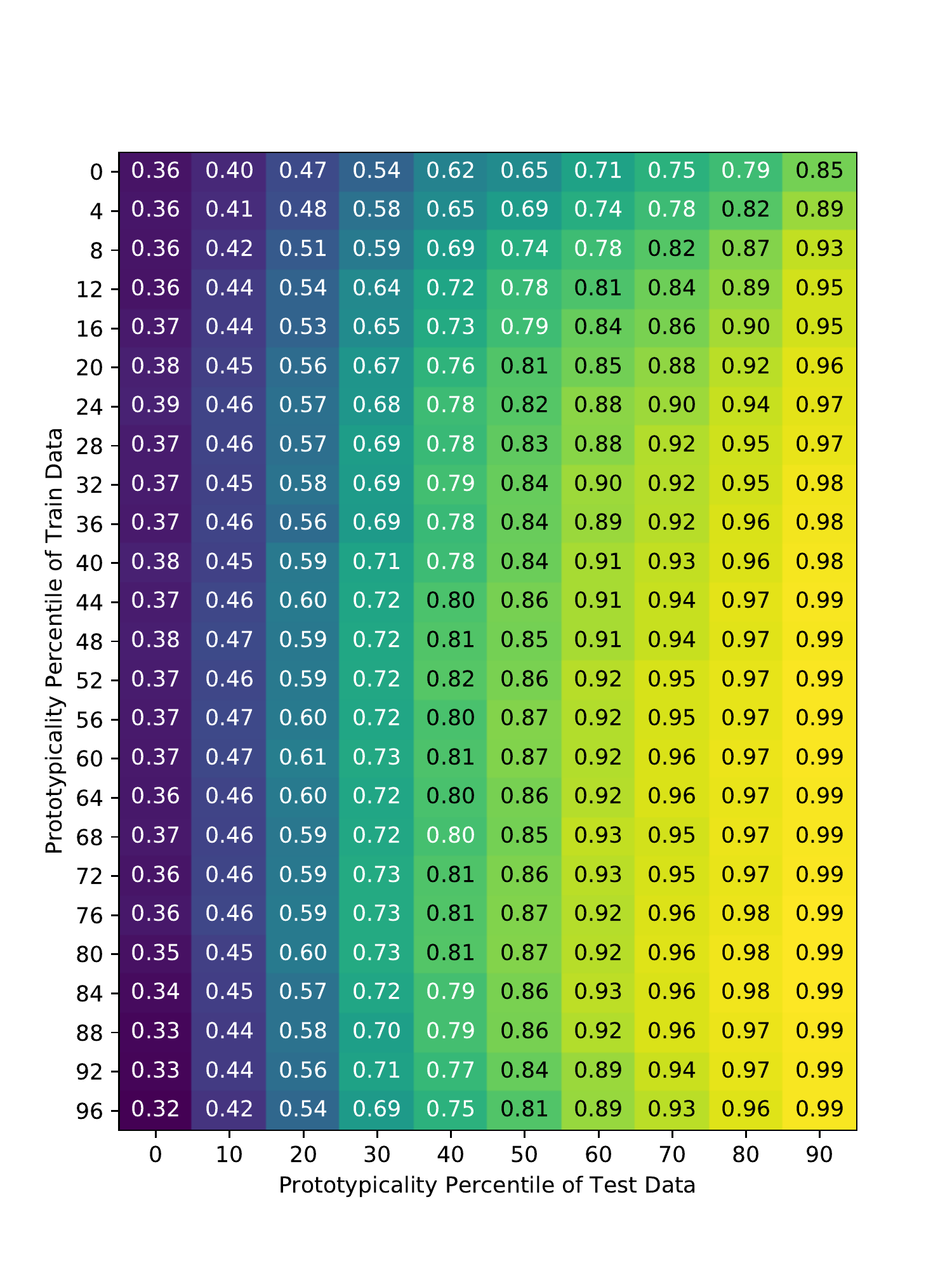}
	\caption{CIFAR-10}
\end{figure}

\newpage
\section{Human Study Example}
\label{apx:humanstudy}
We presented Mechanical Turk taskers with the following webpage, asking
them to select the worst image of the nine in the grid.
\begin{figure}
	\includegraphics[width=0.75\linewidth]{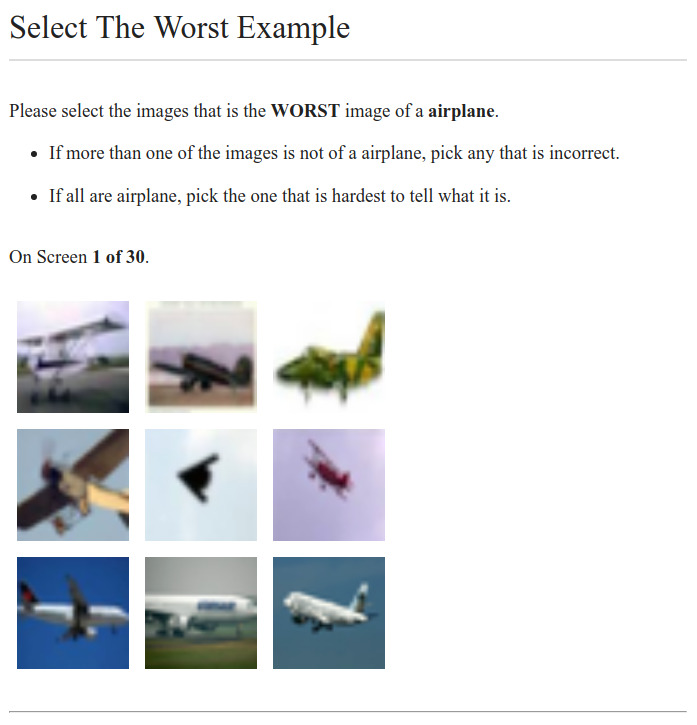}
\end{figure}
\section{Adversarial robustness of model trained on only well represented examples}
\label{apx:adv-ex}
\begin{figure}
	
	\begin{subfigure}{.31\textwidth}
		\includegraphics[width=\linewidth]{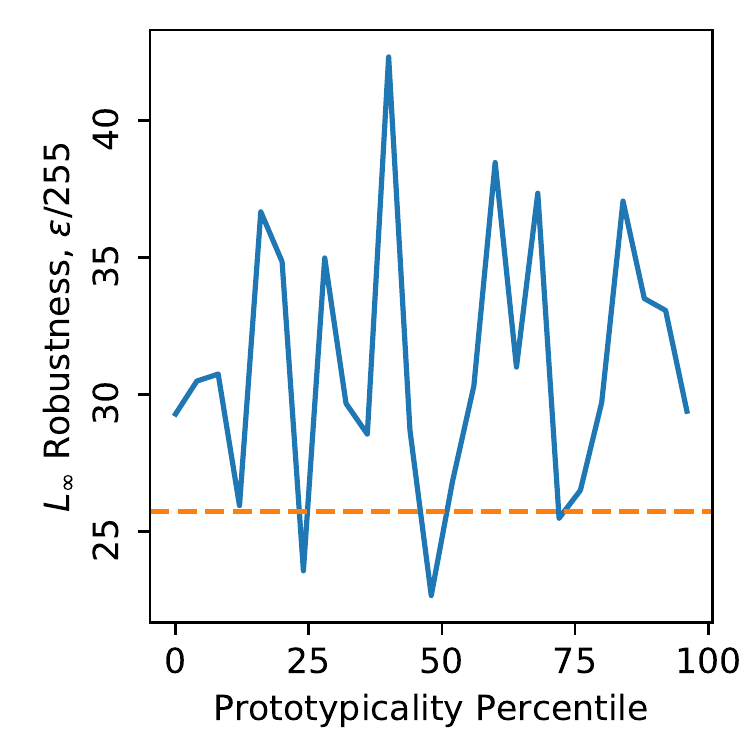}
		\caption{MNIST}
	\end{subfigure}
	\hspace*{\fill}
	\begin{subfigure}{.31\textwidth}
		\includegraphics[width=\linewidth]{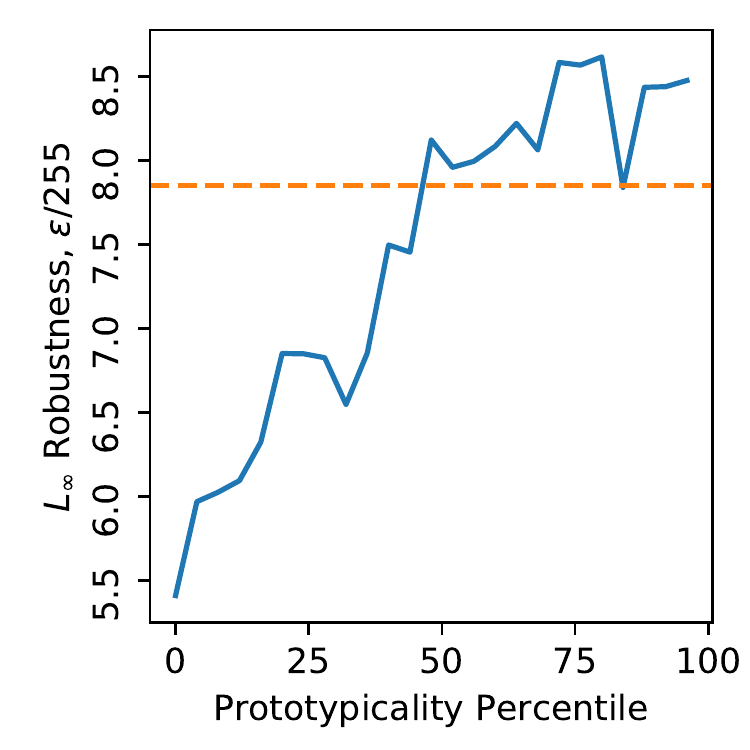}
		\caption{Fashion-MNIST}
	\end{subfigure}
	\hspace*{\fill}
	\begin{subfigure}{.31\textwidth}
		\includegraphics[width=\linewidth]{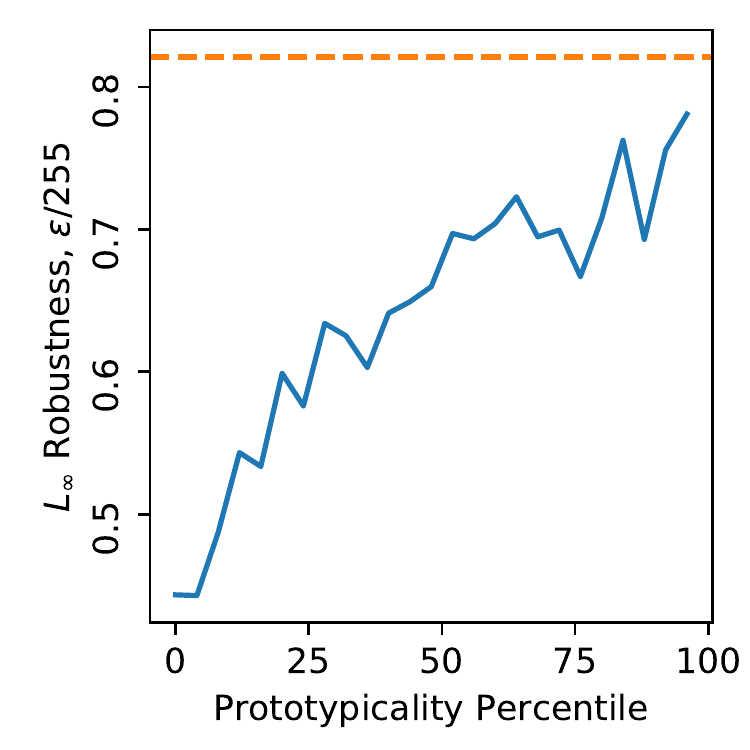}
		\caption{CIFAR-10}
	\end{subfigure}
	
	\caption{The blue curves indicate the training accuracy of models trained
		on slices of $5,000$ examples selected according to their prototypicality---as
		reported on the x-axis. 
		A baseline, obtained by training the model
		on the entire dataset is indicated by the dotted-orange line. 
		Models trained on well represented examples on Fashion-MNIST and CIFAR-10 are
		$2\times$ more robust to adversarial examples, when training on slices
		of $5,000$ prototypical examples as opposed to slices of $5,000$ outlier examples.
		On MNIST there is no significant difference; almost all examples are
		good.}
	\label{fig:protoadv}
\end{figure}

\newpage
\section{Revealing and clustering interesting examples and submodes}
\label{apx:clustering}

\begin{figure}
\includegraphics[width=\linewidth,trim=2cm 1cm 1.6cm 1cm,clip]{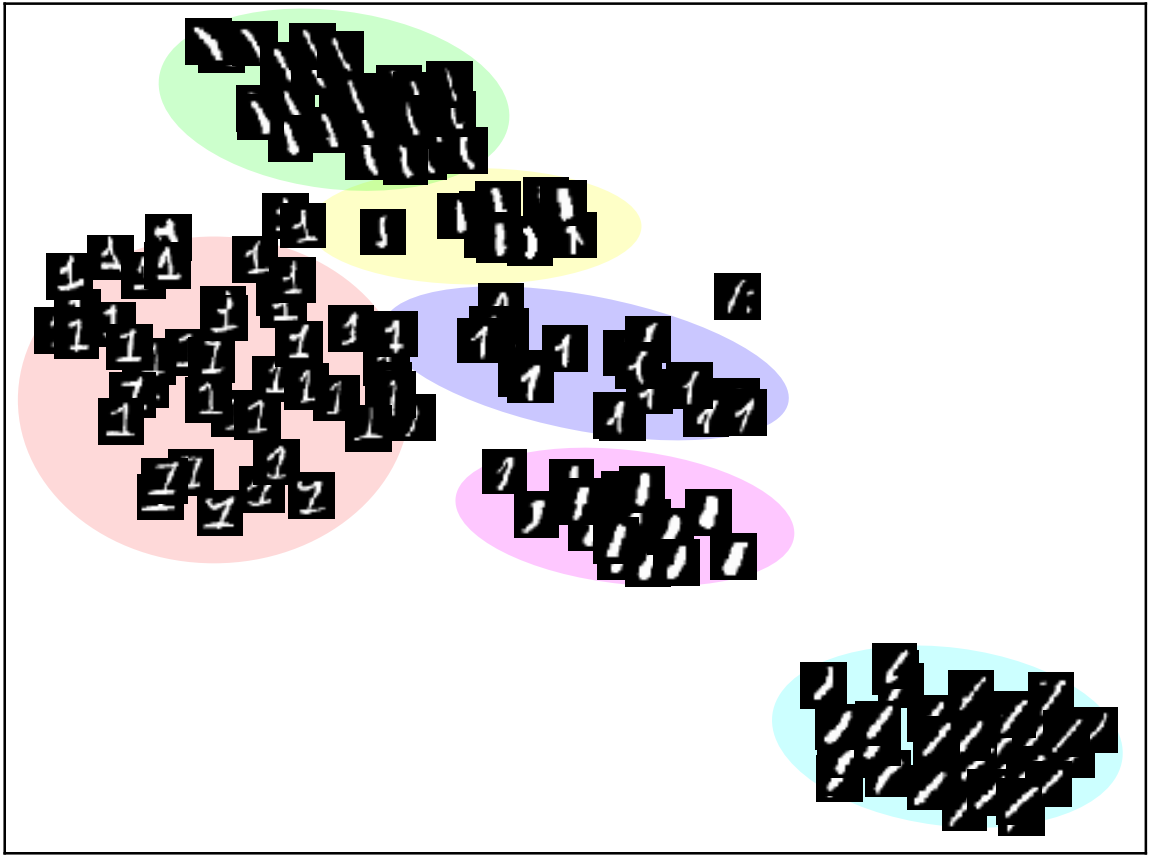}
\caption{Uncommon submodes found within the MNIST ``1'' class, and their HDBSCAN clusters.}
\end{figure}
\clearpage

\begin{figure}[t]
\includegraphics[width=\linewidth,trim=2cm 1cm 1.6cm 1cm,clip]{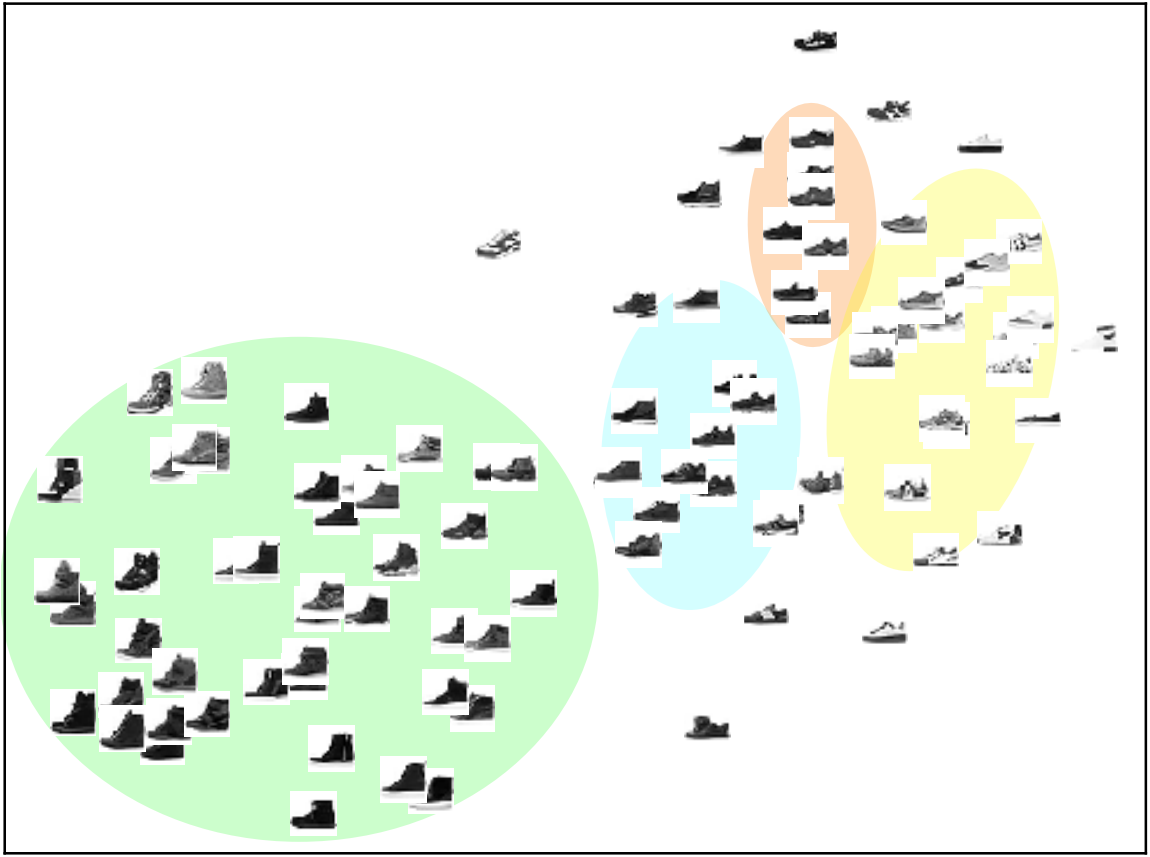}
\caption{Memorized exceptions in the Fashion-MNIST ``sneakers,'' and their HDBSCAN clusters.}
\end{figure}
\clearpage

\begin{figure}[t]
\includegraphics[width=\linewidth,trim=2cm 1cm 1.6cm 1cm,clip]{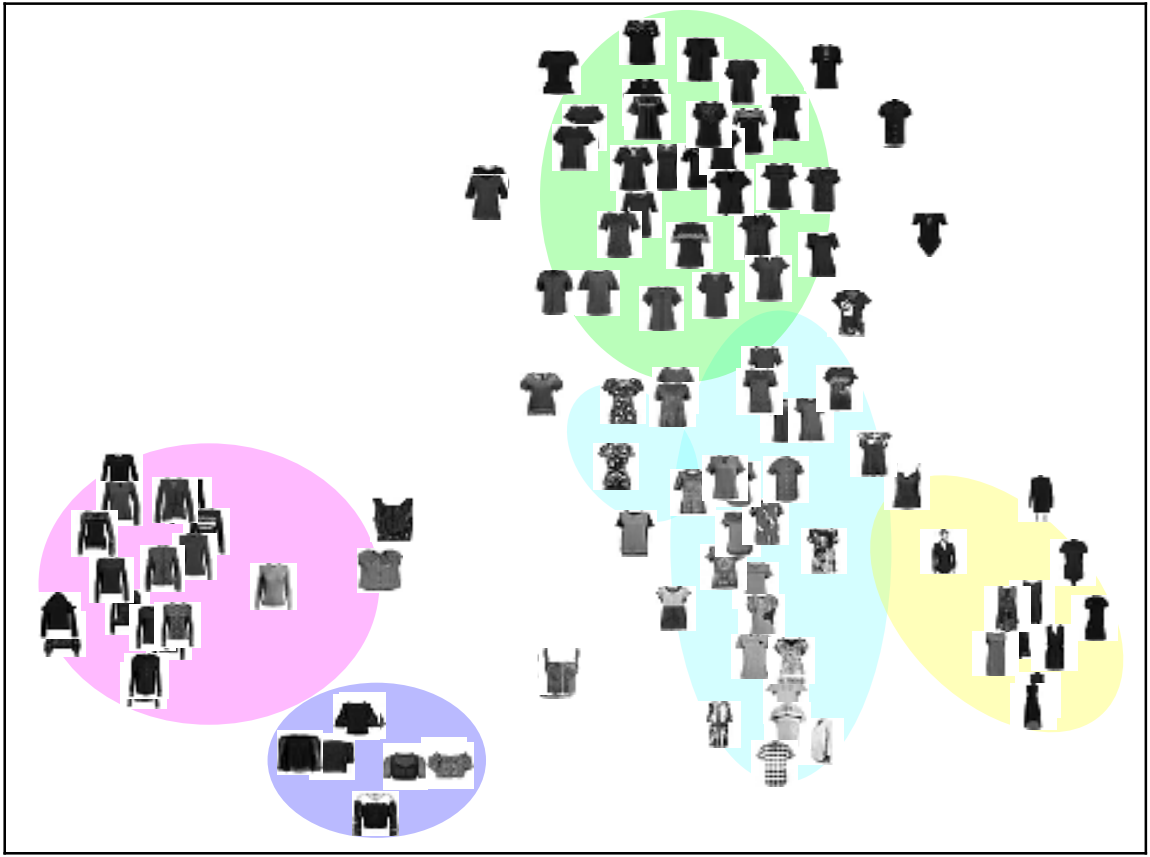}
\caption{Memorized exceptions in the  Fashion-MNIST ``shirts,'', and their HDBSCAN clusters.}
\end{figure}
\clearpage

\begin{figure}[t]
\includegraphics[width=\linewidth,trim=2cm 1cm 1.6cm 1cm,clip]{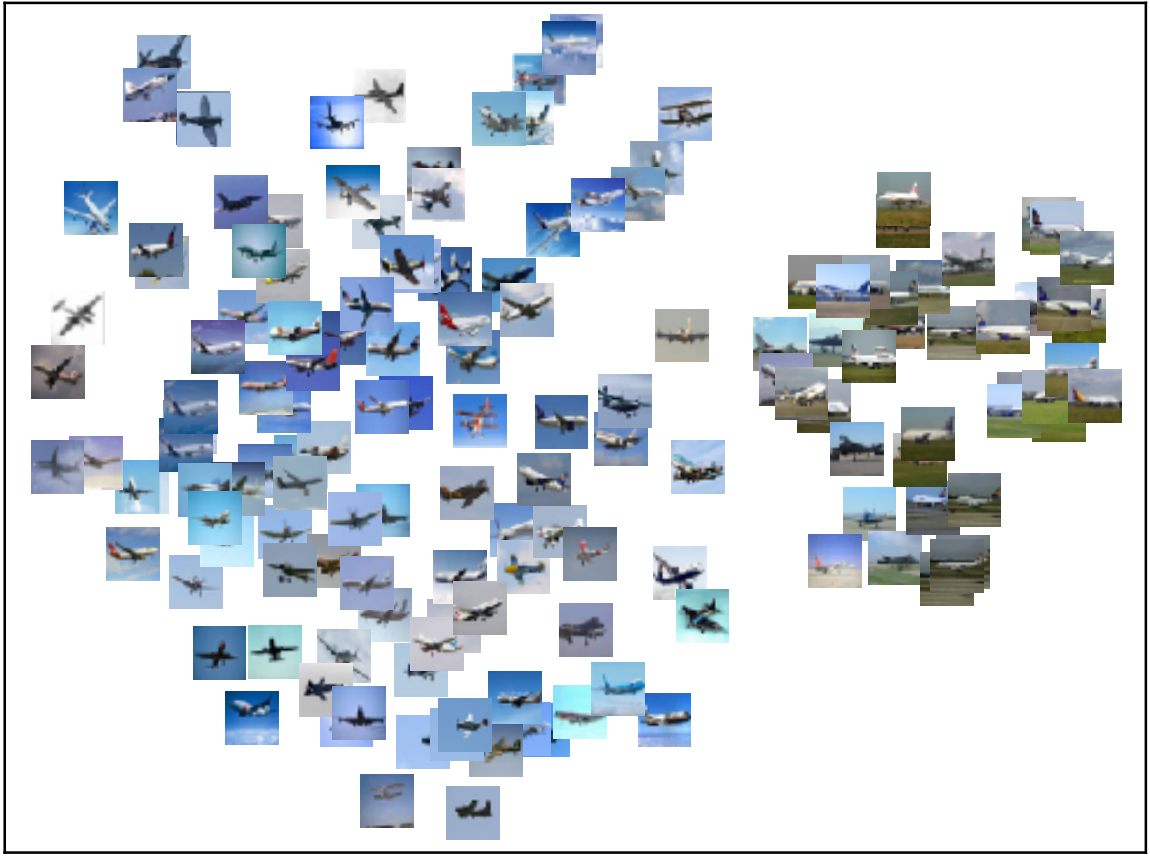}
\caption{Canonical prototypes in the CIFAR-10 ``airplane'' class.}
\end{figure}
\nopagebreak[4]

\section{Comparing density of our metrics for different metrics over the output classes of all learning tasks}
\label{apx:stability}
\includegraphics[width=.45\linewidth,trim=2cm 0cm 6cm 0cm,clip,trim=2cm 0cm 6cm 0cm,clip]{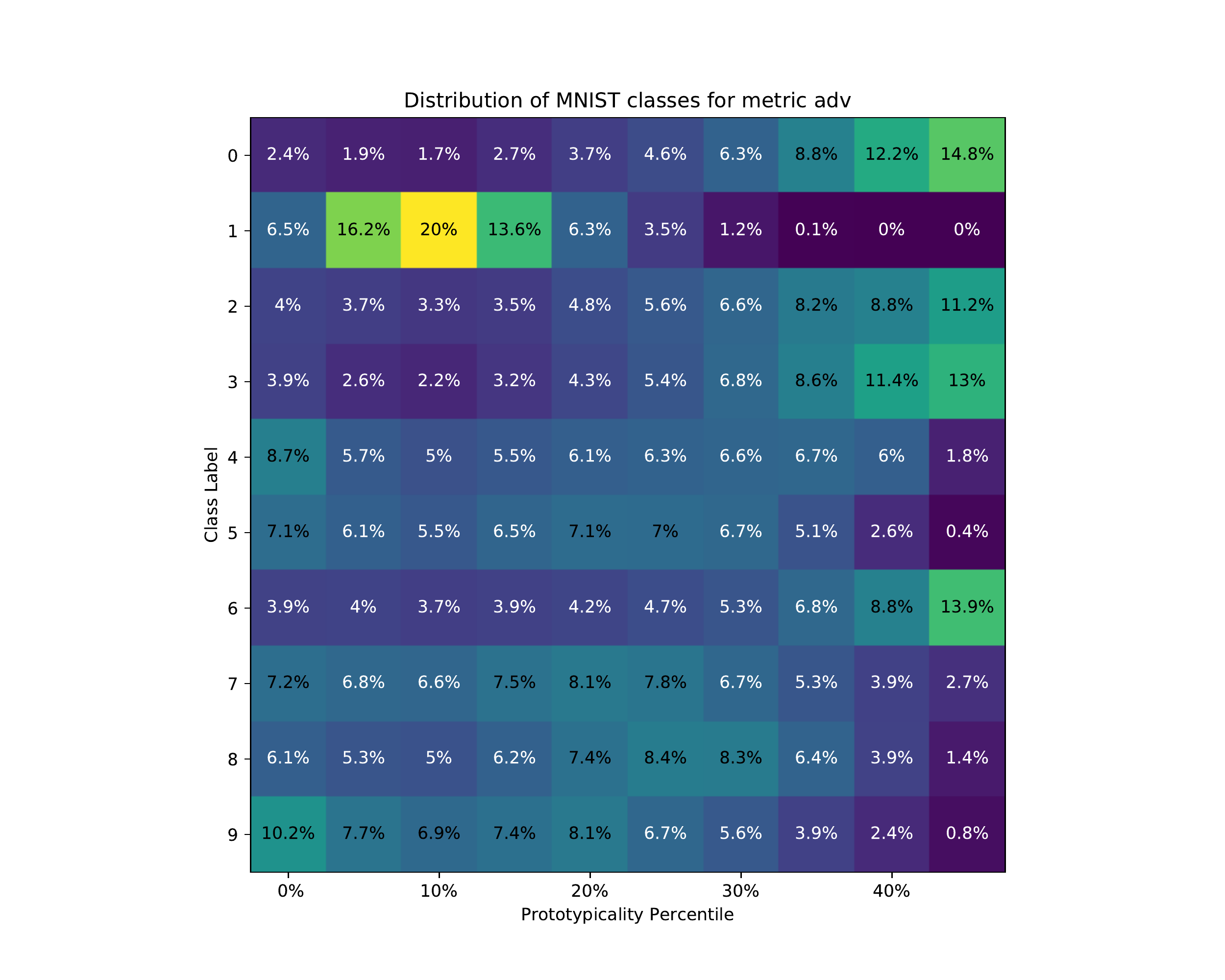}
\includegraphics[width=.45\linewidth,trim=2cm 0cm 6cm 0cm,clip]{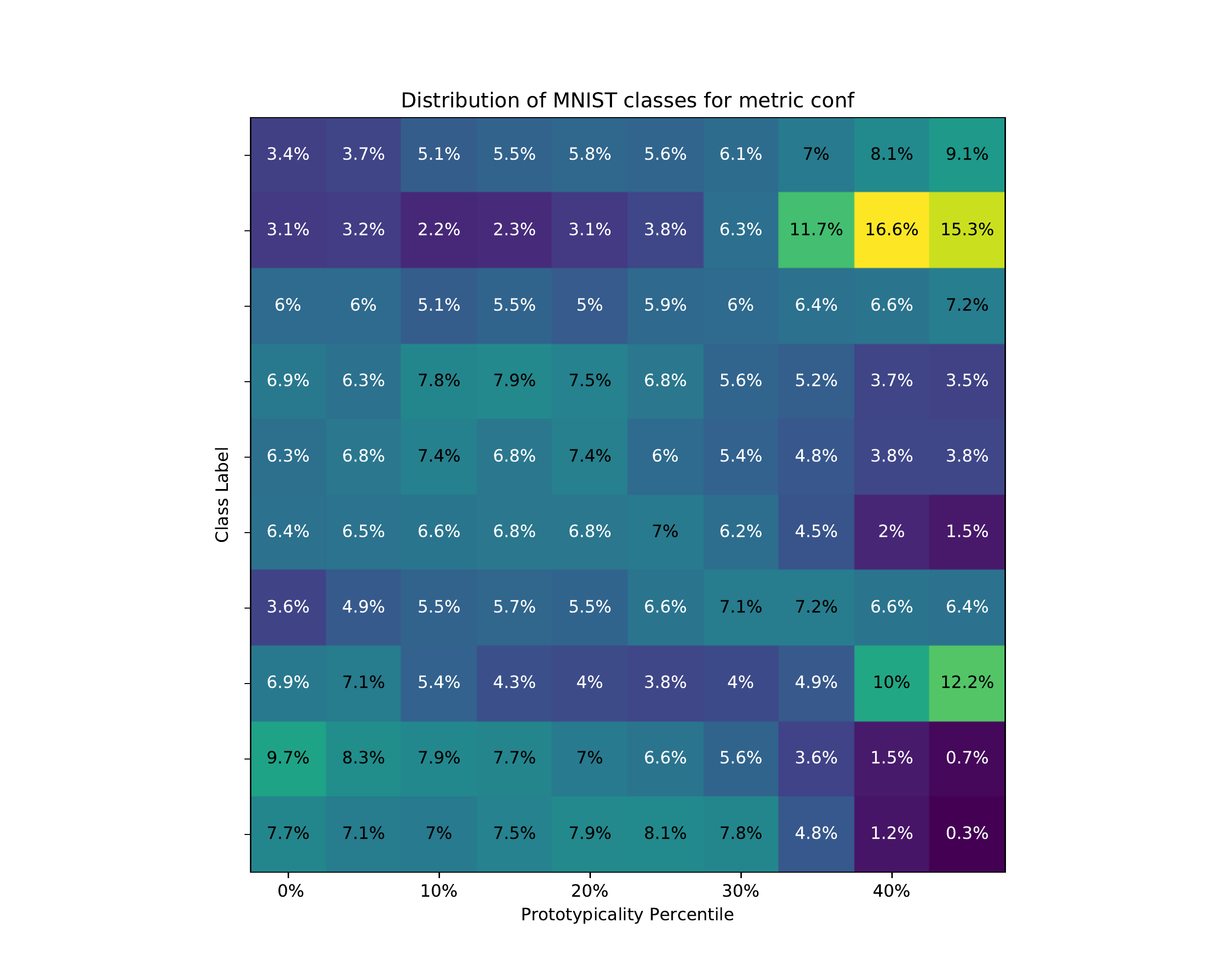} \\
\includegraphics[width=.45\linewidth,trim=2cm 0cm 6cm 0cm,clip]{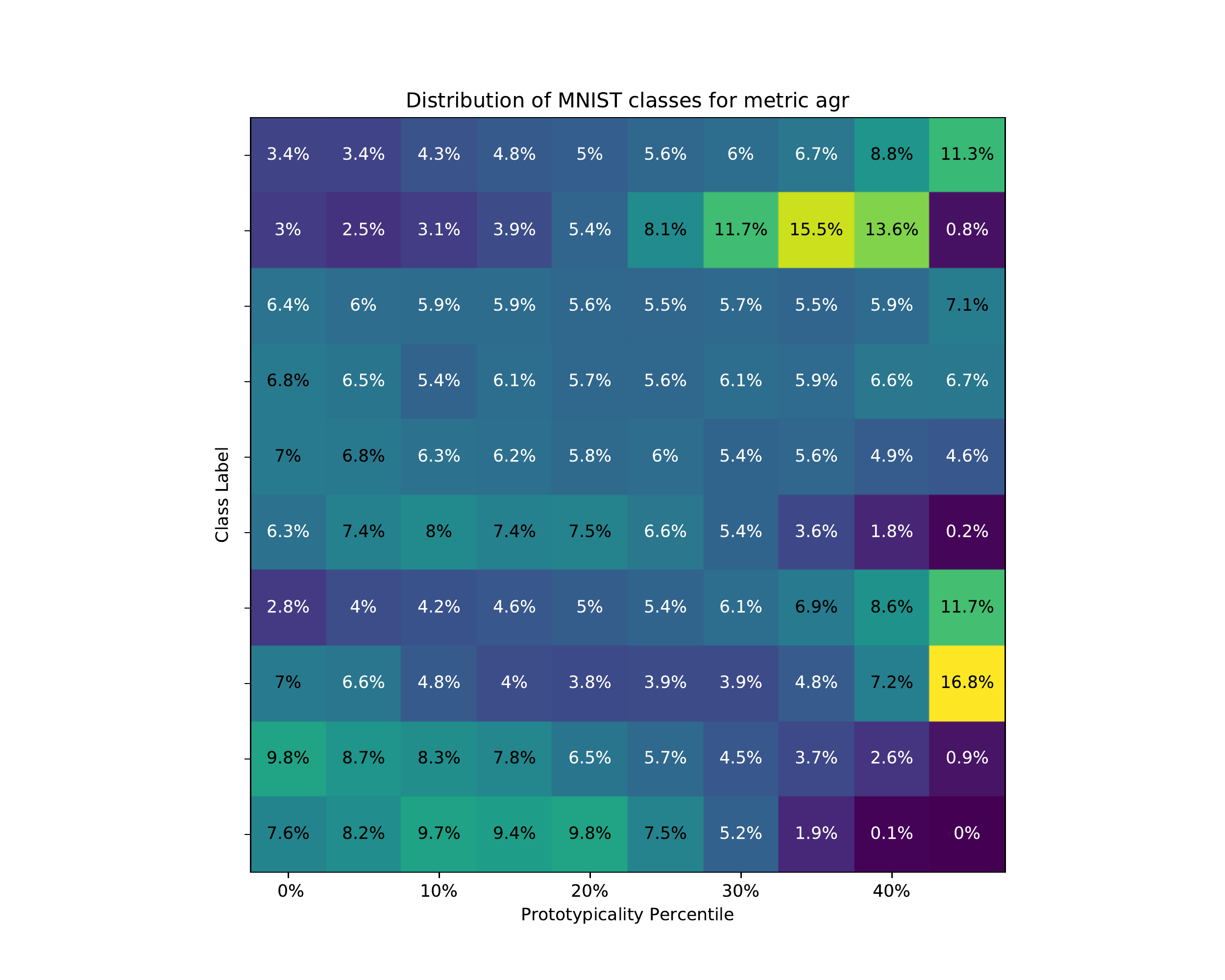}
\includegraphics[width=.45\linewidth,trim=2cm 0cm 6cm 0cm,clip]{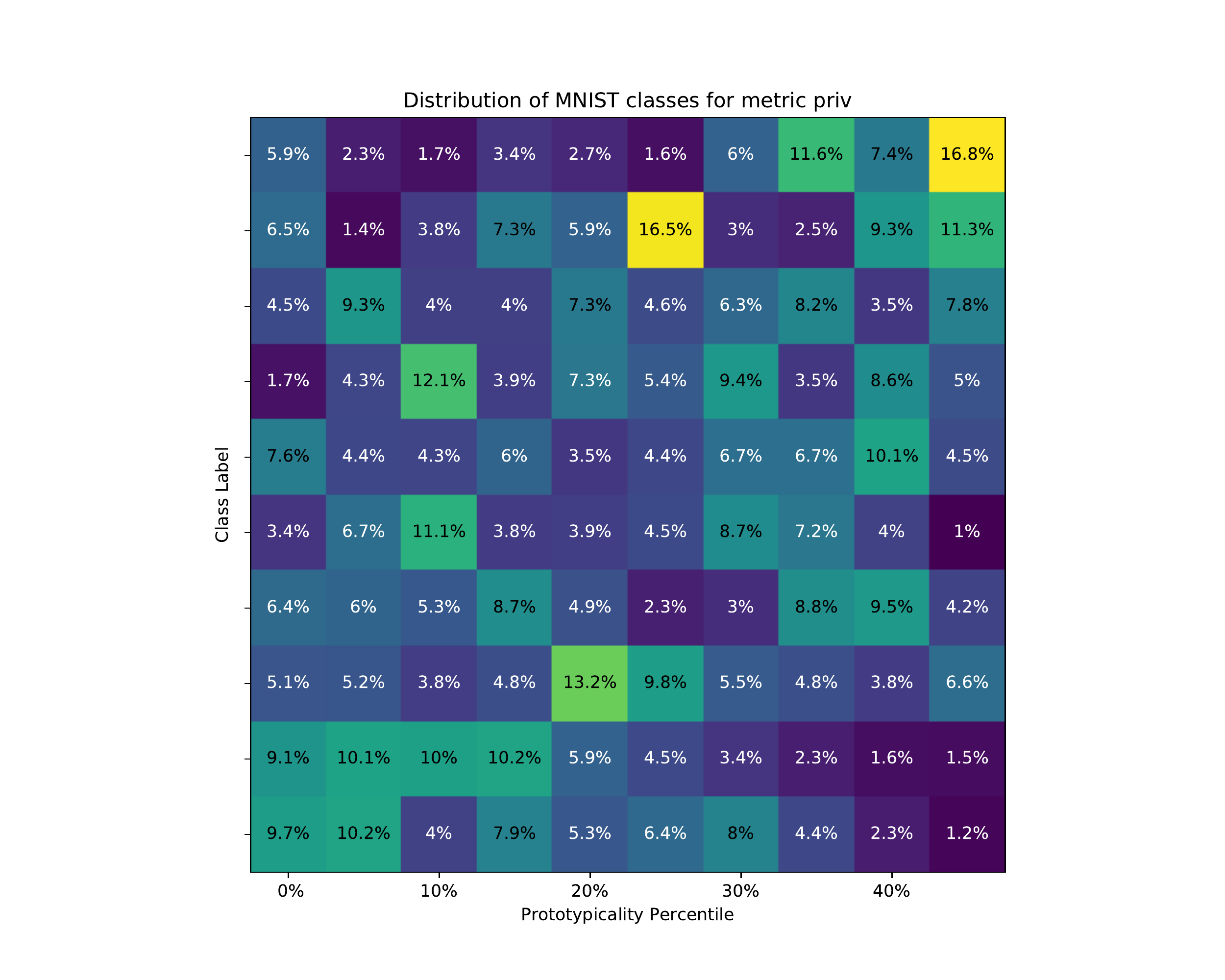} \\
\includegraphics[width=.45\linewidth,trim=2cm 0cm 6cm 0cm,clip]{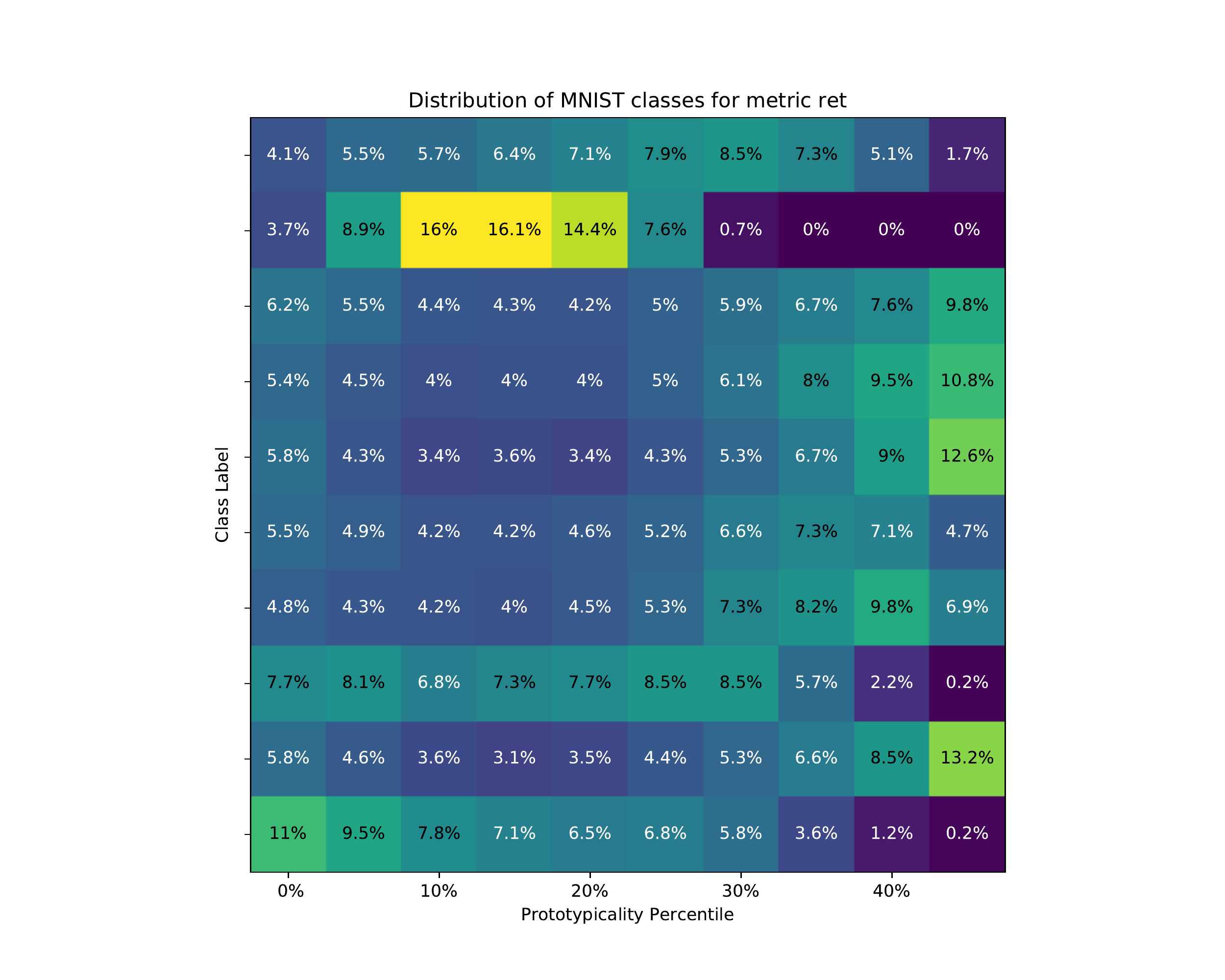}

\includegraphics[width=.45\linewidth,trim=2cm 0cm 6cm 0cm,clip]{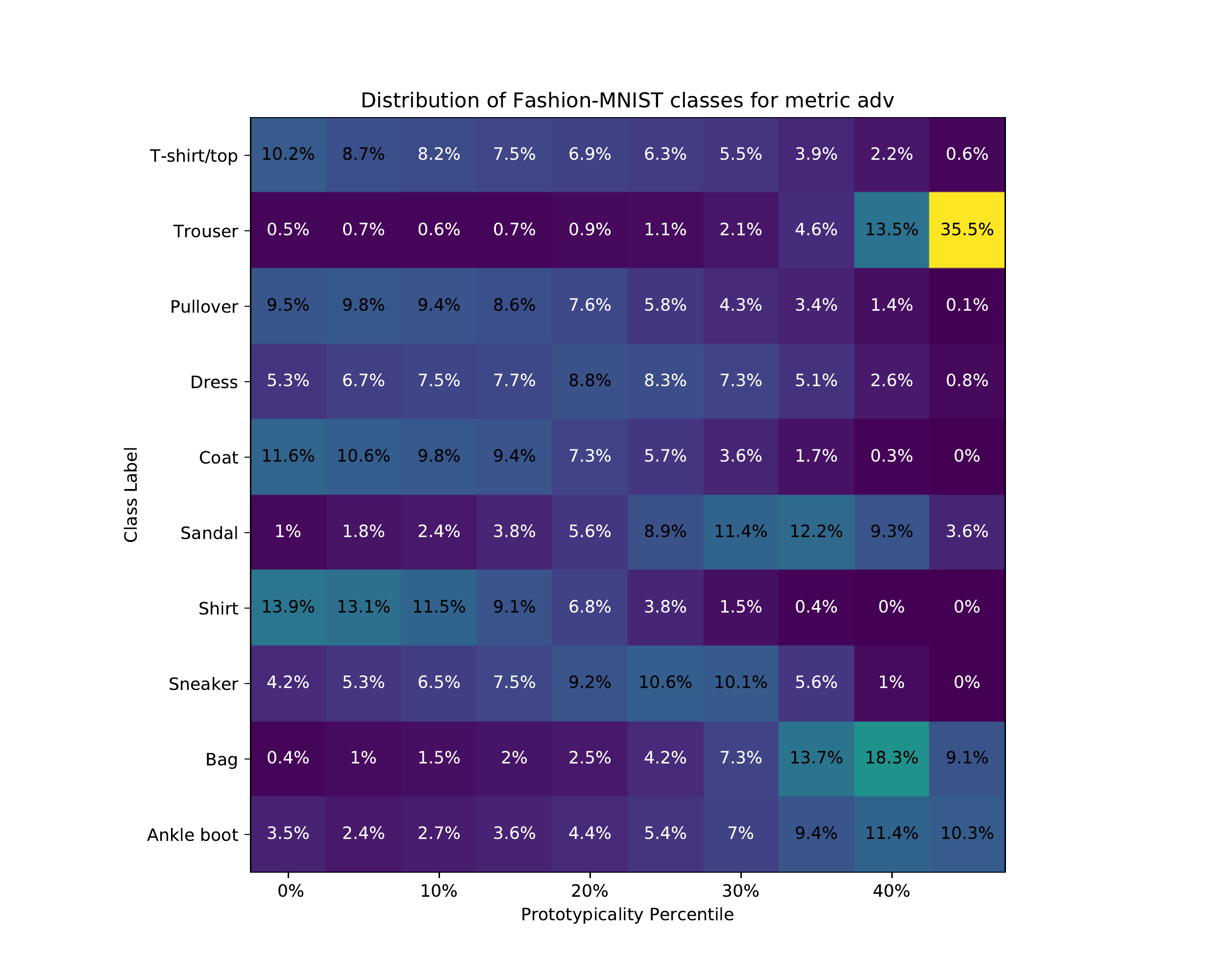}
\includegraphics[width=.45\linewidth,trim=2cm 0cm 6cm 0cm,clip]{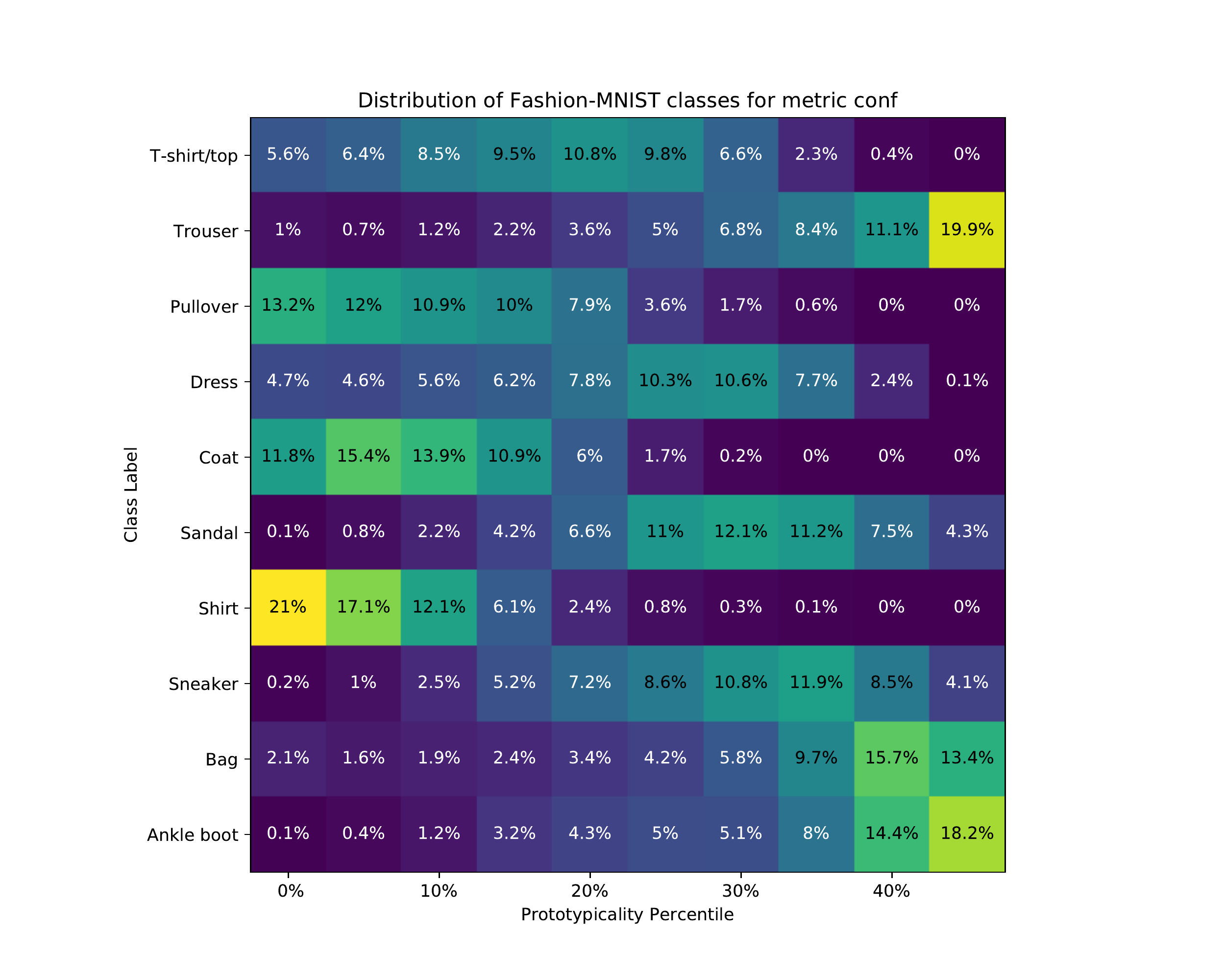} \\
\includegraphics[width=.45\linewidth,trim=2cm 0cm 6cm 0cm,clip]{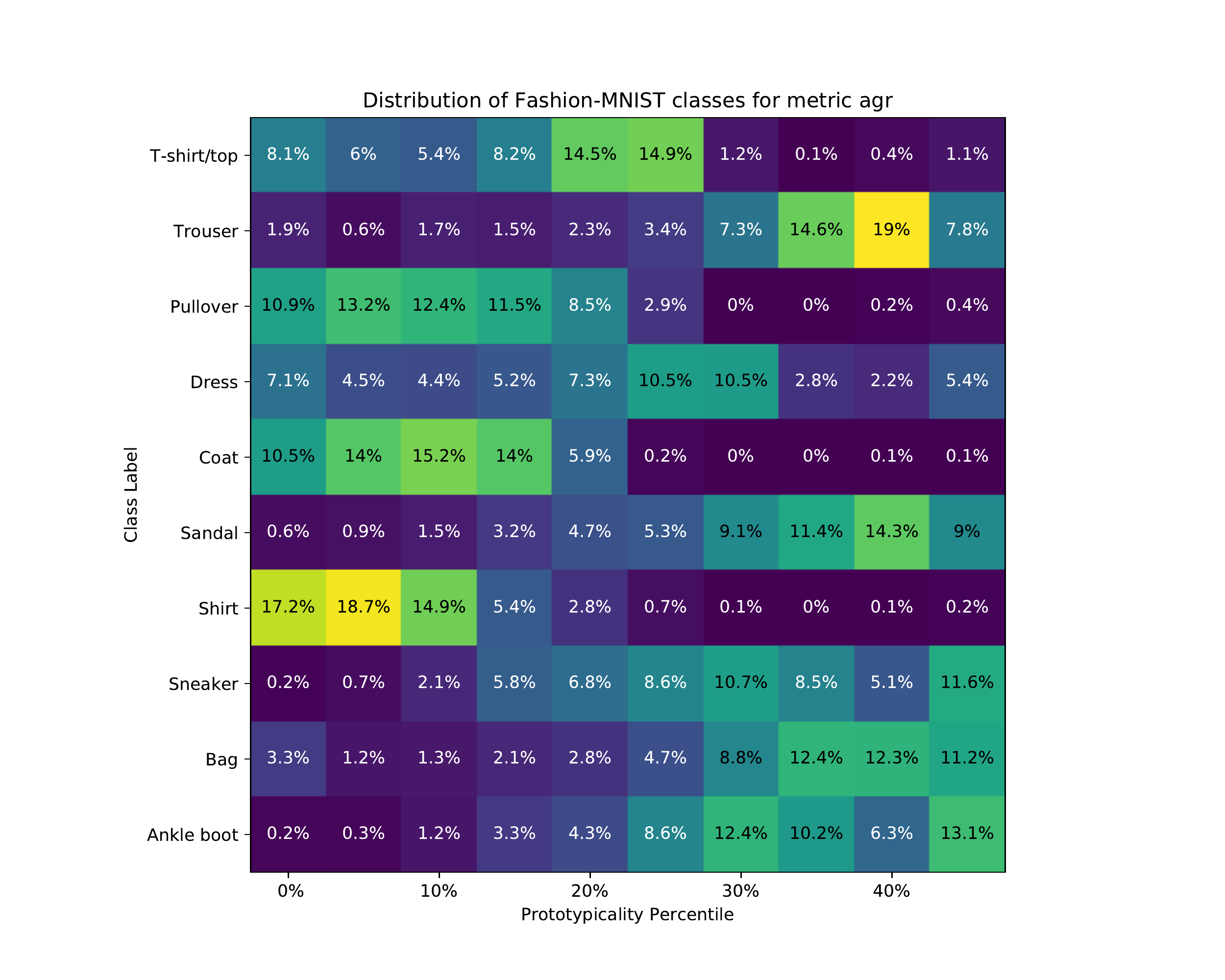}
\includegraphics[width=.45\linewidth,trim=2cm 0cm 6cm 0cm,clip]{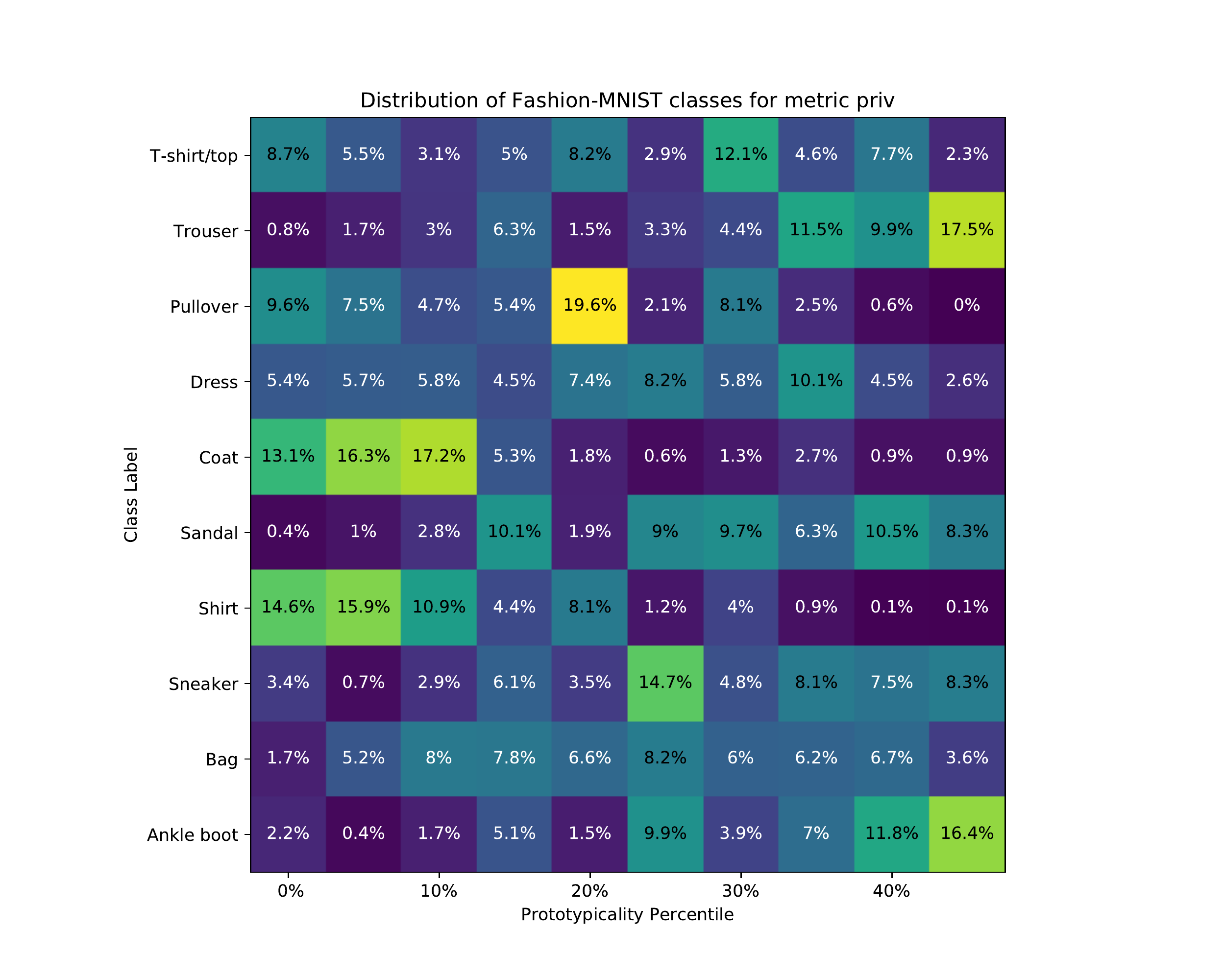} \\
\includegraphics[width=.45\linewidth,trim=2cm 0cm 6cm 0cm,clip]{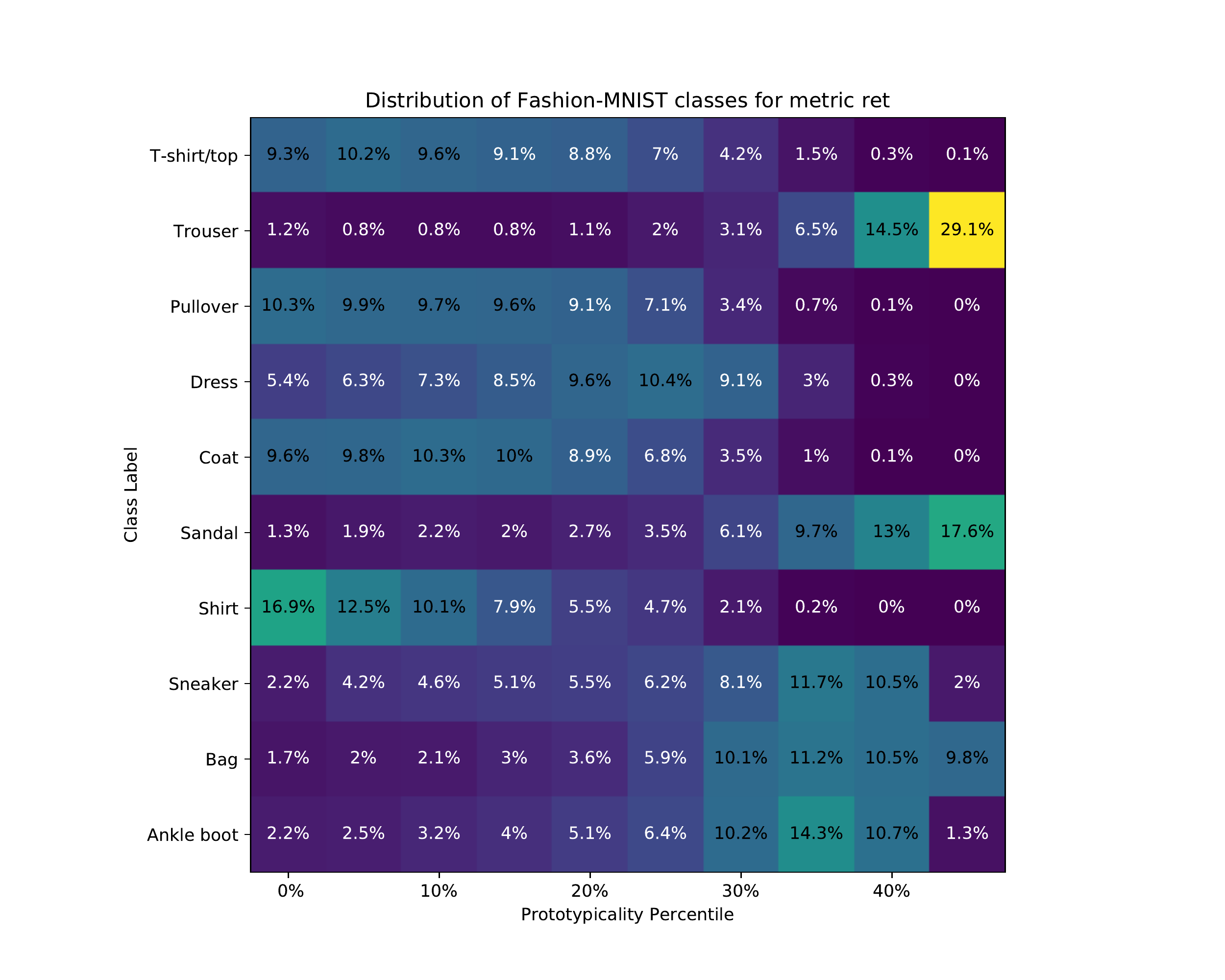}

\includegraphics[width=.45\linewidth,trim=2cm 0cm 6cm 0cm,clip]{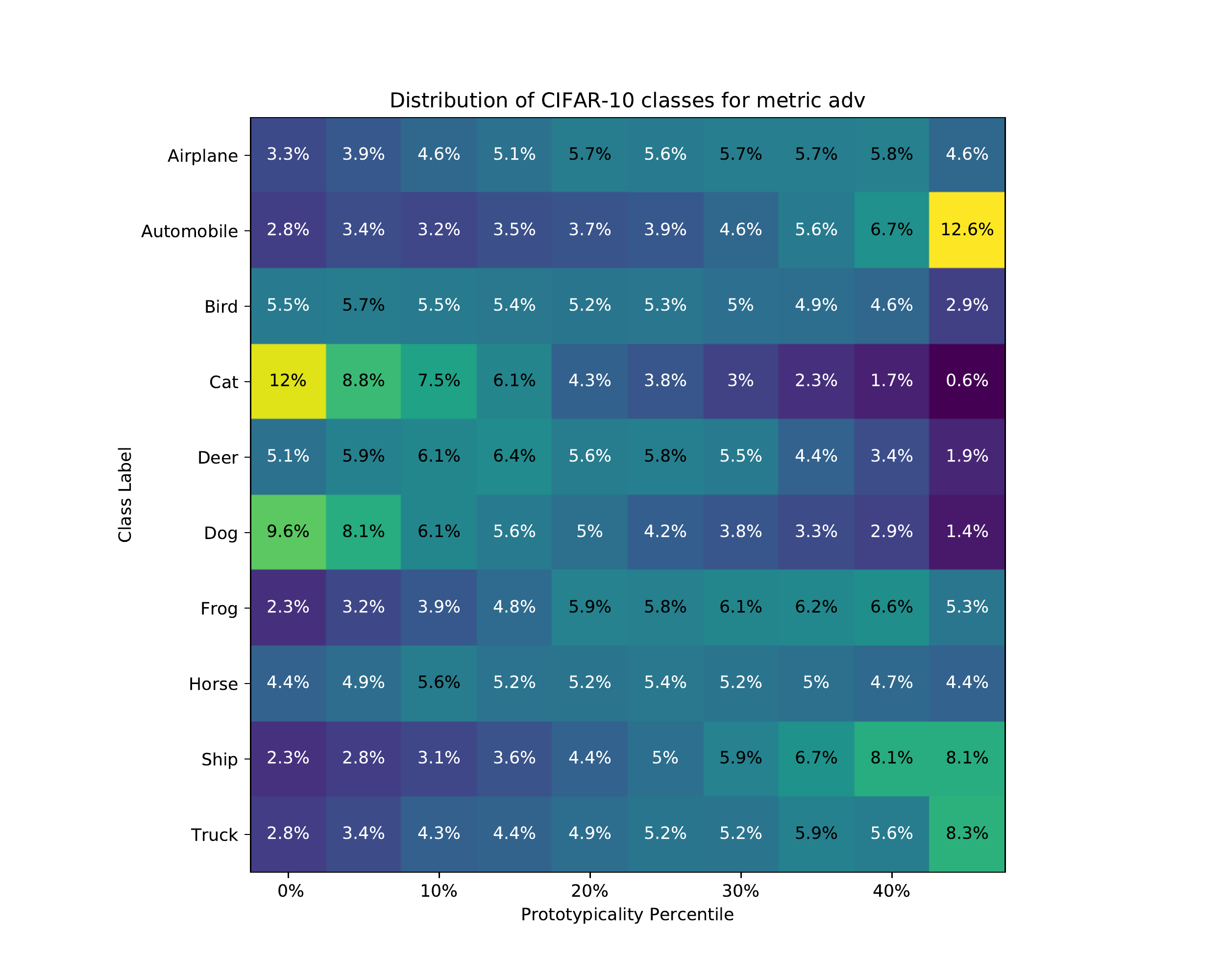}
\includegraphics[width=.45\linewidth,trim=2cm 0cm 6cm 0cm,clip]{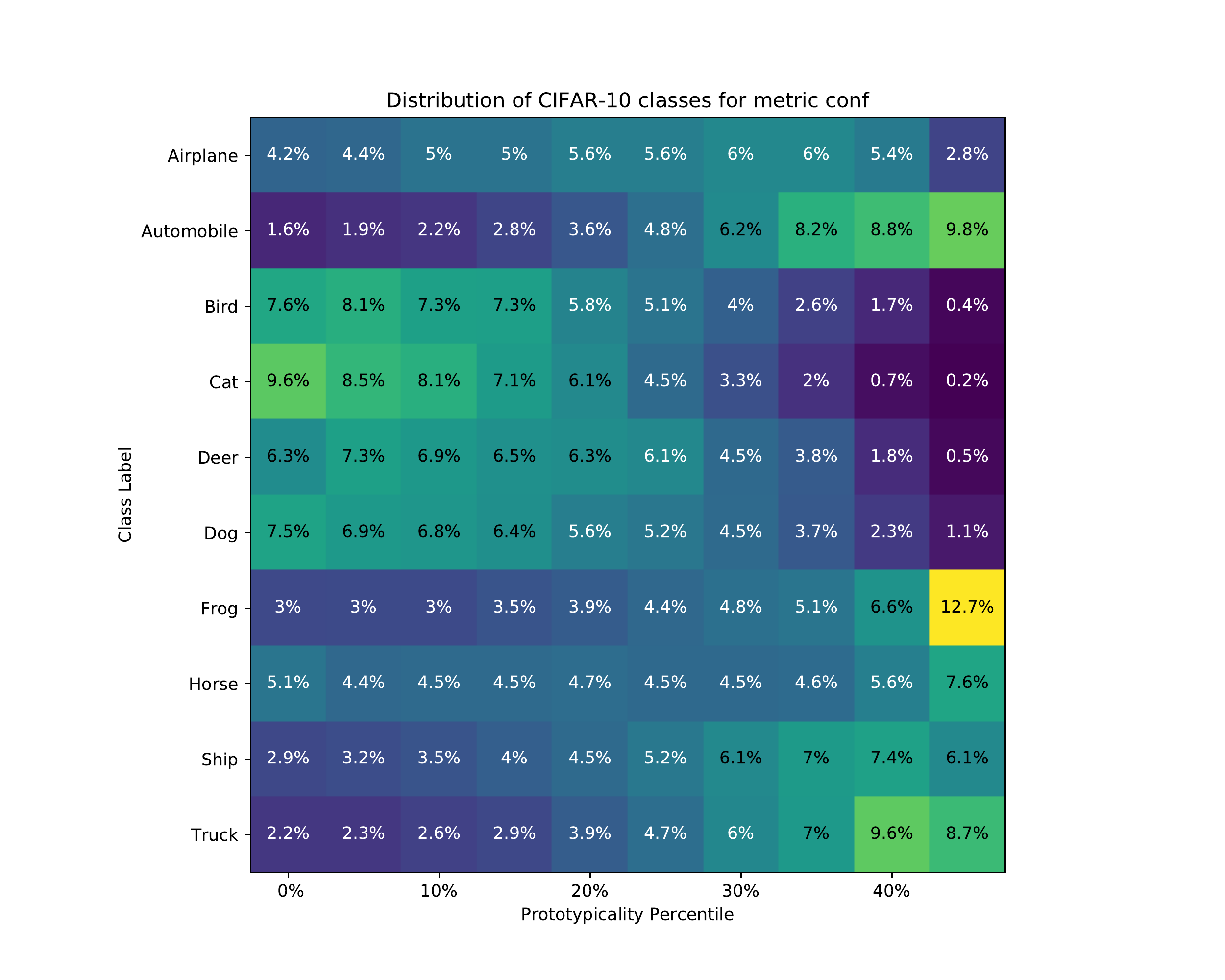}\\
\includegraphics[width=.45\linewidth,trim=2cm 0cm 6cm 0cm,clip]{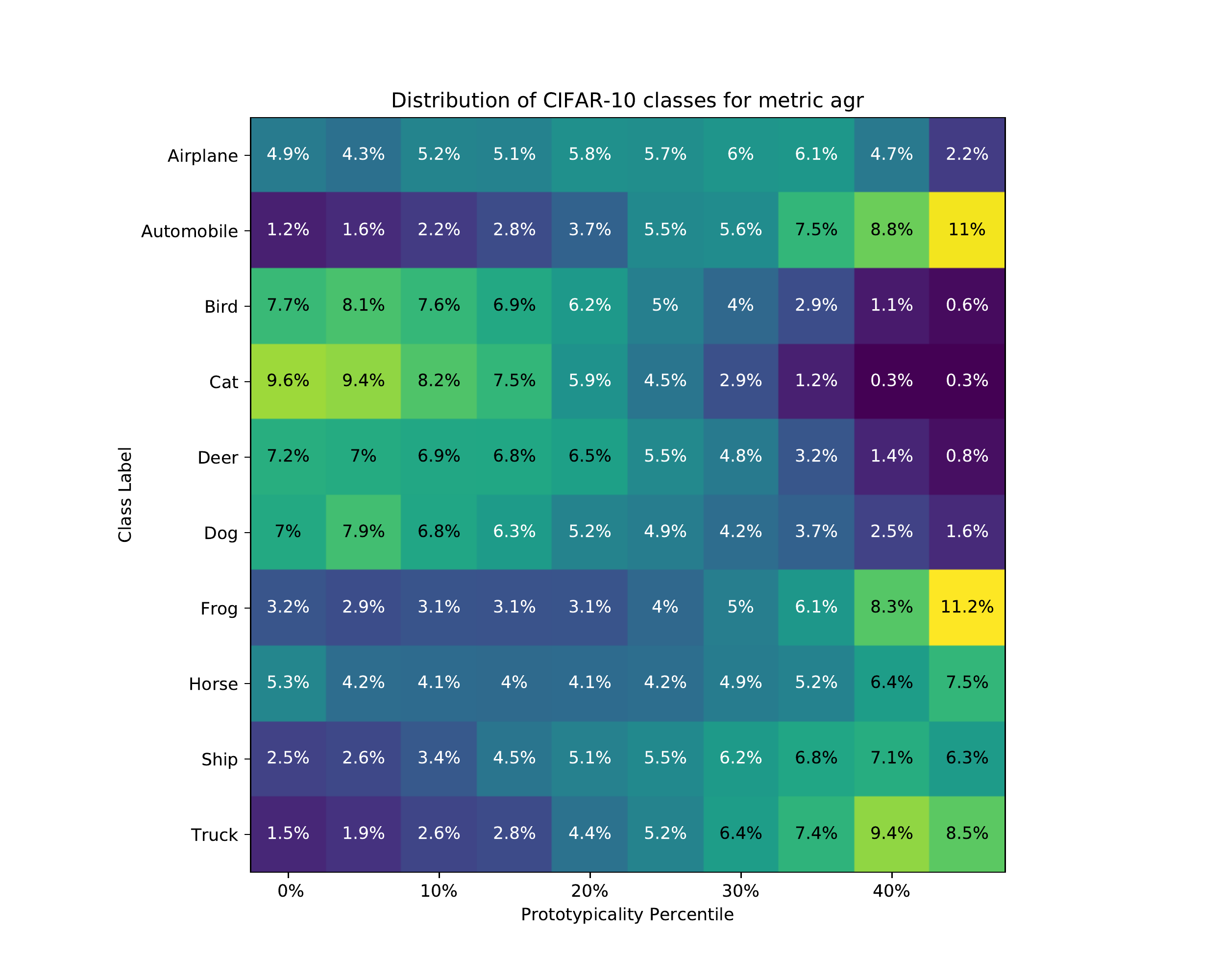}
\includegraphics[width=.45\linewidth,trim=2cm 0cm 6cm 0cm,clip]{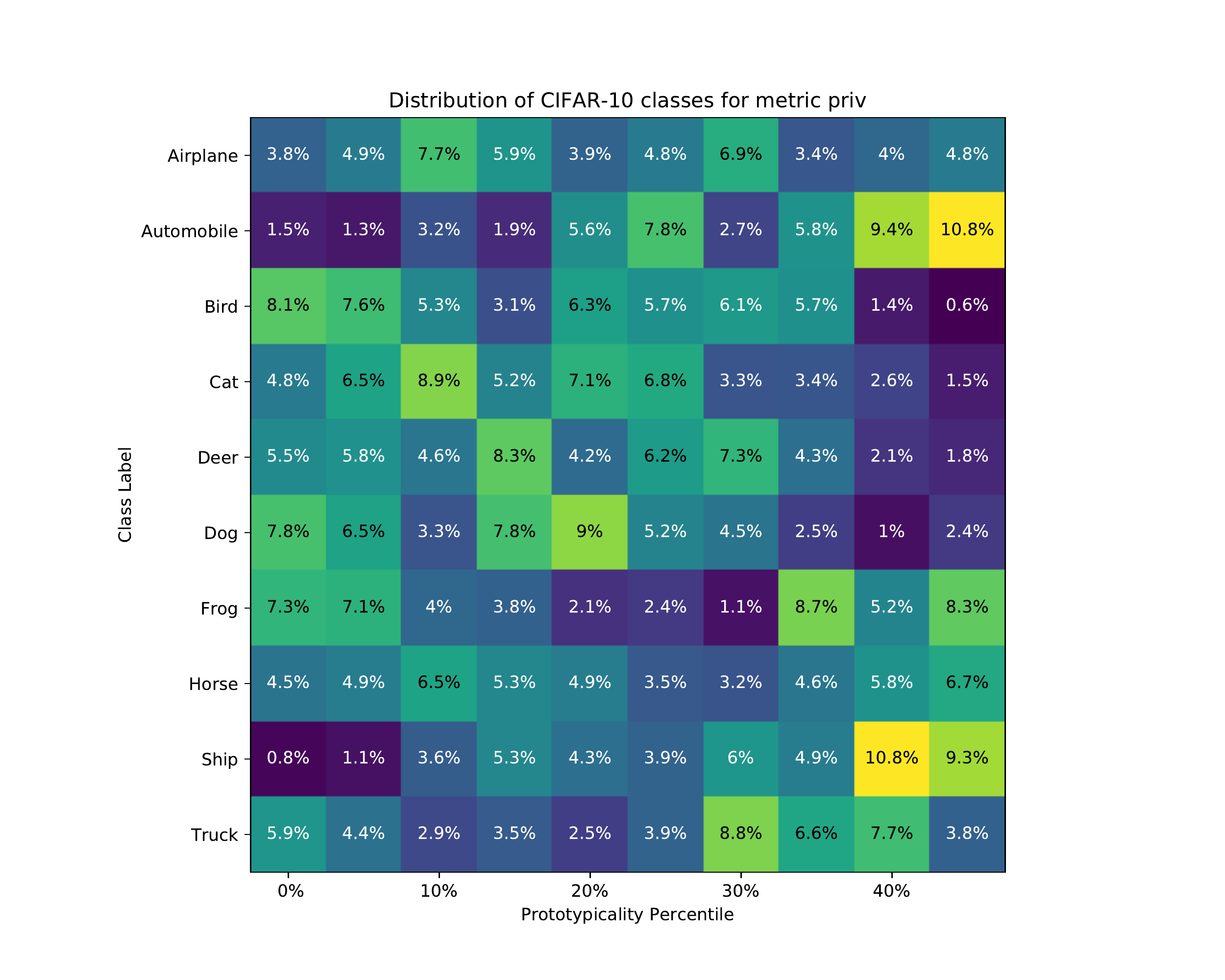}\\
\includegraphics[width=.45\linewidth,trim=2cm 0cm 6cm 0cm,clip]{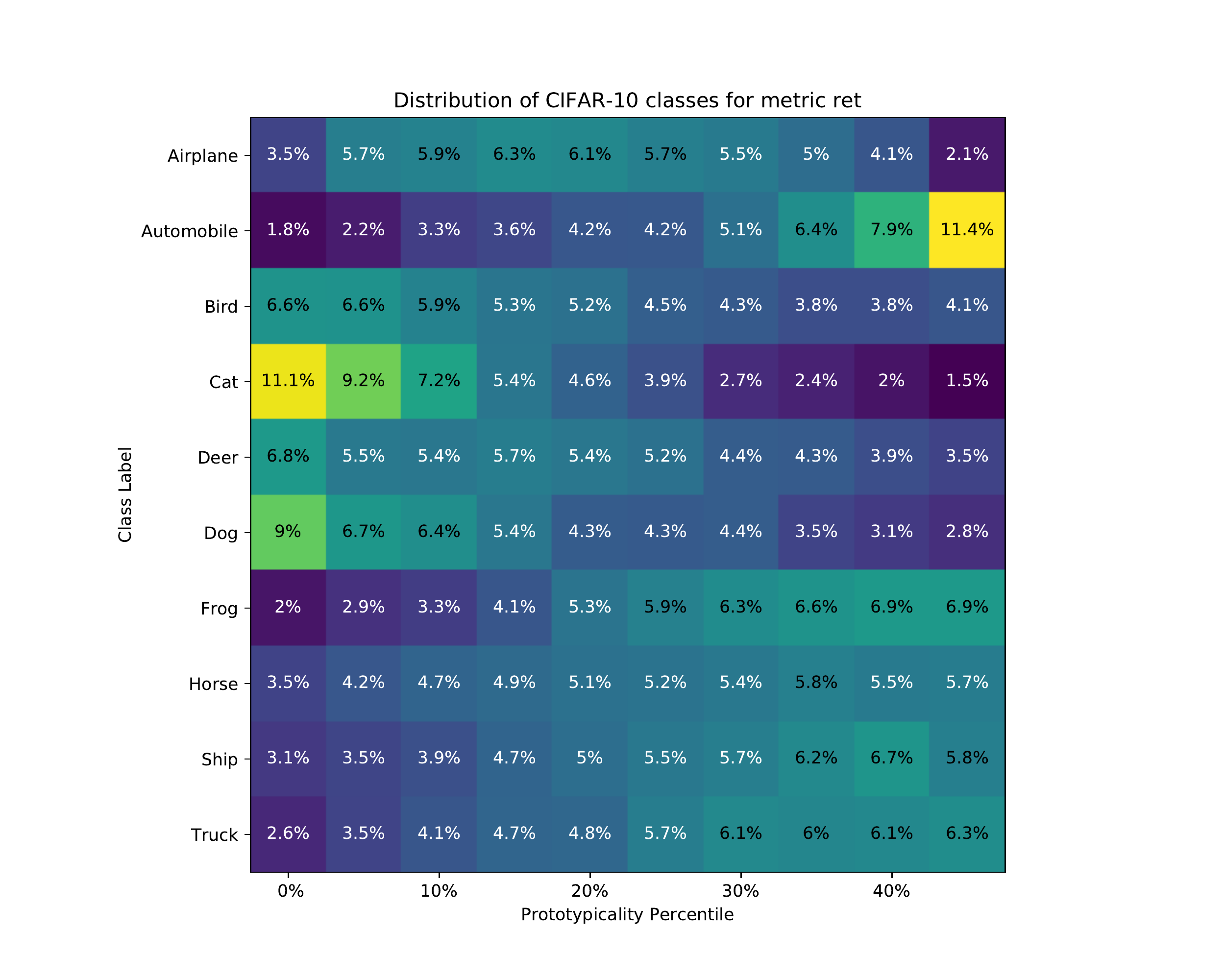}

\newpage
\section{All memorized exceptions for all Fashion-MNIST classes}
Below are all the memorized exceptions, as defined in the body of the paper,
for all Fashion-MNIST output classes:
\begin{itemize}
\item Tshirt/top
\item {Trouser} 
\item {Pullover}
\item {Dress}
\item {Coat} 
\item {Sandal}
\item {Shirt}
\item {Sneaker} 
\item {Bag} 
\item {Ankle boot}
\end{itemize}
\includegraphics[trim=1cm 9cm 1cm 1cm,clip,width=.99\linewidth]{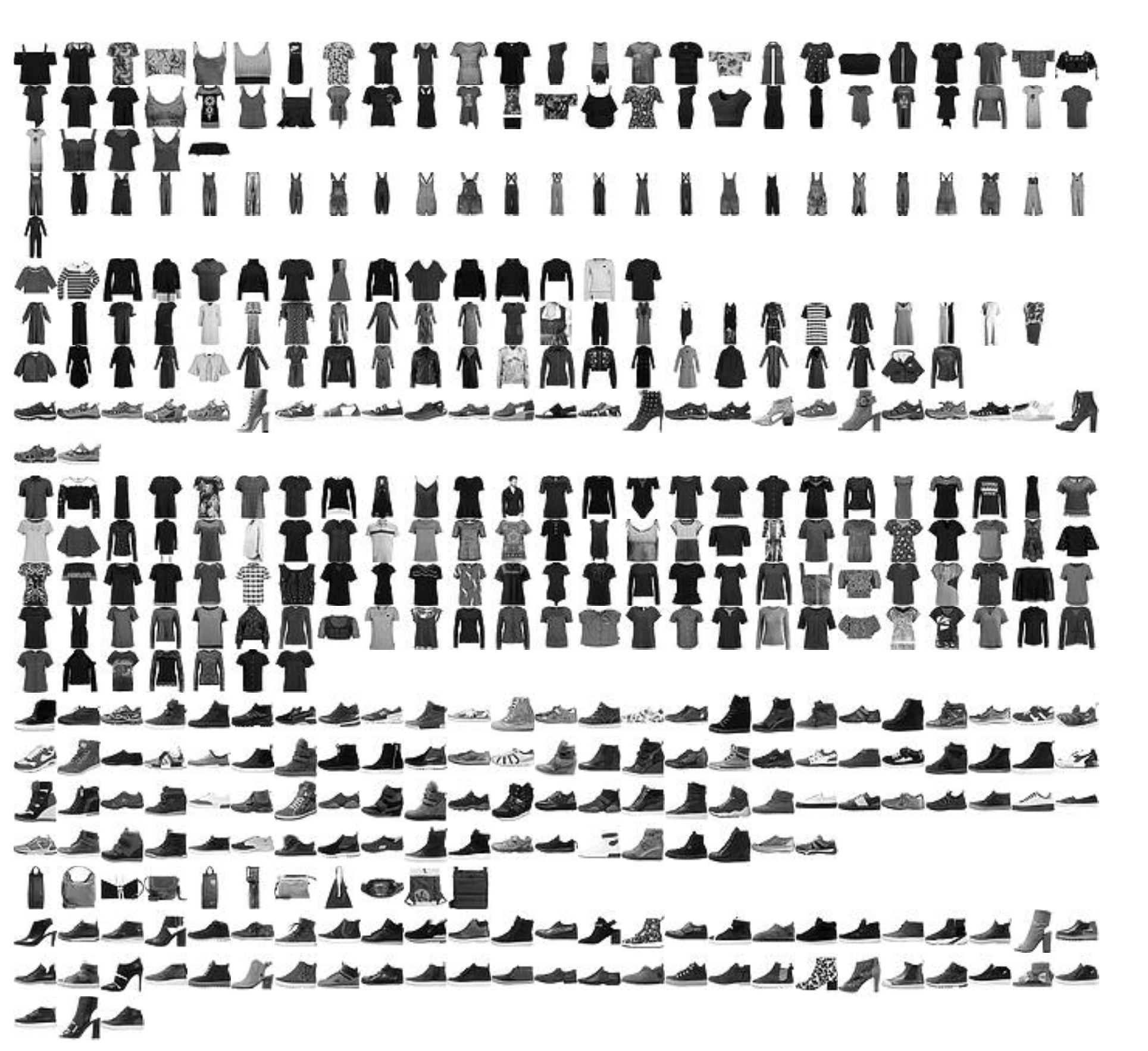}

\newpage
\section{ImageNet Memorized Exceptions}


\section*{Supplementary material}

# 14    ImageNet Memorized Exceptions

Index 6 stingray

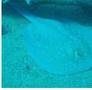

Index 7 cock

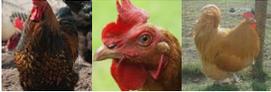

Index 8 hen

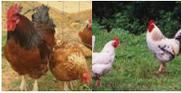

Index 32 tailed frog, bell toad, ribbed toad, tailed toad, Ascaphus trui

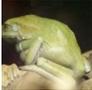

Index 36 terrapin

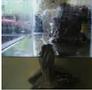

Index 40 American chameleon, anole, Anolis carolinensis

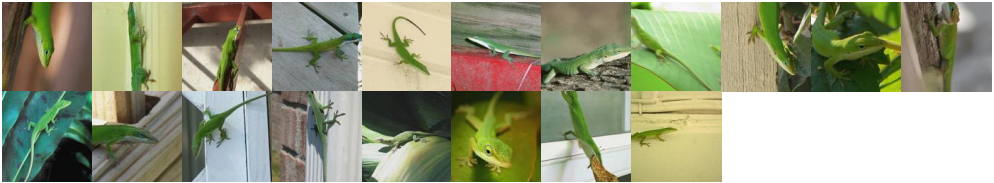

Index 46 green lizard, Lacerta viridis

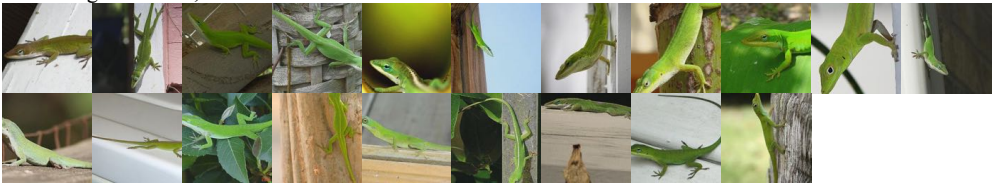

Index 66 horned viper, cerastes, sand viper, horned asp, Cerastes cornutus

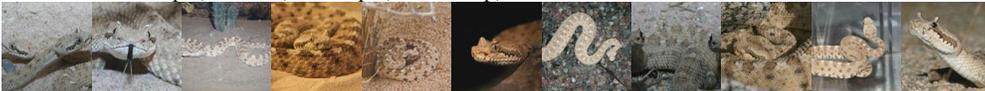

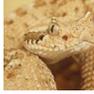

Index 68 sidewinder, horned rattlesnake, Crotalus cerastes

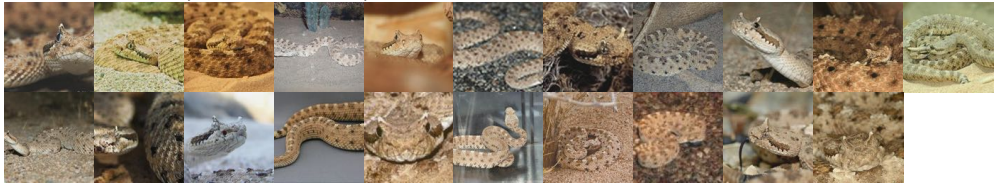

Index 72 black and gold garden spider, Argiope aurantia

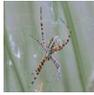

Index 101 tusker

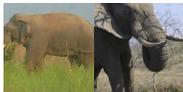

Index 103 platypus, duckbill, duckbilled platypus, duck-billed platypus, Ornithorhynchus anatinus

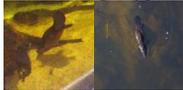

Index 122 American lobster, Northern lobster, Maine lobster, Homarus americanus

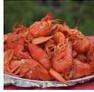

Index 124 crayfish, crawfish, crawdad, crawdaddy

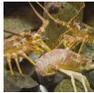

Index 126 isopod

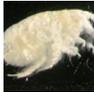

Index 161 basset, basset hound

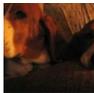

Index 166 Walker hound, Walker foxhound

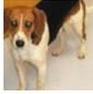

Index 167 English foxhound

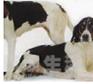

Index 170 Irish wolfhound

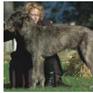

Index 179 Staffordshire bullterrier, Staffordshire bull terrier

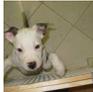

Index 180 American Staffordshire terrier, Staffordshire terrier, American pit bull terrier, pit bull terrier

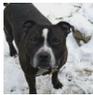

Index 196 miniature schnauzer

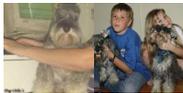

Index 197 giant schnauzer

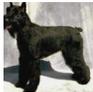

Index 198 standard schnauzer

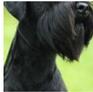

Index 206 curly-coated retriever

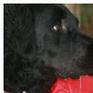

Index 214 Gordon setter

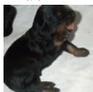

Index 223 schipperke

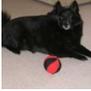

Index 238 Greater Swiss Mountain dog

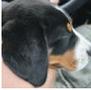

Index 248 Eskimo dog, husky

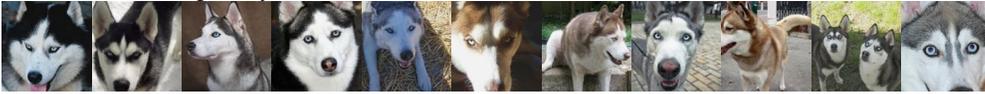

Index 250 Siberian husky

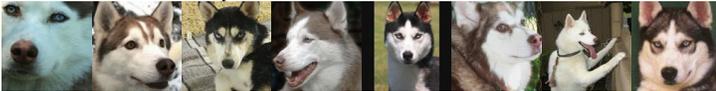

Index 264 Cardigan, Cardigan Welsh corgi

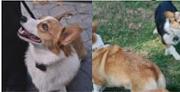

Index 265 toy poodle

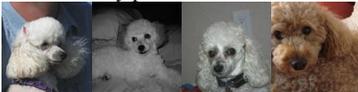

Index 266 miniature poodle

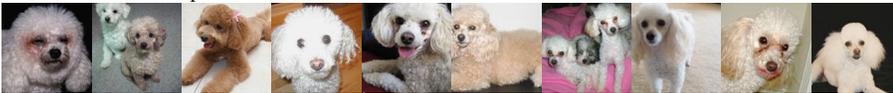

Index 270 white wolf, Arctic wolf, Canis lupus tundrarum

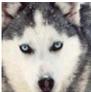

Index 278 kit fox, Vulpes macrotis

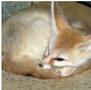

Index 290 jaguar, panther, Panthera onca, Felis onca

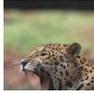

Index 304 leaf beetle, chrysomelid

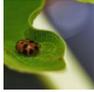

Index 311 grasshopper, hopper

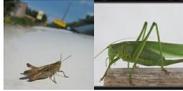

Index 312 cricket

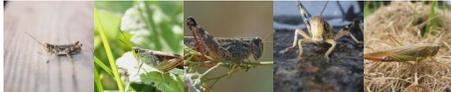

Index 319 dragonfly, darning needle, devil's darning needle, sewing needle, snake feeder, snake doctor, mosquito hawk, skeeter hawk

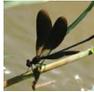

Index 320 damselfly

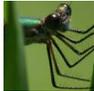

Index 334 porcupine, hedgehog

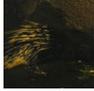

Index 341 hog, pig, grunter, squealer, Sus scrofa

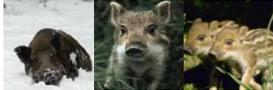

Index 342 wild boar, boar, Sus scrofa

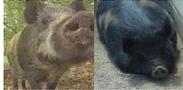

Index 345 ox

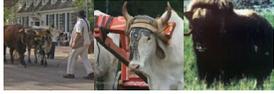

Index 348 ram, tup

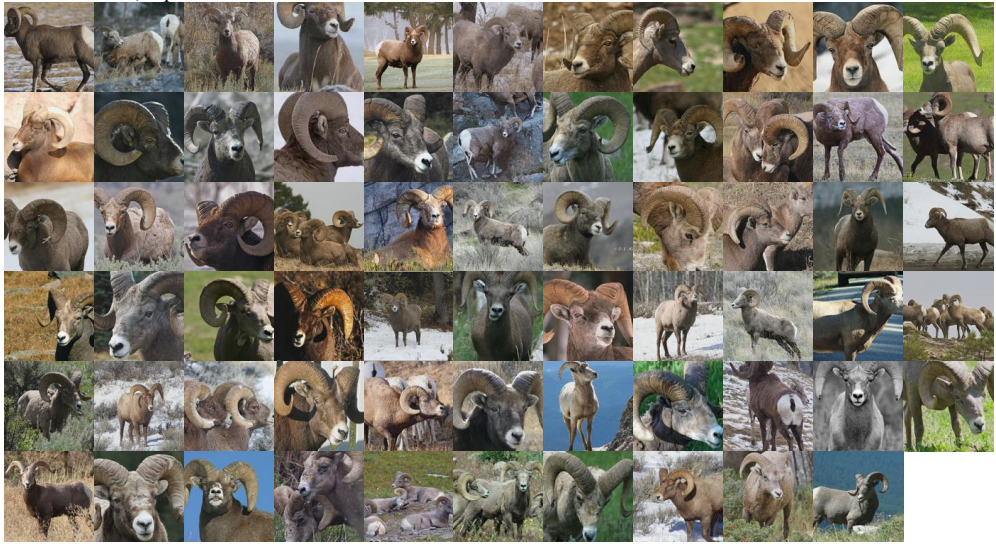

Index 349 bighorn, bighorn sheep, cimarron, Rocky Mountain bighorn, Rocky Mountain sheep, Ovis canadensis

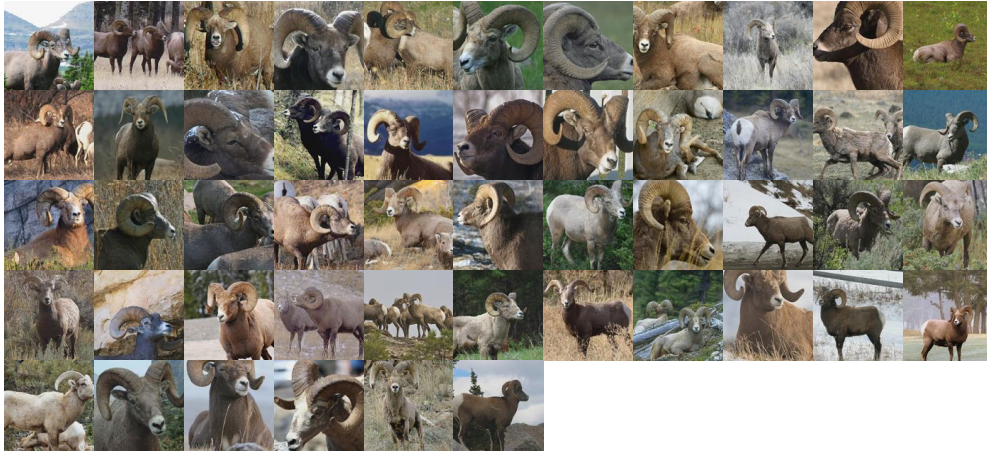

Index 359 black-footed ferret, ferret, Mustela nigripes

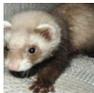

Index 380 titi, titi monkey

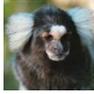

Index 383 Madagascar cat, ring-tailed lemur, Lemur catta

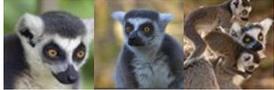

Index 386 African elephant, Loxodonta africana

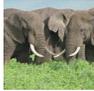

Index 390 eel

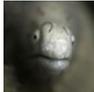

Index 399 abaya

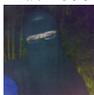

Index 400 academic gown, academic robe, judge's robe

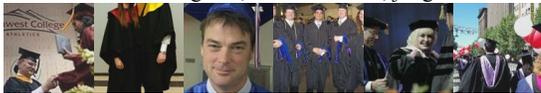

Index 409 analog clock

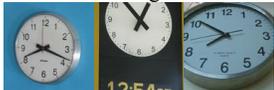

Index 413 assault rifle, assault gun

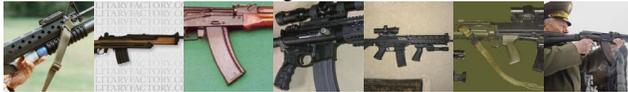

Index 417 balloon

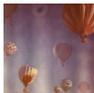

Index 419 Band Aid

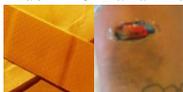

55

Index 423 barber chair

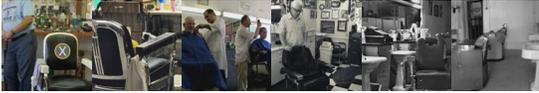

Index 424 barbershop

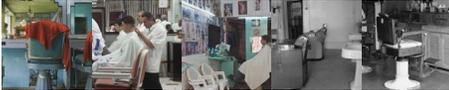

Index 429 baseball

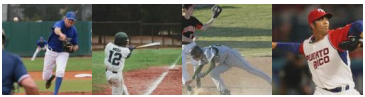

Index 434 bath towel

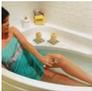

Index 435 bathtub, bathing tub, bath, tub

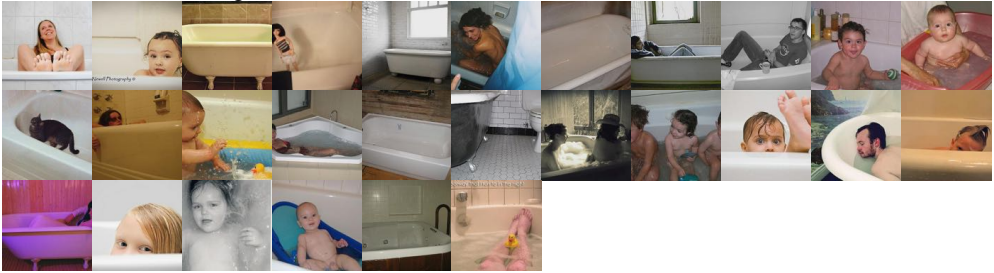

Index 440 beer bottle

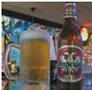

Index 461 breastplate, aegis, egis

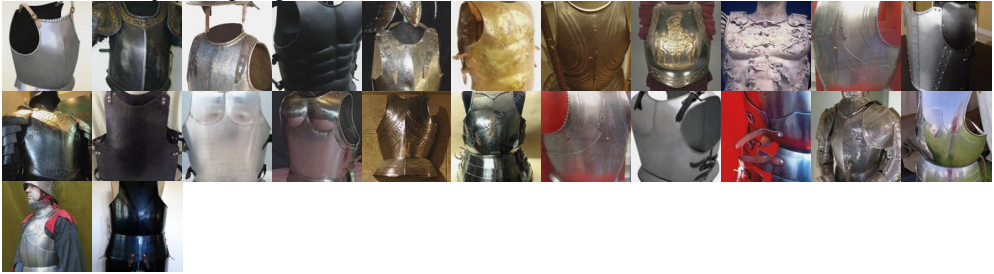

Index 465 bulletproof vest

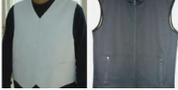

Index 479 car wheel

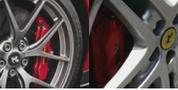

Index 484 catamaran

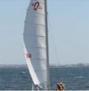

Index 505 coffeepot

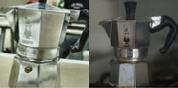

Index 516 cradle

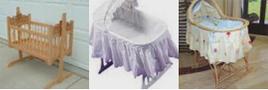

Index 524 cuirass

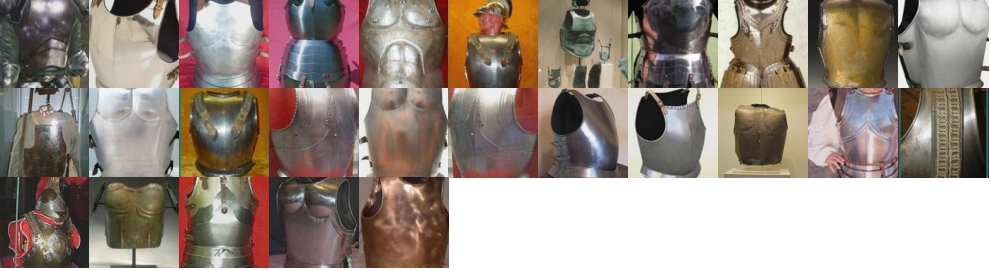

Index 538 dome

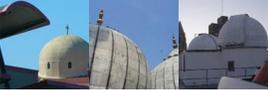

Index 541 drum, membranophone, tympan

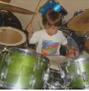

Index 550 espresso maker

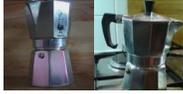

Index 579 grand piano, grand

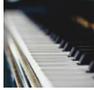

Index 583 guillotine

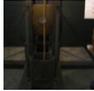

Index 591 handkerchief, hankie, hanky, hankey

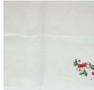

Index 595 harvester, reaper

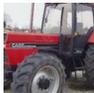

Index 604 hourglass

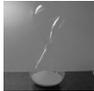

Index 619 lampshade, lamp shade

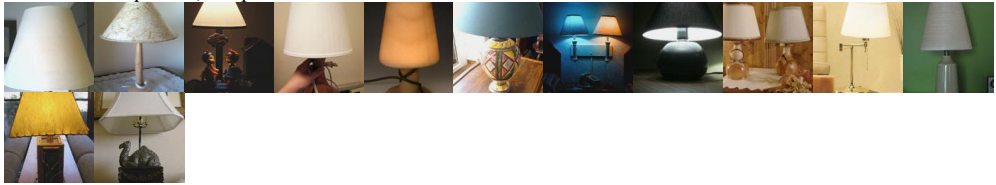

Index 620 laptop, laptop computer

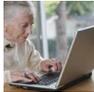

Index 636 mailbag, postbag

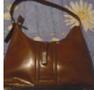

Index 638 maillot

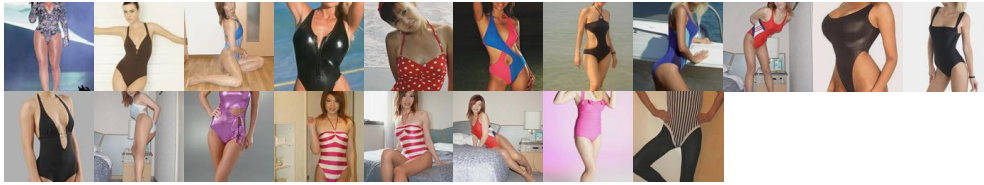

Index 639 maillot, tank suit

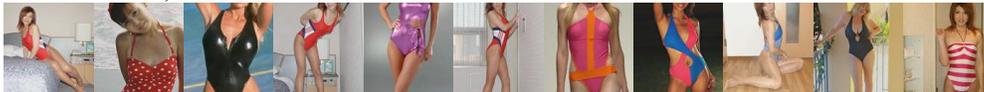

Index 643 mask

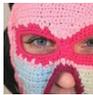

Index 647 measuring cup

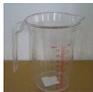

Index 657 missile

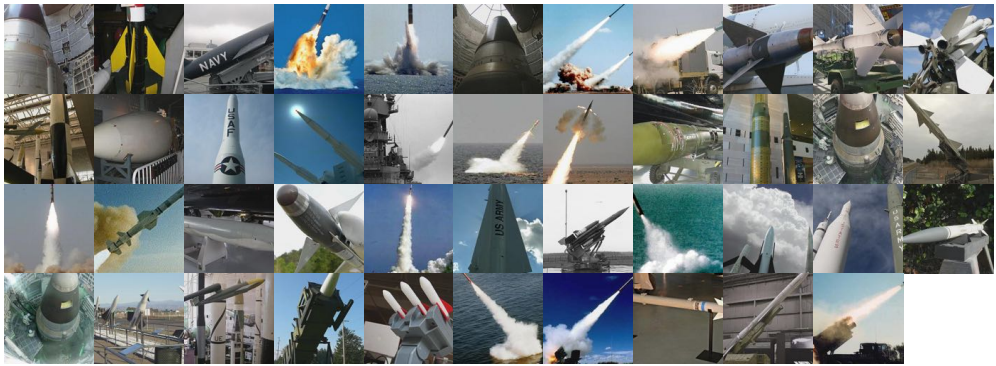

Index 665 moped

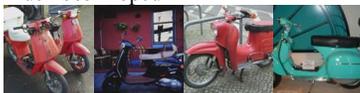

Index 667 mortarboard

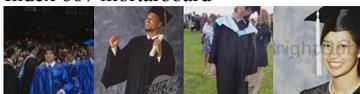

Index 668 mosque

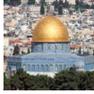

Index 670 motor scooter, scooter

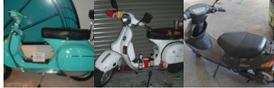

Index 678 neck brace

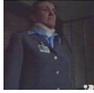

Index 681 notebook, notebook computer

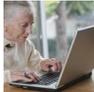

Index 700 paper towel

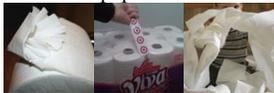

Index 739 potter's wheel

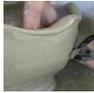

Index 744 projectile, missile

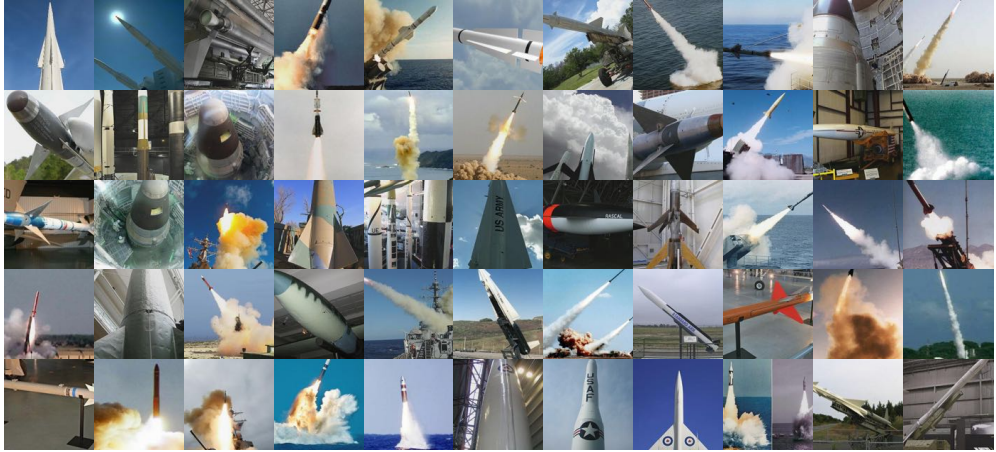

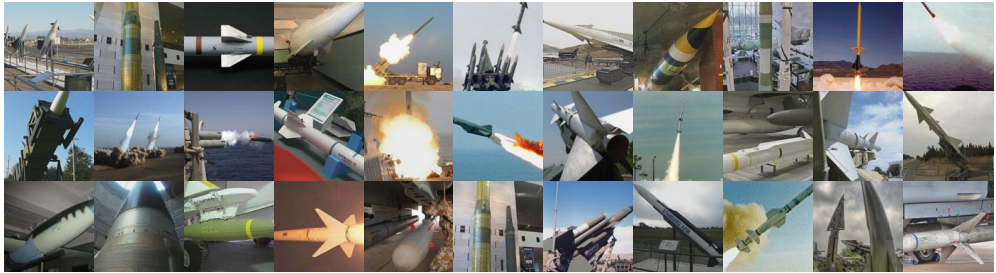

Index 748 purse

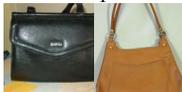

Index 764 rifle

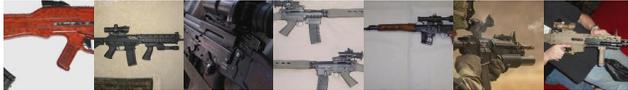

Index 804 soap dispenser

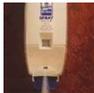

Index 808 sombrero

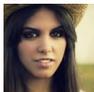

Index 810 space bar

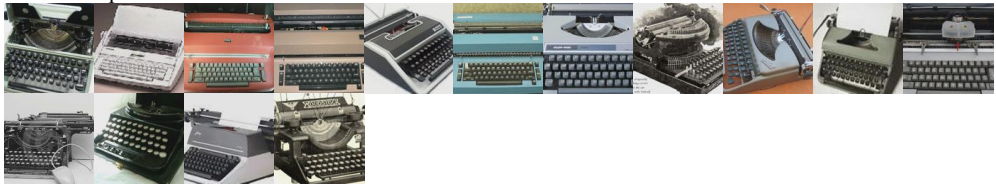

Index 817 sports car, sport car

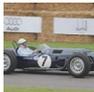

Index 827 stove

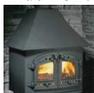

Index 830 stretcher

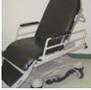

Index 836 sunglass

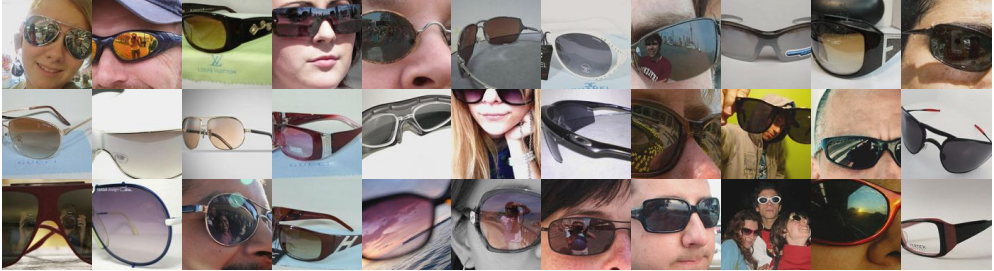

Index 837 sunglasses, dark glasses, shades

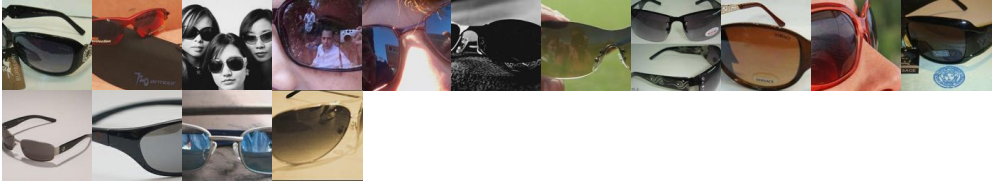

Index 841 sweatshirt

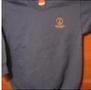

Index 842 swimming trunks, bathing trunks

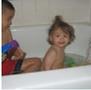

Index 846 table lamp

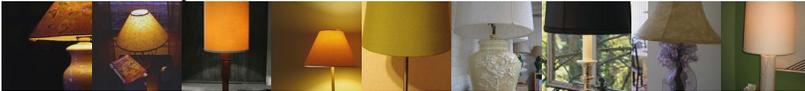

Index 847 tank, army tank, armored combat vehicle, armoured combat vehicle

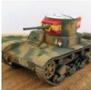

Index 876 tub, vat

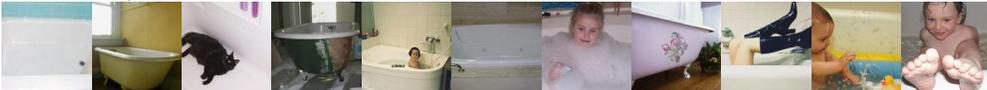

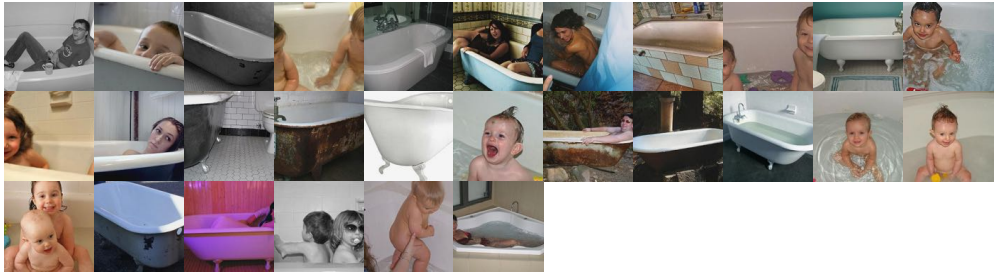

Index 878 typewriter keyboard

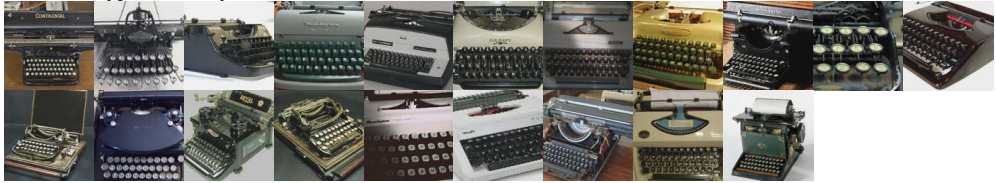

Index 892 wall clock

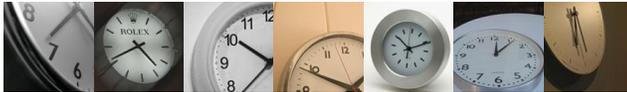

Index 903 wig

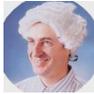

Index 907 wine bottle

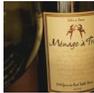

Index 914 yawl

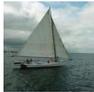

Index 925 consomme

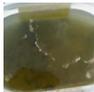

Index 928 ice cream, icecream

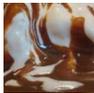

Index 939 zucchini, courgette

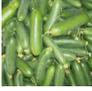

Index 954 banana

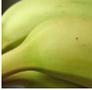

Index 960 chocolate sauce, chocolate syrup

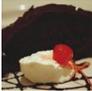

Index 961 dough

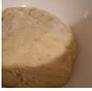

Index 962 meat loaf, meatloaf

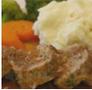

Index 966 red wine

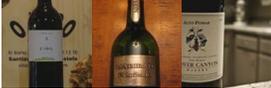

Index 981 ballplayer, baseball player

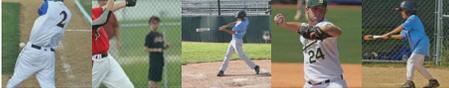

Index 987 corn

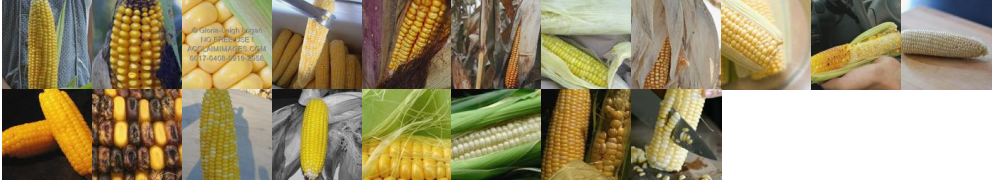



\end{document}